\documentclass[10pt,twocolumn,letterpaper]{article}
\usepackage[pagenumbers]{cvpr}

\usepackage{graphicx}
\usepackage{amsmath}
\usepackage{amsthm}           
\usepackage{booktabs}
\usepackage{epsfig}
\usepackage{tabularx}
\usepackage[table,dvipsnames]{xcolor}      
\usepackage{verbatim}           
\usepackage{relsize}
\usepackage{multirow}
\usepackage{scrextend}          
\usepackage{array}              
\usepackage{makecell}           
\usepackage{subcaption}          
\usepackage[nointegrals]{wasysym}
\usepackage{lipsum}             
\usepackage{esvect}              
\usepackage{soul}               
\usepackage{float}
\usepackage{caption}
\usepackage{bm}
\usepackage{enumitem}           
\usepackage{empheq}
\usepackage{gensymb}            
\usepackage[thicklines]{cancel} 
\usepackage{arydshln}           
\usepackage{pifont}             
\usepackage{hhline}             
\usepackage{tocloft}            
\usepackage{xcolor}

\setlist[itemize, 1]{label =\raisebox{-0.4\height}{\scalebox{1.7}{\textbullet}}}
\setlist[itemize]{noitemsep, topsep=0cm, leftmargin=3mm}
\allowdisplaybreaks             
\graphicspath{{images/}}

\advance\cftsecnumwidth 0.25cm\relax
\advance\cftsubsecnumwidth 0.08cm\relax
\advance\cftsubsecnumwidth 0.15cm\relax

\makeatletter
\@namedef{ver@everyshi.sty}{}
\makeatother
\usepackage{tikz} 
\usetikzlibrary{shapes.geometric}
\usetikzlibrary{calc} 
\usetikzlibrary{arrows}
\usetikzlibrary{arrows.meta}
\usetikzlibrary{shapes.arrows}
\usetikzlibrary{fadings, shadows}
\usetikzlibrary{decorations.text}
\def\centerarc[#1](#2)(#3:#4:#5)
{ \draw[#1] ($(#2)+({#5*cos(#3)},{#5*sin(#3)})$) arc (#3:#4:#5); }

\definecolor{my_blue}{rgb}{0.2, 0.6, 1}  
\definecolor{my_magenta}{rgb}{1.0, 0.2, 0.6}
\definecolor{my_yellow}{rgb}{1.0, 0.8, 0.2} 
\definecolor{my_green}{rgb}{0.0, 0.9, 0.24}
\definecolor{my_green_2}{rgb}{0.0, 0.4, 0.0}
\definecolor{sns_blue}{rgb}{0.21, 0.06, 0.42}   
\definecolor{sns_violet}{rgb}{0.45, 0.12, 0.51} 
\definecolor{sns_orange}{rgb}{0.75, 0.22, 0.46} 
\definecolor{sns_yellow}{rgb}{0.99, 0.91, 0.66}  
\definecolor{white}{rgb}{1.0, 1.0, 1.0}
\definecolor{darkGreen}{rgb}{0.01, 0.8, 0.24}
\definecolor{darkGreen2}{rgb}{0.22,0.42, 0.33}
\definecolor{darkGreen3}{rgb}{0.20,0.66, 0.33}
\definecolor{cvprblue}{rgb}{0.21,0.49,0.74}
\definecolor{LightCyan}{rgb}{0.88,1,1}
\definecolor{lightgreen}{HTML}{90EE90}
\definecolor{new_green}{rgb}{0.75,0.97,0.44}
\definecolor{Gray}{gray}{0.95}
\definecolor{lightgray}{rgb}{0.96, 0.96, 0.96}
\definecolor{set1_cyan}{rgb}{0.23, 0.87, 1.0}
\definecolor{building}{rgb}{0.2, 0.33, 0.33}
\definecolor{my_violet}{rgb}{0.79, 0.40, 1} 
\definecolor{my_yellow_2}{rgb}{0.9, 0.8, 0.54}
\definecolor{my_red}{rgb}{1,0,0}
\definecolor{my_purple}{rgb}{0.27,0.8, 0.8}
\definecolor{my_orange}{rgb}{1.0,0.6,0.35}
\definecolor{my_golden}{rgb}{1.0, 0.75, 0.0}
\colorlet{my_gray}{gray!12}
\definecolor{projectionColor}{rgb}{0.2, 0.6, 1}
\definecolor{rayColor}{rgb}{0.0,0.0,0.0}
\definecolor{axisColor}{rgb}{0.0, 0.0, 0.0}
\colorlet{projectionBorderShade}{rayColor!100}
\colorlet{projectionFillShade}{projectionColor!20}
\colorlet{rayShade}{my_yellow}
\colorlet{axisShade}{axisColor!20}
\colorlet{axisShadeDark}{axisColor!100}
\definecolor{backward_color}{rgb}{1.0, 0.6, 0.2}
\definecolor{forward_color}{rgb}{0.2, 1.0, 0.6}
\definecolor{gain}{HTML}{34a853}
\definecolor{lost}{HTML}{ea4335}
\colorlet{proposedShade}{darkGreen}
\colorlet{vanillaShade}{red!90}
\colorlet{theme_color}{cvprblue}
\colorlet{theme_color_light}{sns_yellow!25}
\colorlet{link_color}{Maroon}

\newcommand{\noIndentHeading}[1]{\noindent\textbf{#1}}

\definecolor{XLcolor}{rgb}{0.858, 0.188, 0.478}

\newcommand{\forExample}{\textit{e.g.}\xspace}
\newcommand{\thatIs}{\textit{i.e.}\xspace}

\newcommand{\myTopRule}{\Xhline{2\arrayrulewidth}}

\newcolumntype{t}{!{\vrule width 1.5\arrayrulewidth}}
\newcolumntype{m}{!{\vrule width 2.5\arrayrulewidth}}
\newcolumntype{a}{>{\columncolor{theme_color_light}}l}
\newcolumntype{b}{>{\columncolor{theme_color_light}}c}

\colorlet{cyan_highlight}{my_blue!85}

\colorlet{darkGreen_highlight}{darkGreen!75}

\colorlet{my_magenta_highlight}{my_magenta!50}

\colorlet{my_yellow_highlight}{my_yellow!55}

\providecommand\rightarrowRHD{\relbar\joinrel\mathrel\RHD}

\newcommand{\uparrowRHD}  {\rotatebox[origin=c]{90}{$\rightarrowRHD$}}

\newcommand{\uparrowRHDSmall}  {\raisebox{0.05\normalbaselineskip}{\scalebox{0.7}{\uparrowRHD}}}

\newcommand{\oneD}{$1$D\xspace}
\newcommand{\twoD}{$2$D\xspace}
\newcommand{\threeD}{$3$D\xspace}

\newcommand{\lidar}{LiDAR\xspace}
\newcommand{\lidars}{LiDARs\xspace}
\newcommand{\radar}{radar\xspace}
\newcommand{\radars}{radars\xspace}
\newcommand{\Radar}{Radar\xspace}

\newcommand{\bev}{BEV\xspace}

\newcommand{\nuscenes}{nuScenes\xspace}

\newcommand{\sota}{SOTA\xspace}
\newcommand{\best}[1]{$\mathbf{#1}$}

\newcommand{\pgd}{PGD\xspace}

\newcommand{\sparseBEV}{SparseBEV\xspace}

\newcommand{\methodName}{RICCARDO\xspace}

\newcommand{\methodNameFull}{Radar Hit Prediction and Convolution for Camera-Radar  3D Object Detection\xspace}
\newcommand{\methodNameFullColored}{\textcolor{theme_color}{R}adar H\textcolor{theme_color}{i}t Predi\textcolor{theme_color}{c}tion and \textcolor{theme_color}{C}onvolution for C\textcolor{theme_color}{a}mera-\textcolor{theme_color}{R}adar\\3D Object \textcolor{theme_color}{D}etecti\textcolor{theme_color}{o}n\xspace}

\newcommand{\paperTitle}{\textcolor{theme_color}{\methodName}: \methodNameFullColored}

\newcommand{\stageOne}{Stage 1\xspace}
\newcommand{\stageTwo}{Stage 2\xspace}
\newcommand{\stageThree}{Stage 3\xspace}
\newcommand{\stageOneName}{RIC\xspace}
\newcommand{\stageOneNameFull}{Radar Hit Prediction\xspace}
\newcommand{\stageTwoNameFull}{Convolving \stageOneName with \Radar\xspace}
\newcommand{\stageThreeNameFull}{Camera-\Radar Candidate Selector\xspace}

\usepackage[pagebackref,breaklinks,colorlinks,allcolors=cvprblue]{hyperref}
\usepackage[capitalize]{cleveref}
\crefname{section}{Sec.}{Secs.}
\Crefname{section}{Section}{Sections}
\Crefname{table}{Table}{Tables}
\crefname{table}{Tab.}{Tabs.}

\def\paperID{17245}
\def\confName{CVPR}
\def\confYear{2025}

\title{\paperTitle}

\author{Yunfei Long\qquad Abhinav Kumar\qquad Xiaoming Liu\qquad Daniel Morris\\
	Michigan State University\\
	\small\tt \{longyunf,kumarab6,liuxm,dmorris\}@msu.edu
}

\begin{document}
	
	\maketitle
	
	\begin{abstract}
		\Radar hits reflect from points on both  the boundary and internal to object outlines.  
		This results in a complex distribution of \radar hits that depends on factors including object category, size, and orientation. Current \radar-camera fusion methods implicitly account for this with a black-box neural network.  
		In this paper, we explicitly utilize a \radar hit distribution model to assist fusion.  First, we build a model to predict \radar hit distributions conditioned on object properties obtained from a monocular detector.  Second, we use the predicted distribution as a kernel to match actual measured \radar points in the neighborhood of the monocular detections, generating matching scores at nearby positions. Finally, a fusion stage combines context with the kernel detector to refine the matching scores. Our method achieves the state-of-the-art \radar-camera detection performance on \nuscenes. Our source code is available at \url{https://github.com/longyunf/riccardo}.
	\end{abstract}
	
	\begin{figure}[!t]
		\centering
		\scalebox{1.0}
		{
			\begin{tabular}{@{}c@{}c@{}}
				\includegraphics[height=1in]{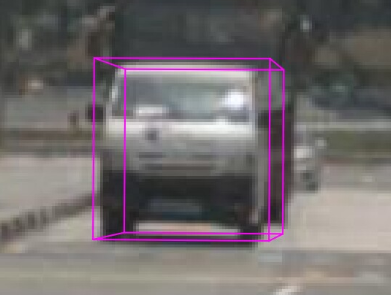}&
				\includegraphics[height=1.05in]{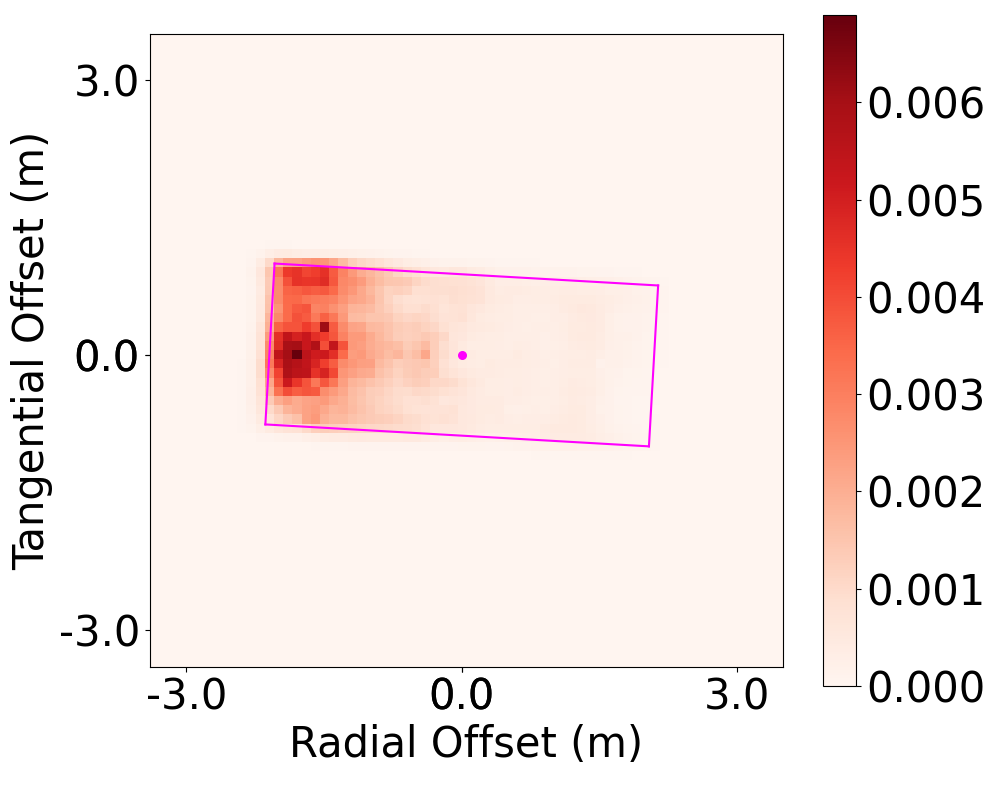}
				\\
				{\small (a)  } & {\small (b)  } \\
				\includegraphics[height=1.05in]{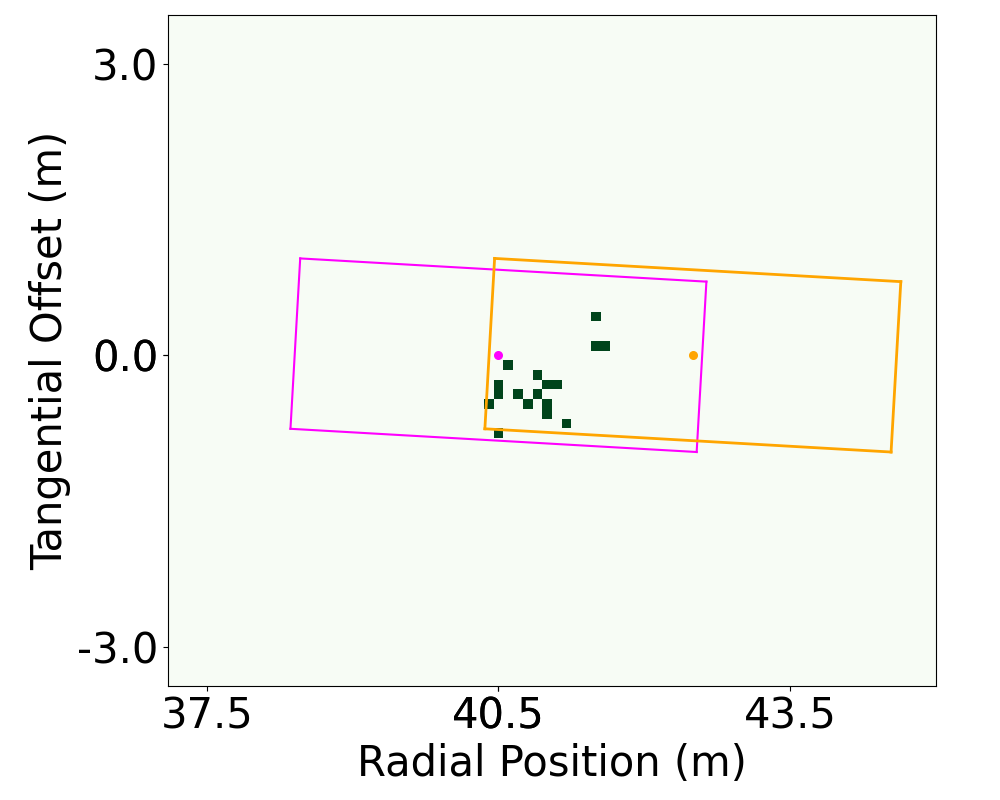}&
				\includegraphics[height=1in]{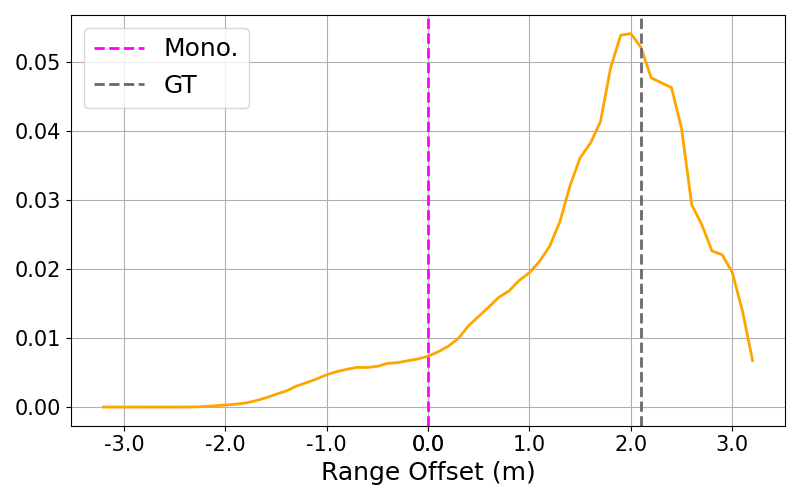}
				\\
				{\small (c)  } & {\small (d)  } \\
			\end{tabular}
		}
		\caption{\small Given a (a) monocular detection, we estimate (b) radar point distribution relative to its bounding box in BEV; then we shift the distribution and convolve it with (c) actual radar measurement in the neighborhood to compute (d) similarity scores and estimate an updated position, where the matching score is maximum. In (c) the monocular bounding box (in magenta) is misaligned with radar points; the updated position (in orange) with peak matching score shifts the box to a farther range so that relative positions of radar points match the predicted distribution (radar hits concentrated at the head of vehicle instead of in the middle). }
		\label{fig:intro}
	\end{figure}

	\section{Introduction}\label{sec:intro}
	
	\threeD object detection~\cite{geiger2012we, caesar2020nuscenes} is a key component of scene understanding in autonomous vehicles (AVs). It predicts nearby objects and their attributes including \threeD location, size, orientation, and category, setting the stage for navigation tasks such as path planning. The primary sensors used for \threeD object detection are cameras, \lidars{}~\cite{zamanakos2021comprehensive}, and \radar{}s, with the focus here on the two-sensor combination that is the least expensive and already ubiquitous on vehicles, namely cameras and \radar{}s. 
	This paper asks how to combine camera  and \radar data in order to achieve the best performance improvement over a single modality detection.
	
	Detection can be performed on camera and \radar individually, with each sensor having its  strengths and drawbacks. Cameras are inexpensive and capture high-resolution details and texture with state-of-the-art (\sota) methods~\cite{lu2021geometry, park2021pseudo, kumar2022deviant} achieving accurate object classification, as well as estimating size and orientation. One of the primary limitations is the depth-scale ambiguity, resulting in relatively inaccurate object depth estimation~\cite{lu2021geometry, kumar2022deviant}. In contrast, current automotive \radar is another inexpensive sensor that directly measures range to target, as well as Doppler velocity, and is robust to adverse weather such as rain, snow, and darkness \cite{nabati2021centerfusion}. The drawbacks of \radar are its very sparse scene sampling and lack of texture, making it challenging to perform tasks such as object categorization, orientation, and size estimation. For example, \radar point clouds collected in \nuscenes dataset~\cite{caesar2020nuscenes} are \twoD points on \radar bird's-eye view (\bev) plane without height measurements (the default height is zero). This comparison shows that the strengths of \radar and camera are complementary, and indeed this paper explores how we can combine information from these two modalities to improve \threeD object detection.
	
	While \radar and \lidar are both widely used depth sensors for AVs, they differ significantly in their target sampling characteristics.   \lidar points align densely with the edges of objects, and for vehicles typical form a distinct ``L" or ``I" distribution.  These regular distributions are aligned with the object pose and enable precise shape and pose estimation from \lidar scans as evidenced by top-performing \lidar methods~\cite{zhang2023fully} on nuScenes achieving a mean Average Precision (mAP) of 0.702 and nuScenes detection score (NDS) of 0.736.  In contrast, \radars have wide beam-width with low angular resolution and often penetrate objects or reflect from their undersides.  This leads to a much sparser and dispersed distribution of hits on objects.  From this distribution it is much more challenging to estimate target shape, category, and pose, and the top-performing \radar-only method, RadarDistill~\cite{bang2024radar}, achieves a far lower mAP of 0.205 and NDS of 0.437 on nuScenes.
	
	The combination of camera with radar has potential to significantly enhance \radar-only methods, with camera data providing strong results on object category, shape, and pose.  Radar, with its direct range measurements, can contribute range and velocity to a combined detector.  Existing \radar-camera models combine camera and radar via concatenation~\cite{nabati2021centerfusion}, (weighted) sum~\cite{kim2020grif}, or attention-based operations~\cite{kim2023craft, kim2023crn}. But obtaining measurable gains by fusing sparse radar and camera is difficult, with the top \radar-camera method~\cite{kim2023crn} having lower performance than camera-only methods~\cite{liu2023sparsebev}.
	
	To address the difficulty in combining disparate modalities,  we conjecture that directly \textit{modeling the \radar distribution's dependency on target properties} will enable \radar returns to be more effectively aligned and leveraged in object detection.
	As shown in Fig.~\ref{fig:intro}, we introduce a model that generates a \radar hit prediction (RIC) in \bev from a given monocular detection priors, \thatIs, bounding box size, relative pose, and object class. 
	By moving a distribution kernel in the neighborhood of monocular estimated center and computing the similarity between the kernel and actual \radar measurements, we identify potential object centers with high similarity. Those position candidates are passed through a refinement stage for final position estimation.
	
	In nutshell, this paper makes the following contributions:
	\begin{itemize}
		\item Builds a model to predict \radar hit distributions relative to reflecting objects on \bev.
		\item Proposes position estimation by convolving predicted \radar distribution with actual \radar measurements.
		\item Achieves \sota performance on the \nuscenes dataset.
	\end{itemize}

	\begin{figure*}[!t]
		\centering
		\includegraphics[width=0.96\linewidth]{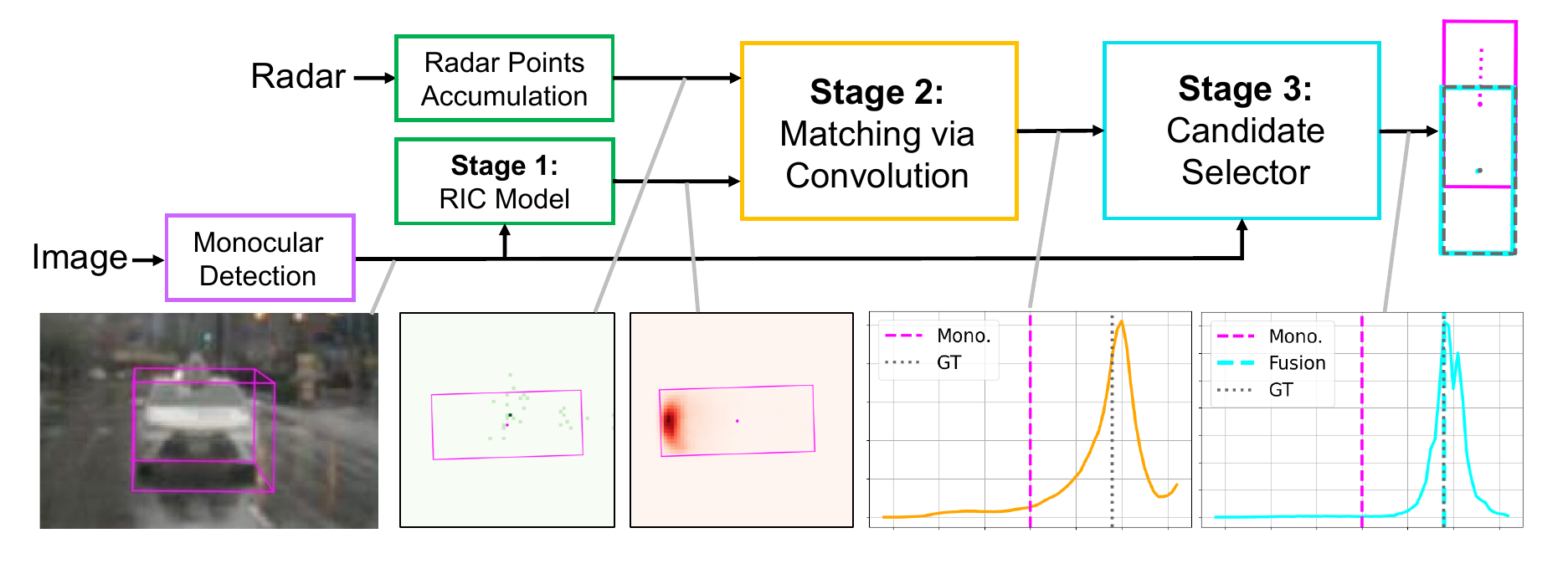}
		\vspace{-0.3cm}
		\caption{\textbf{\methodName inference}. \methodName leverages a monocular detector to identify objects and estimate their attributes (category, size, orientation, and approximate range) and involves three stages. Its \textbf{\stageOne} then predicts the \radar hit distribution (RIC) for each object. \textbf{\stageTwo} bins and convolves the observed accumulated \radar returns with the RIC, to generate a matching score over range. A final \textbf{\stageThree} fusion refines these scores to yield a precise target range estimate.
		}
		\label{fig:overall}
	\end{figure*}

	\section{Related Works}
	\noIndentHeading{Monocular 3D Detection.}
	Monocular \threeD detection is known for its low cost and simple setup, attracting extensive research optimizing every component in the detection pipeline, \forExample{}, architectures~\cite{wang2021detr3d,brazil2023omni3d,brazil2019m3d}, losses~\cite{simonelli2020disentangling,kumar2024seabird,brazil2020kinematic}, and NMS~\cite{kumar2021groomed}. Efforts to alleviate the intrinsic ambiguities from camera image to \threeD include incorporating estimated monocular depth~\cite{wang2019pseudo,simonelli2021we,reading2021categorical}, designing special convolutions \cite{kumar2022deviant}, considering camera pose \cite{zhou2021monoef}, and taking advantage of CAD models \cite{liu2021autoshape}. However, depth ambiguities still remain a bottleneck to performance \cite{ma2021delving,kumar2024seabird} and hence, fusion with \radar is a promising strategy for enhancing depth estimation while keeping low computation cost.
	
	\noIndentHeading{Camera-Radar Fusion for Detection.} 
	Camera-radar fusion has been widely applied to different vision tasks, {\it e.g.}, depth estimation~\cite{lin2020depth, long2021radar, singh2023depth}, semantic segmentation~\cite{yao2023radar}, target velocity estimation~\cite{long2021full, pandya2021velocity}, detection~\cite{yao2023radar, long2023radiant}, and tracking~\cite{tang2021road}. For detection, various camera-\radar methods have been proposed which differ in  representations~\cite{kim2023craft,nabati2021centerfusion}, fusion level \cite{kim2020grif,nabati2021centerfusion}, space \cite{kim2023crn,kim2020grif,zhou2023bridging}, and strategies \cite{wu2023mvfusion}.
	There are different \radar point representations. As \twoD \radar measures \bev XY locations without height, it is natural to represent \radar points in binned \bev space~\cite{kim2023crn}. To associate \radar points with camera source, \radar points are also modeled as pillars~\cite{nabati2021centerfusion, wu2023mvfusion} with fixed height in \threeD space. In addition, \radar points are represented as point feature~\cite{kim2023craft} and each point as a multi-dimensional vector with elements of \radar locations and other properties from measurement such as radar cross-section.
	
	\Radar and image sensors can interact at either the feature-level~\cite{kim2020grif} or the detection-level~\cite{nabati2021centerfusion}. While intermediate image features offer a wealth of raw information, such as textures, detection-level outputs provide directly interpretable information with clear physical meanings, making them well-suited for fusion tasks. With the rapid advancement of monocular \threeD detectors, leveraging detection-level image information allows us to capitalize on their accurate estimates of object category, size, orientation, and focus on refining position estimation, particularly range estimation, where \radar excels as a range sensor. Therefore, in this paper, we utilize camera information at the detection level.
	
	The fusion is conducted in image view~\cite{nabati2021centerfusion,long2023radiant} , \bev~\cite{kim2023crn} or a mix of the two~\cite{kim2020grif}. 
	Image features are naturally in image view, while \radar is in \bev. 
	To combine them in one space, fusion methods either project \radar points to image space~\cite{nabati2021centerfusion,long2023radiant} or lift image features to \threeD space~\cite{kim2023crn, zhou2023bridging}. Image view suffers from overlappings and occlusions while it is imprecise to transform from image view to \bev without reliable depths. In this work, we adopt \bev space for fusion as we use image source monocular detections, which are already in \threeD space.
	
	The \radar-camera association is conducted by associating the \radar pillars with monocular boxes in \threeD space~\cite{nabati2021centerfusion} or projecting the pillars on image to extract corresponding image features~\cite{wu2023mvfusion}. 
	However, none of these methods explicitly leverage \radar point distributions to address the misalignment problem in \radar-camera fusion.

	\section{\methodName}
	Our goal is to enhance object position estimation using \radar returns, surpassing the capabilities of monocular vision. The challenge lies in the sparse and non-obvious alignment of \radar returns with object boundaries and features, unlike the dense and consistent \lidar returns.
	To address this, we propose a method that explicitly models the statistics of \radar returns on objects, taking its category, size, orientation, range, and azimuth into account. 
	These statistics enable \radar returns to improve monocular detections. 
	
	Our approach, called \methodNameFull (\methodName), is illustrated in \cref{fig:overall} and involves three stages. The first stage predicts the \radar distribution returns on an object based on monocular detector outputs. The second stage convolves the predicted distribution with accumulated and binned \radar measurements to obtain a range-based score. The third and final stage refines the range-based score to obtain a final range estimate.
	We describe the details of each stage below.

	\subsection{\stageOne: \stageOneNameFull (\stageOneName) Model}
	
	The \stageOneName model aims to predict \radar hit position distributions on objects in BEV, conditioned on object category, size, orientation, range, and azimuth. This model leverages monocular detection data to predict \radar returns as a distribution, enabling comparison with actual measured returns (\cref{sec:matching}). This section details the construction and learning of this predictive model. 
	We model \radar hit distributions as a probability of \radar return over a set of grid cells.

	\noIndentHeading{Coordinate System.} A key choice is the coordinate system to predict this distribution. Possibilities include object-aligned, sensor-aligned or ego-vehicle aligned systems.  
	We choose an object-aligned system for modeling \radar hit position distributions. This choice allows for more gradual changes in probabilities as a function of relative sensor location compared to ego-vehicle or sensor-aligned systems, which exhibit significant variations with changes in relative object pose. Consequently, the object-aligned system facilitates learning from limited data.
	
\begin{figure}[!tb]
		\centering
		\includegraphics[width=\linewidth]{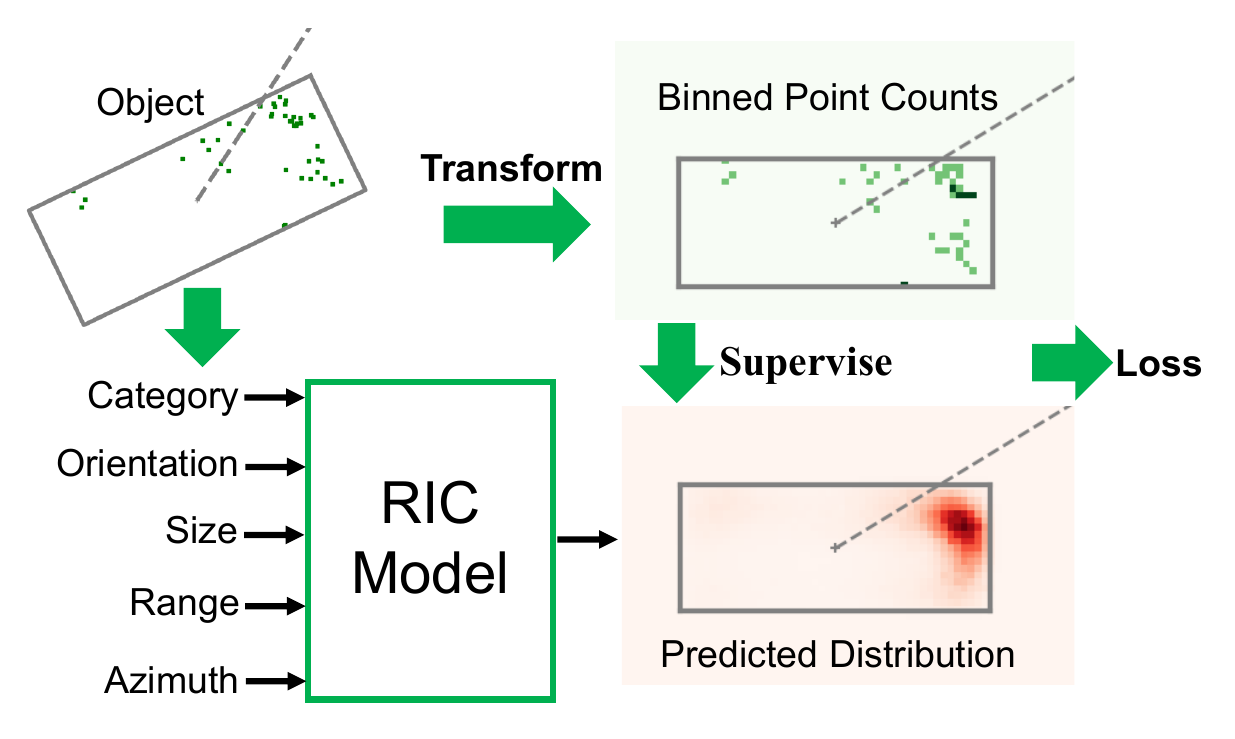}
		\caption{\textbf{Stage-1 \stageOneName Model} Training. The radial ray from ego to target is plotted as dashed line for reference.}
		\label{fig:stg1}
	\end{figure}

	\noIndentHeading{Architecture.}
	\cref{fig:stg1} shows the overview of \stageOne in \methodName.
	\stageOne employs a neural network to predict the distribution of \radar points in object-centric \bev coordinates, conditioned on object category, size, orientation, range, and azimuth. 
	The output is a \twoD quantized \bev probability map centered on the object, with X and Y axes 
	aligned to the object's length and width dimensions.
	The \stageOneName pixel value represents the density of \radar hits at those locations.  
	The network architecture comprises a multilayer perceptron (MLP) model, with parallel preprocessing branches for input parameters and a main branch that fuses these features and predicts the \stageOneName. 
	
	\noIndentHeading{Ground Truth Distribution.} To construct a ground truth (GT) distribution for a given target, we define a grid relative to the target's known position and accumulate \radar hits over a short time interval. 
	The normalized density of returns for a grid cell $i$, $\bar{P}_i$, is calculated as:
	\begin{align}
		\bar{P}_i &= \frac{c_i}{\sum_{i=1}^{N} c_i},
		\label{eq:gt_prob}
	\end{align}
	where $c_i$ is the number of \radar returns in grid cell $i$, and $N$ is the total number of cells on the  \stageOneName map. 
	This model only includes the points within the object boundary.  
	
	Targets may move during an accumulation period in a driving scene, and directly accumulating \radar hits causes smearing.  Thus, we offset each \radar hit position by the object motion between the \radar return and the final target position.  A pair of sequential annotated locations can provide GT object velocity, $\bm{v}_\text{O}$, for calculating this.  
	Yet, \radar Doppler velocity is more accurate for the radial component; thus, we combine the Doppler velocity, $\bm{v}_\text{D}$, with the tangential component of $\bm{v}_\text{O}$.  Given a unit vector perpendicular to radar ray,  $\bm{n}_\text{T}$, the offset $\bm{m}$ of a \radar point is:
	\begin{align}
		\bm{m} = \left( (\bm{v}_\text{O} \cdot \bm{n}_\text{T}) \bm{n}_\text{T}  + \bm{v}_\text{D}\right)\Delta t,
		\label{eq:motion}
	\end{align}
	where $\Delta t$ is the time between the \radar measurement and current sweep.  
	We apply these offsets before calculating the probabilities in \cref{eq:gt_prob}.

	\noIndentHeading{Loss Function.}
	The loss function incorporates cross-entropy loss $L_\text{CE}$ between the predicted \radar distribution $P$, and the GT distribution from the accumulated \radar returns, $\bar{P}$. In addition, we include a smoothness regularization term $L_\text{S}$ on $P$ to encourage spatial smoothness: 
	\begin{align}
		L_\text{S} &= \frac{1}{N_\text{c}(N_\text{r}\!-\!1)}\sum_{j=1}^{N_\text{c}} \sum_{i=1}^{N_\text{r}-1} |P(i,j)\!-\!P(i+1,j)| \nonumber \\  
		&+ \frac{1}{N_\text{r}(N_\text{c}\!-\!1)}\sum_{i=1}^{N_\text{r}} \sum_{j=1}^{N_\text{c}-1} |P(i,j)\!-\!P(i,j+1)|,
		\label{eq:l_s}
	\end{align}
	where $i$ and $j$ are pixel indices within $N_\text{r}\times N_\text{c}$ grid of rows and columns.
	Thus, the total loss for training is $L_\text{CE} + L_\text{S}$.
	
	\begin{figure}[!t]
		\centering
		\includegraphics[width=0.9\linewidth]{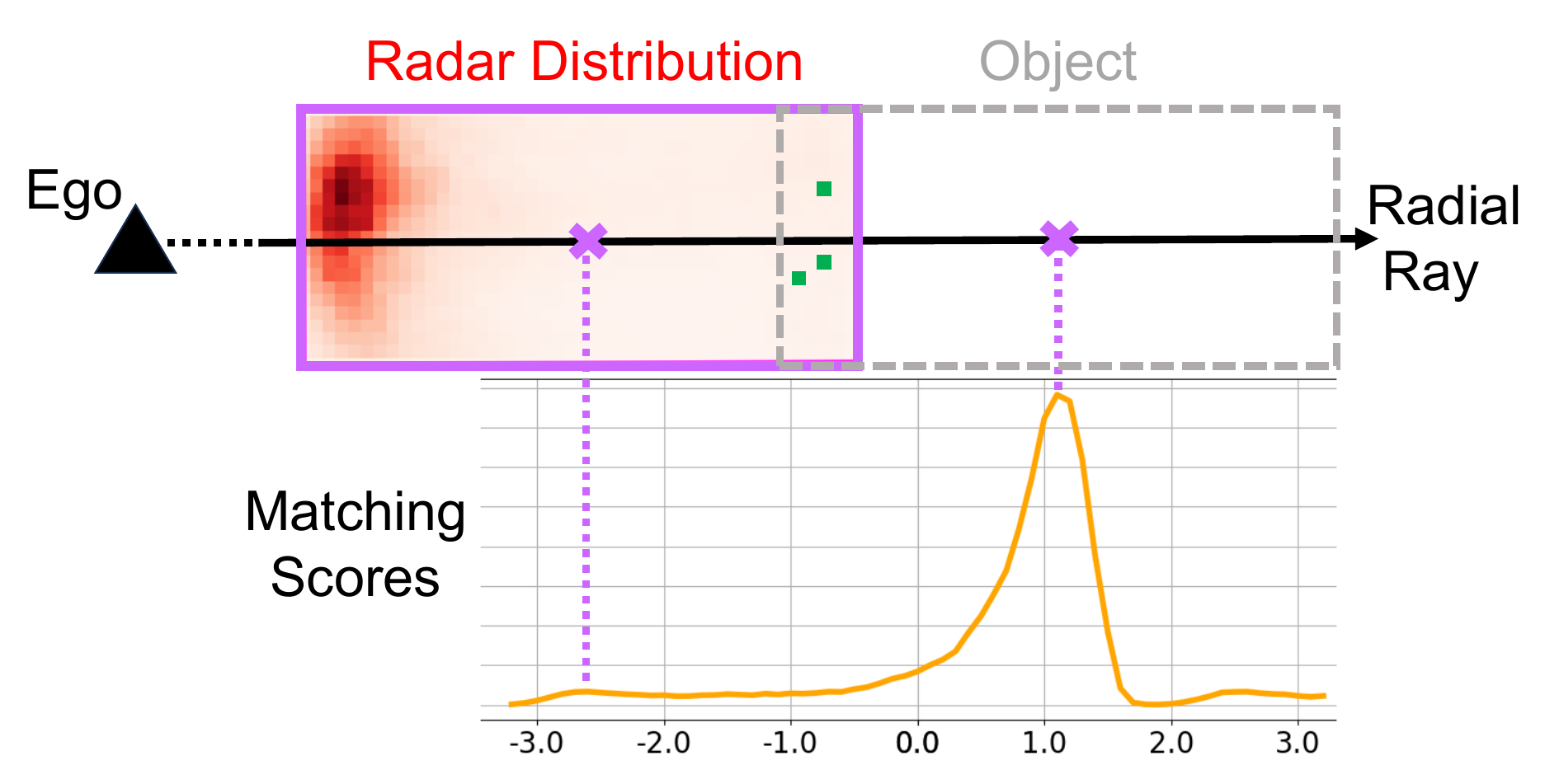}
		\caption{\textbf{\stageTwo} takes the \stageOneName predictions and neighboring radar points. It matches predicted \radar distribution with \radar points in the radial direction and computes binned and range-dependent matching scores. Peak positions indicates range estimations.}
		\label{fig:stg2}
	\end{figure}

	\subsection{\stageTwo: \stageTwoNameFull}\label{sec:matching}
	\stageTwo uses the predicted \radar density  from the \stageOneName model to score the consistency of object positions with the measured \radar hits.  By binning measured \radar returns, we can use a convolution of the \stageOneName to obtain similarity scores as a function of range, and so locate potential target positions.

	\cref{fig:stg2} shows the overview of \stageTwo.
	Both the predicted and measured \radar data are resampled into a \bev space with the X-axis along radial direction (from ego to target, approximately parallel to \radar rays), and the Y-axis along the tangential direction. 
	We align the central row of the \stageOneName map with the central row of the \radar measurement map and restrict the search to be along the X-axis ({\it i.e.}, the radial axis) near to the monocular detected center.  Our motivation is that monocular detections are more accurate tangentially ({\it i.e.}, image space) and less radially.
	To compute the similarity between the predicted and measured densities, we slide the \stageOneName  kernel along the radial axis and calculate by sum of dot product (convolution) between the \stageOneName and the actual point count at each position. This results in a \oneD matching score profile along the radial dimension, indicating potential target locations in the vicinity of the monocular centers. 
	
	Specifically, given a binned radar distribution $P$  with center $(L_\text{P},L_\text{P})$ and resolution of  $(2L_\text{P} + 1, 2L_\text{P} + 1)$ and radar binned count $C$ with center $(L_\text{C},L_\text{C})$, {\it i.e.}, object center from monocular prediction, and size of 
	$(2L_\text{C} + 1, 2L_\text{C} + 1)$, the row convolution (or cross-correlation) is   
	\begin{multline}
		S_\text{STG2}(n) = 
		\sum_{i=-L_\text{P}}^{L_\text{P}}\sum_{j=-L_\text{P}}^{L_\text{P}}P\left(L_\text{P}  + i + 1, L_\text{P} + j + 1\right)\\C\left(L_\text{C} + i + 1, L_\text{C} + j + n + 1\right),
		\label{eq:conv}
	\end{multline}
	where $n$ is offset to the object center along the row.
	
	To create the measured densities, we first accumulate multiple \radar sweeps over a short interval prior to the detection.  Object motions of \radar points are compensated by Doppler velocity and object velocity from monocular detections similar to \cref{eq:motion}. The difference is that object velocity, $\bm{v}_\text{O}$, is obtained from the monocular detector.

	\subsection{\stageThree: \stageThreeNameFull}
	Matching scores from Stage 2 provide position candidates, which have high scores. The purpose of \stageThree is to select the best range predictions among Stage-2 candidates by considering the combined evidence from monocular detection and \radar measurements. 
	\stageThree trains a neural network to rescore the candidate positions using additional evidence. 
	
	This model should consider multiple factors that indicate the confidence of each choice. For instance, high detection scores imply high confidence in monocular detection; large matching scores indicate more accurate matched positions from \stageTwo; monocular detector excels at low ranges while suffers at long ranges where Stage-2 candidates may be a better choice because of the help of \radar measurement.

	\noIndentHeading{Architecture and Loss.} The Stage-3 network takes two types of inputs: (a) monocular detection parameters (class, range, and size); (b) Stage-2 matching scores at binned ranges.
	The network preprocesses  inputs via separate linear layers, concatenates the results, feeds them to an MLP to extract features, and predicts
	a confidence score per range candidate. 
	Cross-entropy loss is used for training the network. Training data are generated from true positive monocular detections with non-zero Stage-2 matching scores. 
	
	\noIndentHeading{Inference.}  Denoting predicted Stage-3 scores as $S_\text{STG3}(i)$, where $i$ is quantized offset to monocular predicted range,
	we estimate the range offset by finding the peak position as
	\begin{equation}
		\Delta n = \text{argmax}_i\left(S_\text{STG3}(i)\right), 
	\end{equation}
	and corresponding 
	Stage-3 score $S_\text{STG3}(\Delta n)$. The final predicted range $R_\text{F}$ can be expressed as
	\begin{equation}
		R_\text{F} = R_\text{CAM} + \Delta n \, b_\text{p},
	\end{equation}
	where $R_\text{CAM}$ is monocular predicted range, and $b_\text{p}$ is bin size. Typically, the estimated range is approximately at one of the peak positions from Stage-2 score or at monocular estimated range; thus Stage 3 implicitly learns to select the best position candidates from previous stages.
	We also update detection score by combining monocular detection score $S_\text{CAM}$ and Stage-3 score as
	\begin{equation}
		S_\text{F} = S_\text{CAM} + \alpha S_\text{STG3}(\Delta n),
	\end{equation}
	where $\alpha$ is a Stage-3 weighting parameter. Note $S_\text{STG3}$ has been processed by the Softmax function, and its value ranges from 0 to 1.

	\section{Experiments}
	
	\noIndentHeading{Dataset.} Our experiments are based on the widely-used \nuscenes dataset \cite{caesar2020nuscenes}, with both images and \radar points collected in urban driving scenarios.
	Equipped with six cameras and five \radar{}s, the ego-vehicle scans traffic environments in $360$ degrees. There are $700$ training scenes, $150$ validation scenes, and $150$ test scenes, each with $10$ classes of objects specified with bounding boxes.
	
	\noIndentHeading{Data Splits.} We follow the standard splits of the \nuscenes detection benchmark: the test results are obtained from model trained on \nuscenes training plus validation set ($34$K frames) and evaluated on the test set ($6$K frames); the validation results trained on \nuscenes training set ($28$K frames) and evaluated on the validation set ($6$K frames). 
	
	\noIndentHeading{Implementation Details.} 
	Our code uses PyTorch~\cite{paszke2019pytorch} and detection package MMDetection3D~\cite{mmdet3d2020}. 
	Our experiments use \sparseBEV \cite{liu2023sparsebev} as the monocular detector. Note that \methodName is flexible and easily adaptable to other monocular methods. To preserve the premium detection performance of monocular component and focus on training the Stage-3 model, we use pretrained weights for the monocular branch, which are frozen in training and inference. 
	We train Stage-1 and Stage-3 models with RMSProp optimizer for $120$ epochs with an initial learning rate of $1\times10^{-6}$, which is reduced by half at the 60th epoch. 
	We list the number of parameters in \methodName Stage-1 and Stage-3 models as well as in underlying monocular models with two backbones in \cref{tab:model_size}.  \methodName Stage 1 and Stage 3 are relatively lightweight compared to monocular models.
	
	\begin{table*}[!tb]
		\caption{\textbf{Detection Performance on \nuscenes Test Set}. \methodName achieves \sota performance for camera-radar fusion. 
		}
		\label{tab:nusc_test_detection}
		\centering
		\scalebox{0.88}{
			\rowcolors{3}{white}{lightgray}
			\setlength\tabcolsep{0.13cm}
			\begin{tabular}{m cc m c m c m c | c | c | c | c | c m}
				\myTopRule
				\multicolumn{2}{mcm}{Modality} 
				& \multirow{2}{*}{Method}
				& \multirow{2}{*}{NDS ($\uparrow$)}
				& \multirow{2}{*}{mAP ($\uparrow$)} 
				& \multirow{2}{*}{mATE ($\downarrow$)}
				& \multirow{2}{*}{mASE ($\downarrow$)}
				& \multirow{2}{*}{mAOE ($\downarrow$)}
				& \multirow{2}{*}{mAVE ($\downarrow$)}
				& \multirow{2}{*}{mAAE ($\downarrow$)}\\
				\Radar & Camera & & & & & & & & \\ 
				\myTopRule
				& \checkmark & \pgd~\cite{wang2021probabilistic} 
				& $0.448$ & $0.386$  & $0.626$ & $0.245$ & $0.451$ & $1.509$ & $0.127$ \\
				& \checkmark & \sparseBEV~\cite{liu2023sparsebev}
				& $0.675$ & $0.603$  & $0.425$ & $0.239$ & $0.311$ & $0.172$ & $0.116$ \\
				\hline\hline
				\checkmark & \checkmark & MVFusion~\cite{wu2023mvfusion}
				& $0.517$ & $0.453$  & $0.569$ & $0.246$ & $0.379$ & $0.781$ & $0.128$ \\
				\checkmark & \checkmark & CRN~\cite{kim2023crn} 
				& $0.624$ & $0.575$  & $0.416$ & $0.264$ & $0.456$ & $0.365$ & $0.130$ \\			
				\checkmark & \checkmark & RCBEVDet~\cite{lin2024rcbevdet} 
				& $0.639$ & $0.550$  & $0.390$ & \best{0.234} & $0.362$ & $0.259$ & \best{0.113} \\
				\checkmark & \checkmark & HyDRa~\cite{wolters2024unleashing} 
				& $0.642$ & $0.574$  & $0.398$ & $0.251$ & $0.423$ & $0.249$ & $0.122$ \\
				\checkmark & \checkmark & HVDetFusion~\cite{lei2023hvdetfusion} 
				& $0.674$ & $0.609$  & $0.379$ & $0.243$ & $0.382$ & $0.172$ & $0.132$ \\
				\checkmark & \checkmark & \sparseBEV{} + \textbf{\methodName{}}
				& \best{0.695} & \best{0.630}  & \best{0.363} & $0.240$ & \best{0.311} & \best{0.167} & $0.118$ \\			
				\myTopRule
			\end{tabular}
		} 
	\end{table*}
	
	\begin{table*}[!t]
		\caption{\textbf{\nuscenes Validation Results}.} 
		\label{tab:nusc_val_detection}
		\centering
		\scalebox{0.88}{
			\rowcolors{3}{white}{lightgray}
			\setlength\tabcolsep{0.13cm}
			\begin{tabular}{m cc m c m c m c | c | c | c | c | c m}
				\myTopRule
				\multicolumn{2}{mcm}{Modality} 
				& \multirow{2}{*}{Method}
				& \multirow{2}{*}{NDS ($\uparrow$)}
				& \multirow{2}{*}{mAP ($\uparrow$)} 
				& \multirow{2}{*}{mATE ($\downarrow$)}
				& \multirow{2}{*}{mASE ($\downarrow$)}
				& \multirow{2}{*}{mAOE ($\downarrow$)}
				& \multirow{2}{*}{mAVE ($\downarrow$)}
				& \multirow{2}{*}{mAAE ($\downarrow$)}\\
				\Radar & Camera & & & & & & & & \\ 
				\myTopRule
				& \checkmark & \pgd~\cite{wang2021probabilistic} 
				& $0.428$ & $0.369$  & $0.683$ & $0.260$ & $0.439$ & $1.268$ & $0.185$ \\
				& \checkmark & \sparseBEV~\cite{liu2023sparsebev}
				& $0.592$ & $0.501$  & $0.562$ & $0.265$ & $0.320$ & $0.243$ & $0.195$ \\
				\hline\hline
				\checkmark & \checkmark & MVFusion~\cite{wu2023mvfusion}
				& $0.455$ & $0.380$  & $0.675$ & \best{0.258} & $0.372$ & $0.833$ & $0.196$ \\
				\checkmark & \checkmark & CRN~\cite{kim2023crn} 
				& $0.607$ & \best{0.545}  & $0.445$ & $0.268$ & $0.425$ & $0.332$ & \best{0.180} \\			
				\checkmark & \checkmark & RCBEVDet~\cite{lin2024rcbevdet} 
				& $0.568$ & $0.453$  & $0.486$ & $0.285$ & $0.404$ & $0.220$ & $0.192$ \\
				\checkmark & \checkmark & HyDRa~\cite{wolters2024unleashing}
				& $0.617$ & $0.536$  & \best{0.416} & $0.264$ & $0.407$ & $0.231$ & $0.186$ \\
				\checkmark & \checkmark & HVDetFusion~\cite{lei2023hvdetfusion} 
				& $0.557$ & $0.451$  & $0.527$ & $0.270$ & $0.473$ & \best{0.212} & $0.204$ \\
				\checkmark & \checkmark & \sparseBEV{} + \textbf{\methodName{}}
				& \best{0.622} & $0.544$  & $0.481$ & $0.266$ & \best{0.325} & $0.237$ & $0.189$ \\			
				\myTopRule
			\end{tabular}
		} 
	\end{table*}
	
	\textit{\stageOne:} We implement the Stage-1 network with a lightweight MLP-like network. An input sample consists of a series of vectors representing different properties of object, {\it e.g.}, size, orientation, and range. They are first processed by separate linear projection layers before being concatenated and fed into an MLP of $3$ hidden layers. The network output ({\it i.e.}, the binned distribution map) is defined in object local coordinates with X-axis parallel to object length, Y-axis to width, and map center at object center. It has a resolution of $129\times129$ with a pixel size of $0.1\times0.1$ meters for small and medium-sized object categories and pixel size of $0.2\times0.2$ meters for large-sized categories such as buses and trailers. 
	GT distribution is generated by accumulating $13$ neighboring \radar sweeps ($6$ previous sweeps, $1$ current, and $6$ future ones). We assume \radar points distribute within the GT bounding boxes on BEV, and thus points outside the bounding boxes are ignored and not used for training. We train Stage-1 and Stage-3 models separately since Stage-1 model is invariant to its underlying monocular model while Stage-3 network depends on the monocular model. 
	
	\textit{\stageTwo:} To perform Stage-2 convolution along the radial direction, the measured \radar positions are binned in radial-tangential coordinates to generate a \radar measurement map, with X-axis parallel to the ray from ego vehicle to target center and map center at target center detected by a monocular method. It has a resolution of $193\times193$ with $2$ pixel sizes mentioned above. Before performing convolution, the predicted \radar distribution map of \stageOne is rotated from object local coordinates to radial-tangential coordinates. The search range of convolution is $-3.2$m to $3.2$m relative to the range of target center estimated by the monocular method.
	
	\textit{\stageThree:} We implement the Stage-3 network via a lightweight MLP similar to \stageOne. To create labels for GT range, we associate monocular detections with GT bounding boxes under the conditions that the GT bounding boxes fall in the search range of associated monocular detections (with radial distance $\le3.2$m) and on the ray from ego to monocular detections (with tangential distance $\le \text{min}(0.5\text{m}, L)$, where $L$ is the object length).

	\begin{table}[!tb]
		\caption{Number of parameters in SparseBEV with different backbones as well as in \methodName Stage 1 and Stage 3.}
		\label{tab:model_size}
		\centering
		\scalebox{0.8}{
			\rowcolors{2}{lightgray}{white}
			\begin{tabular}{| c | c | c | c | c |}
				\myTopRule
				Models & 
				\begin{tabular}{@{}c@{}}SparseBEV \\ (V2-99)\end{tabular}  & 
				\begin{tabular}{@{}c@{}}SparseBEV \\(ResNet101)\end{tabular} & 
				Stage 1 & Stage 3 \\
				\myTopRule
				Params (M)  &  $94.0$  &  $63.6$ &  $19.2$   &  $0.3$  \\
				\myTopRule
		\end{tabular}}
	\end{table}

	\subsection{Quantitative Results on \nuscenes}
	\cref{tab:nusc_test_detection,tab:nusc_val_detection} show the performance of \methodName on test and validation set, respectively. The proposed fusion of \radar points with monocular detection proposals improves the performance of position estimation, which is evaluated with mAP and mean Average Translation Error (mATE) for true positive detections. 
	We compare  performance of \sota methods of monocular and \radar-camera fusion. 
	Note that for a fair comparison, the monocular component in \methodName, \textit{i.e.}, SparseBEV, uses exactly the same weights within \cref{tab:nusc_test_detection} and likewise uses the same weights within \cref{tab:nusc_val_detection}. Specifically, SparseBEV uses V2-99~\cite{lee2020centermask} and ResNet101~\cite{he2016deep} as its backbones in \cref{tab:nusc_test_detection,tab:nusc_val_detection}, respectively. 
	
	By comparing \methodName with its monocular counterpart (\thatIs{}, 8th row vs.~2nd row in \cref{tab:nusc_test_detection,tab:nusc_val_detection}), we see a significant improvement in mAP and a reduction in mATE when using \methodName with \radar data as inputs. Meanwhile it preserves good performance of its monocular component in other aspects, \textit{e.g.}, size and orientation estimation. 
	
	\methodName also achieves \sota performance among published \radar-camera fusion methods: in test set performance shown in \cref{tab:nusc_test_detection} and validation set results shown in \cref{tab:nusc_val_detection}, \methodName achieves the best overall performance measured by NDS and comparable performance in other metrics. As a detection-level fusion, final performance of \methodName depends on the quality of its underlying monocular model. We observe a stable and significant improvement in overall performance over its monocular models in both \cref{tab:nusc_test_detection,tab:nusc_val_detection}, although the monocular components adopt different backbones. The ease of plugging in different monocular components in our fusion architecture allows \methodName to capitalize on \sota monocular models and achieve better fusion performance.

	\begin{figure*}[t]
		\centering
		\scalebox{1.0}
		{
			\begin{tabular}{@{}c@{}c@{}c@{}c@{}c@{}}
				\\
				{\small (a)} &
				\includegraphics[height=1.2in]{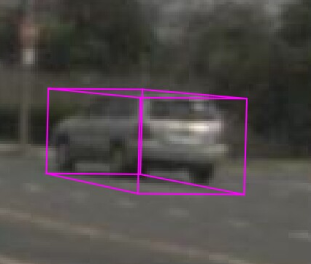}&
				\includegraphics[height=1.2in]{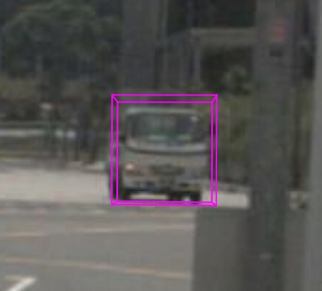}&	
				\includegraphics[height=1.2in]{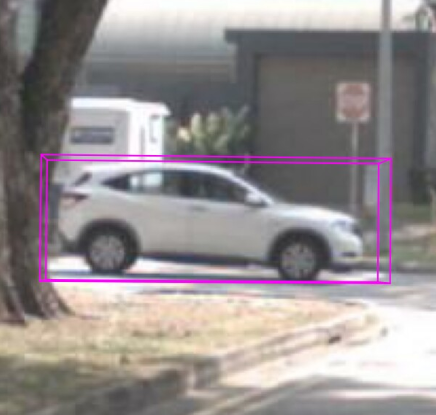}&
				\includegraphics[height=1.2in]{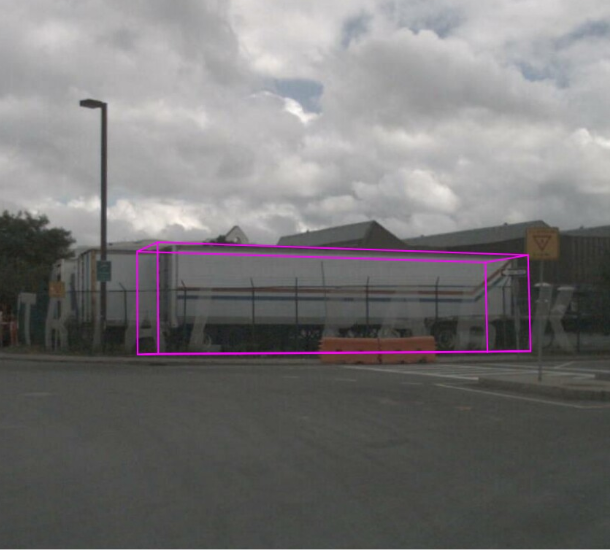}
				\\
				{\small (b)} &
				\includegraphics[height=1.4in]{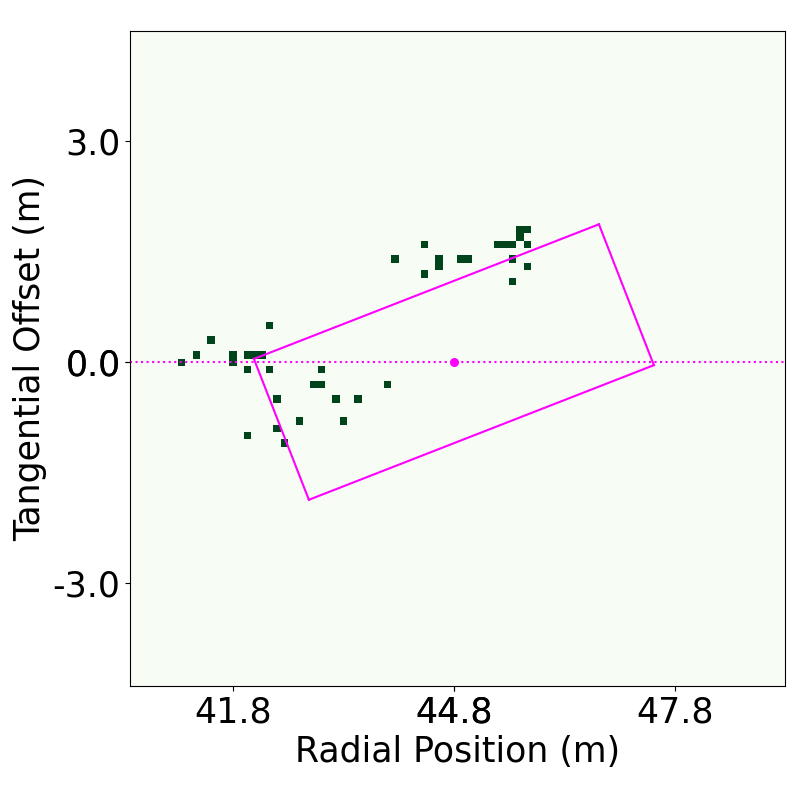}&
				\includegraphics[height=1.4in]{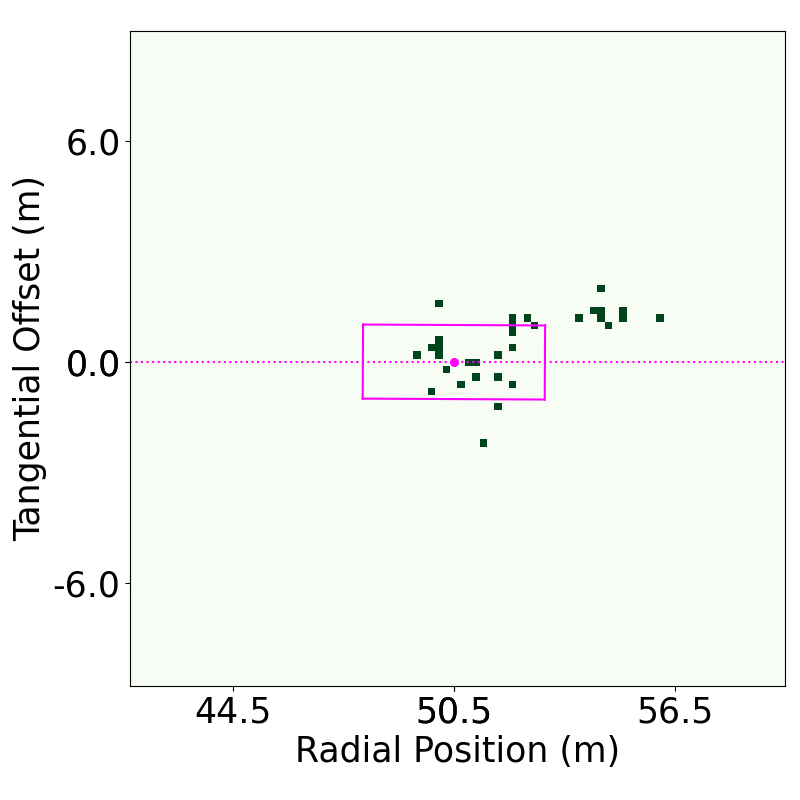}&	
				\includegraphics[height=1.4in]{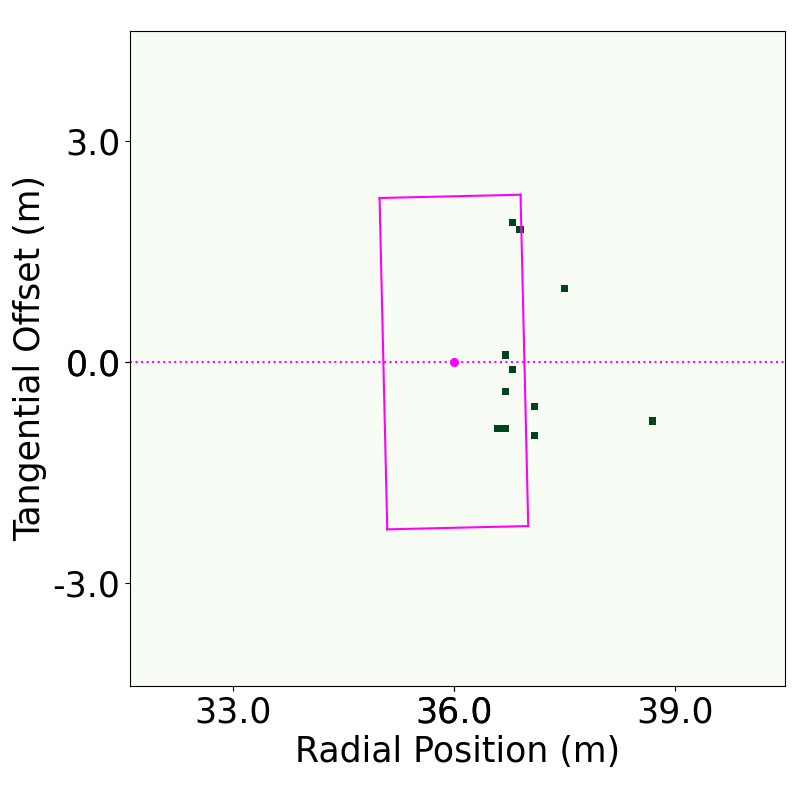}&
				\includegraphics[height=1.4in]{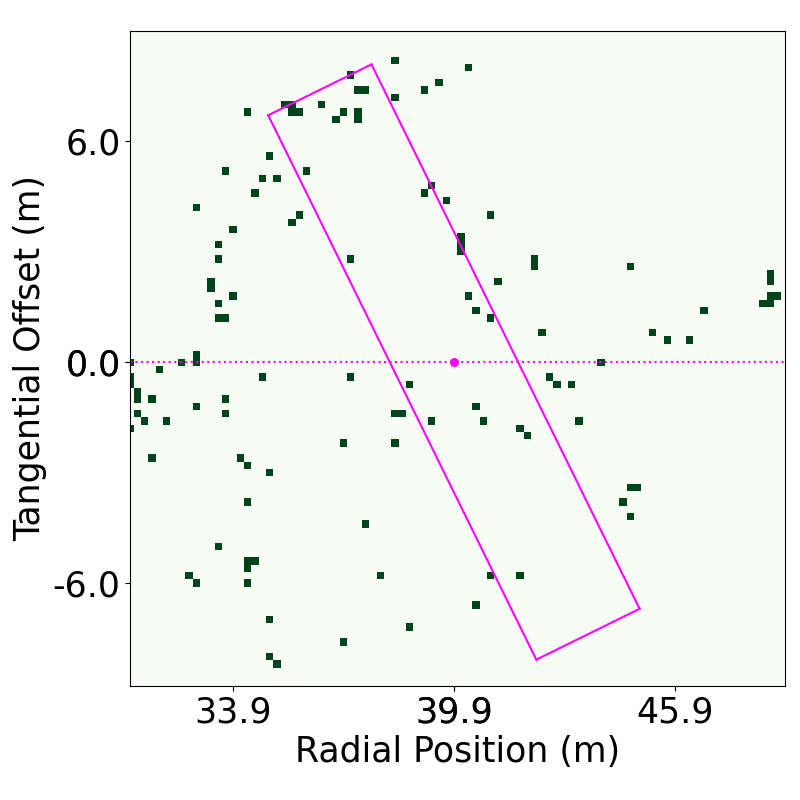}
				\\
				{\small (c)} &
				\includegraphics[height=1.3in]{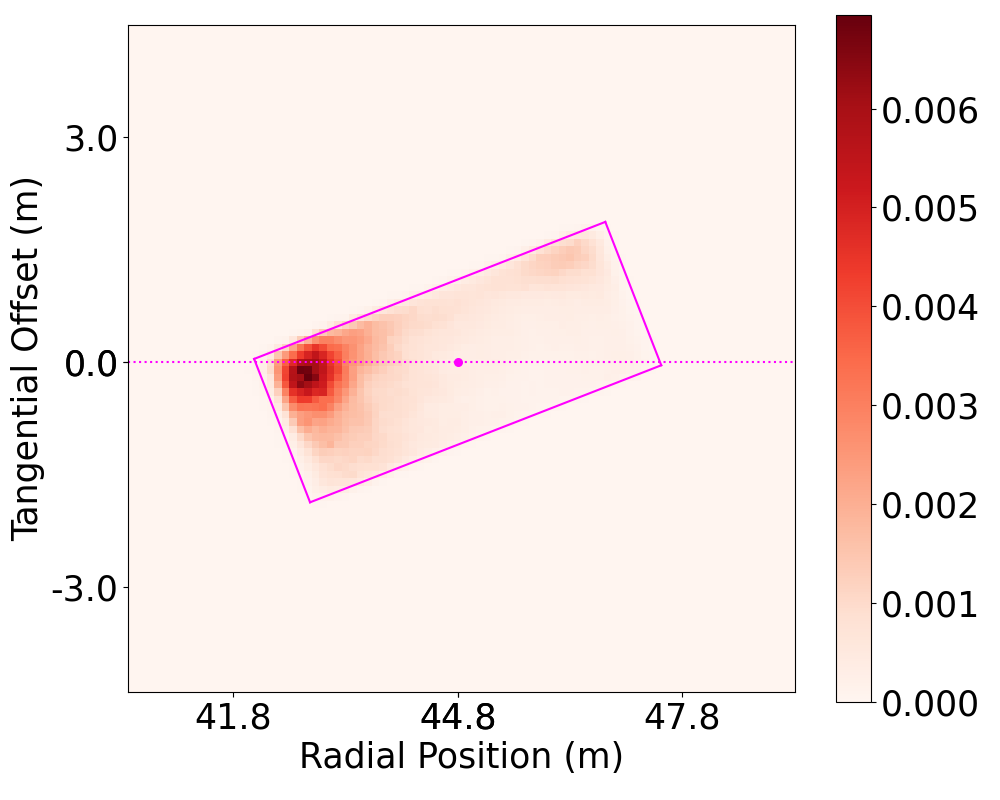}&
				\includegraphics[height=1.3in]{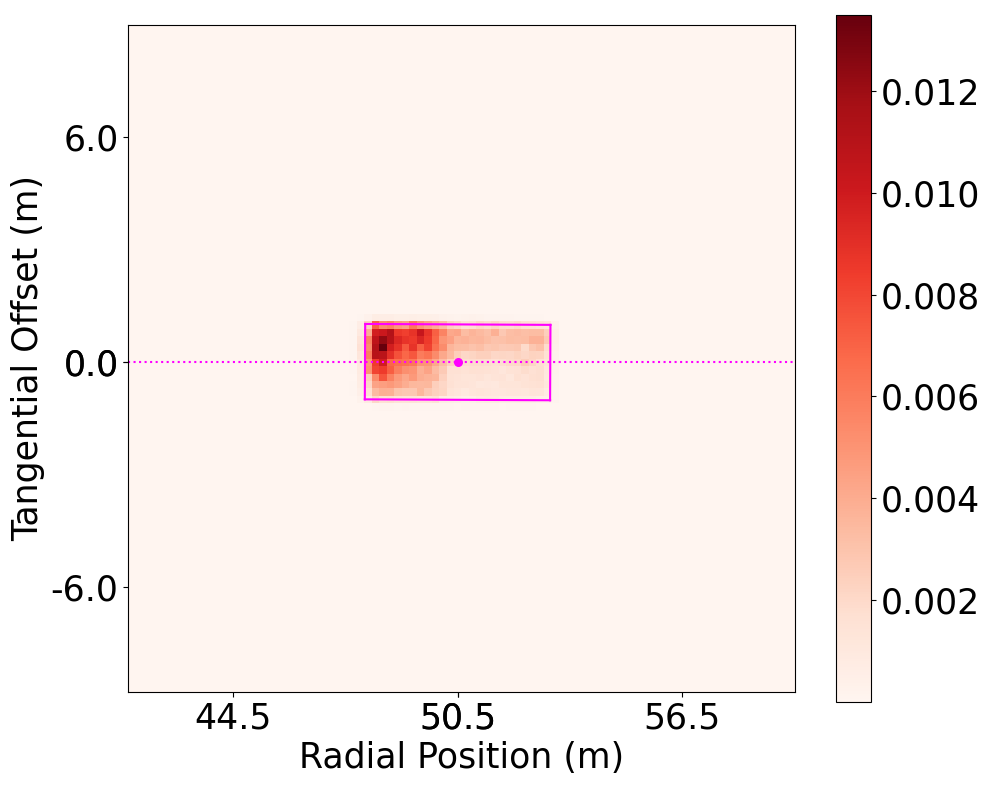}&	
				\includegraphics[height=1.3in]{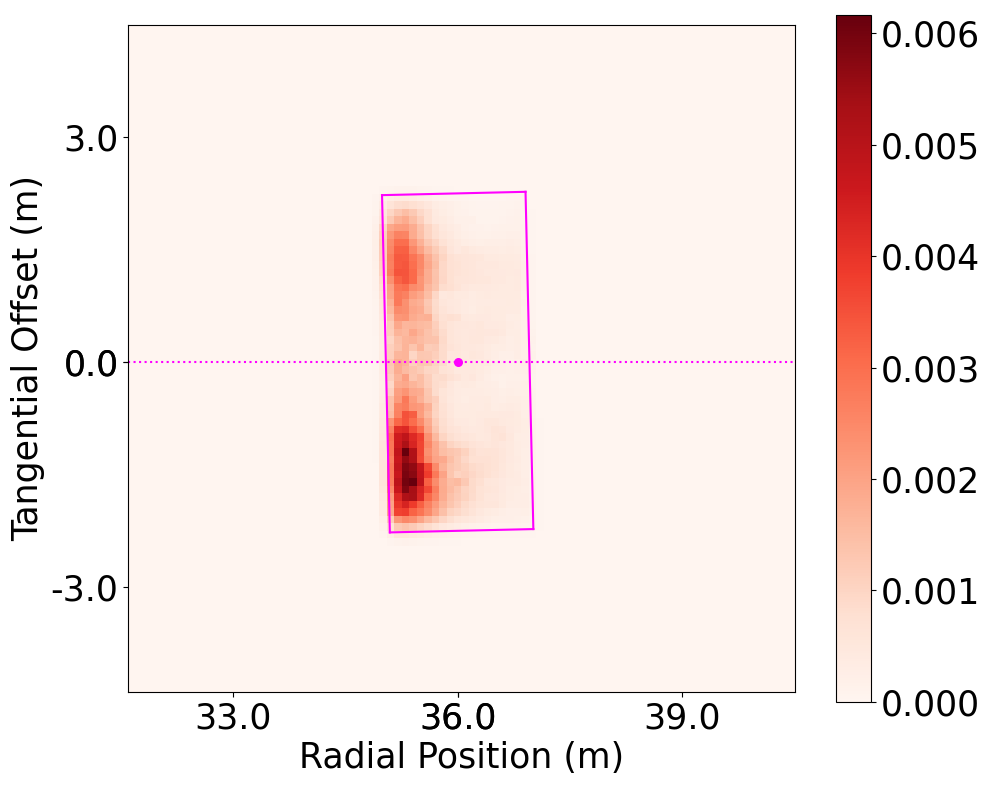}&
				\includegraphics[height=1.3in]{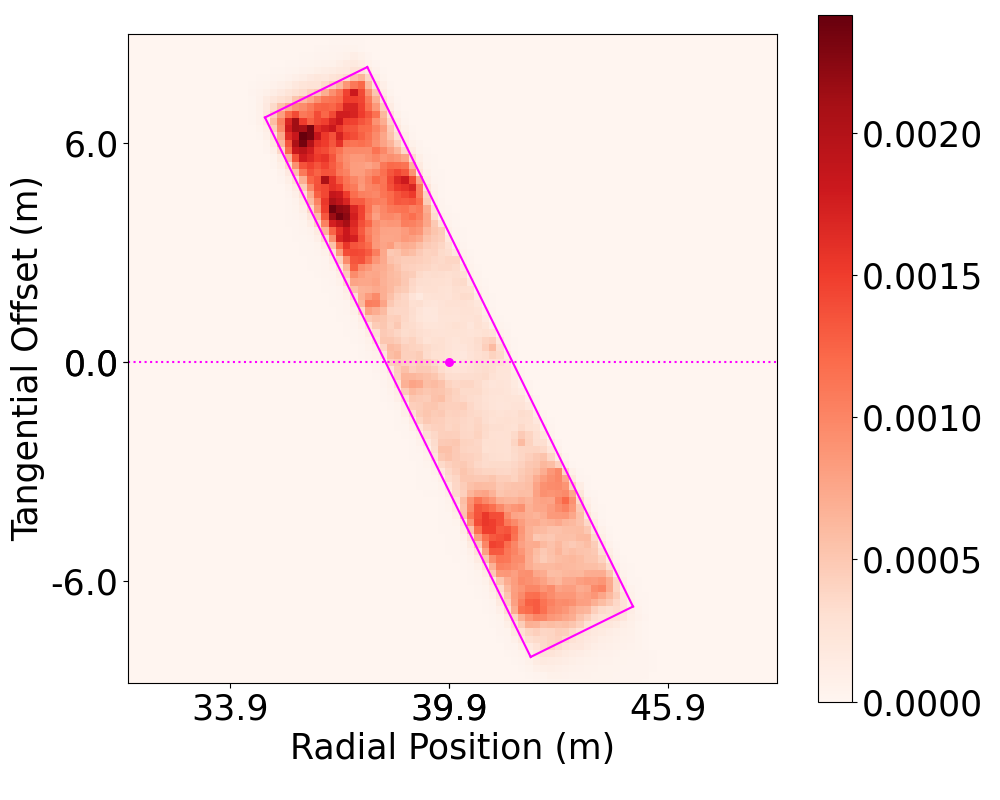}
				\\
				{\small (d)} &
				\includegraphics[height=0.8in]{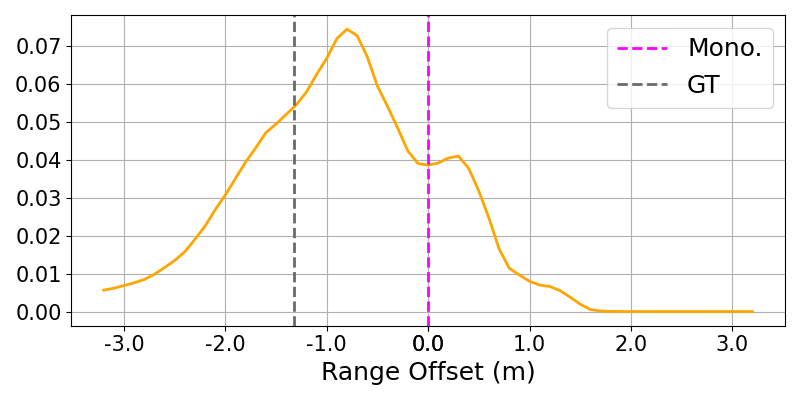}&
				\includegraphics[height=0.8in]{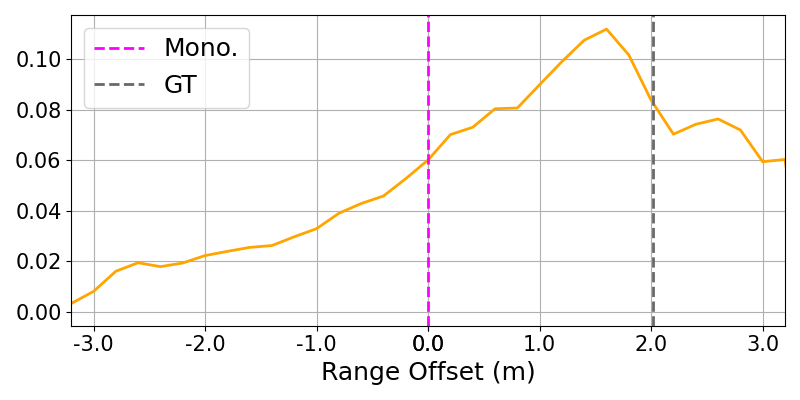}&	
				\includegraphics[height=0.8in]{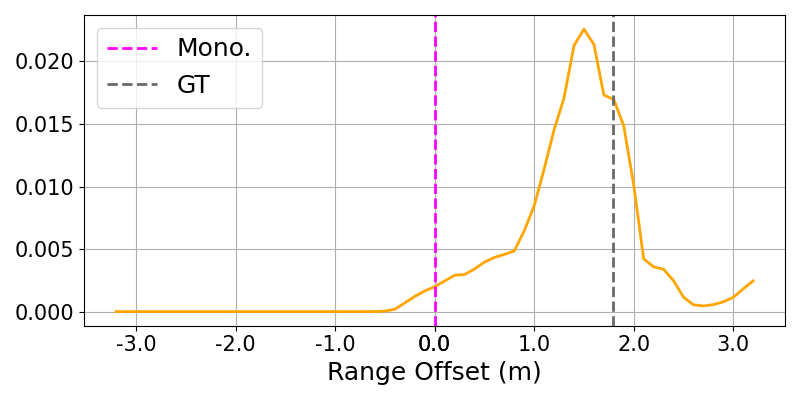}&
				\includegraphics[height=0.8in]{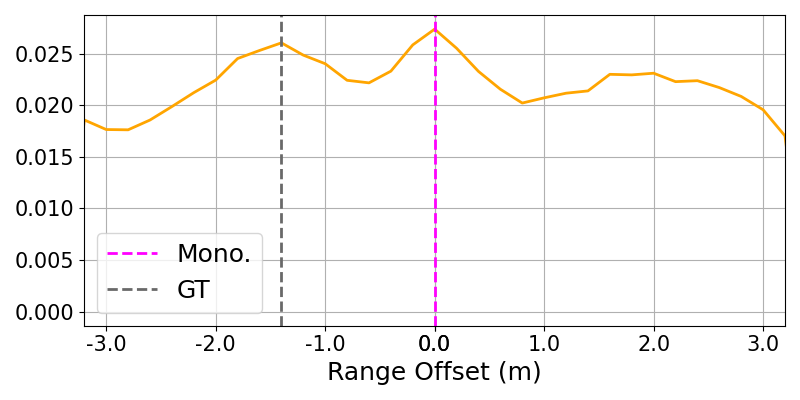}
				\\
				{\small (e)} &
				\includegraphics[height=0.8in]{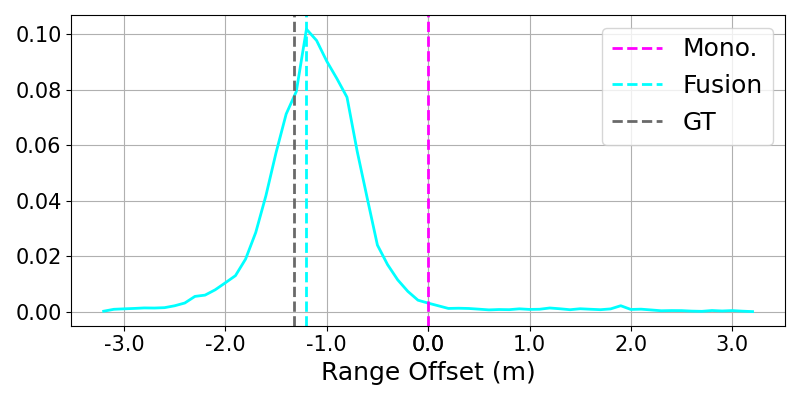}&
				\includegraphics[height=0.8in]{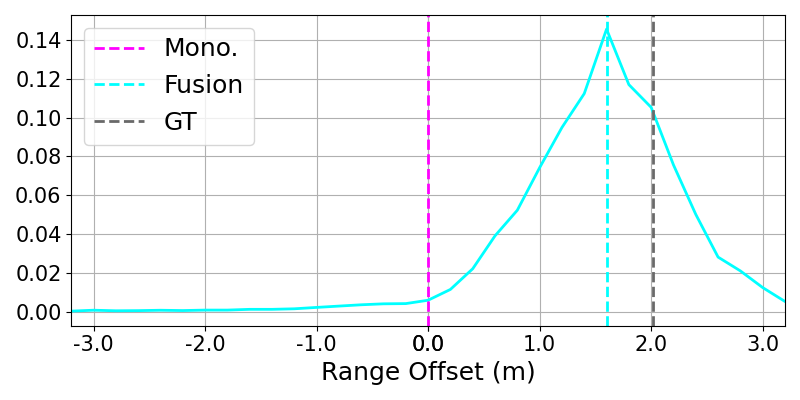}&	
				\includegraphics[height=0.8in]{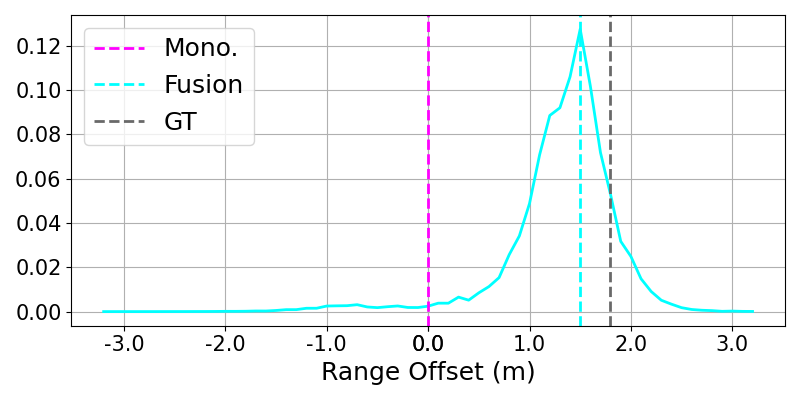}&
				\includegraphics[height=0.8in]{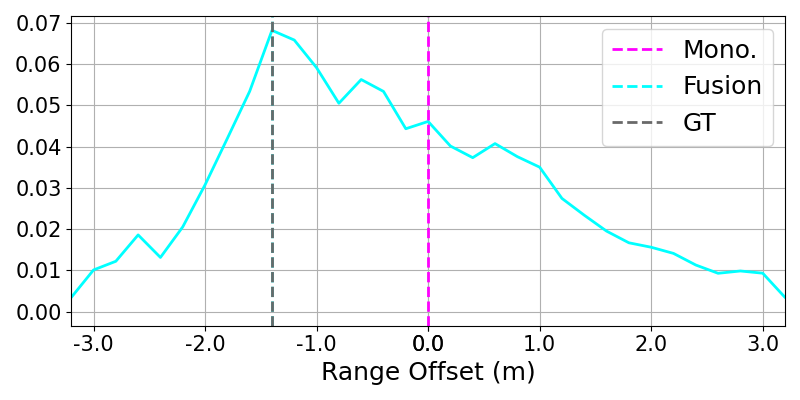}
				\\
				{\small (f)} &
				\includegraphics[height=1.2in]{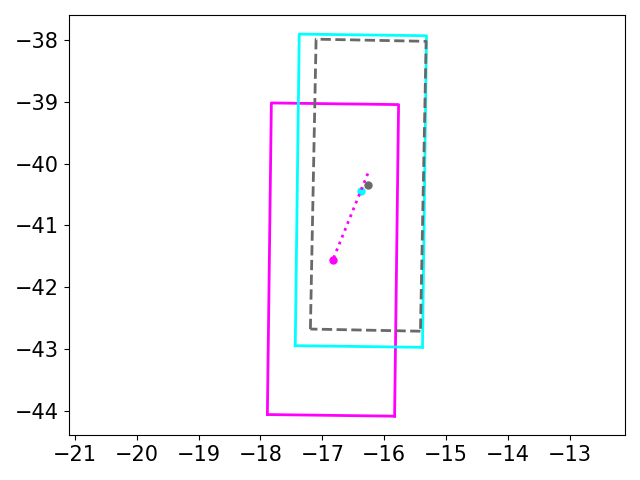}&
				\includegraphics[height=1.2in]{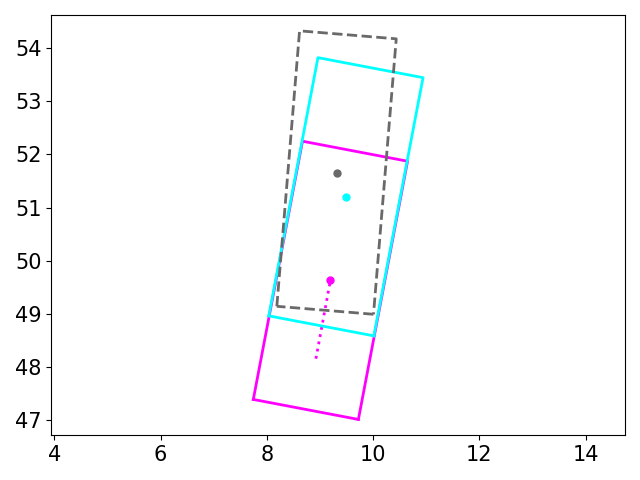}&	
				\includegraphics[height=1.2in]{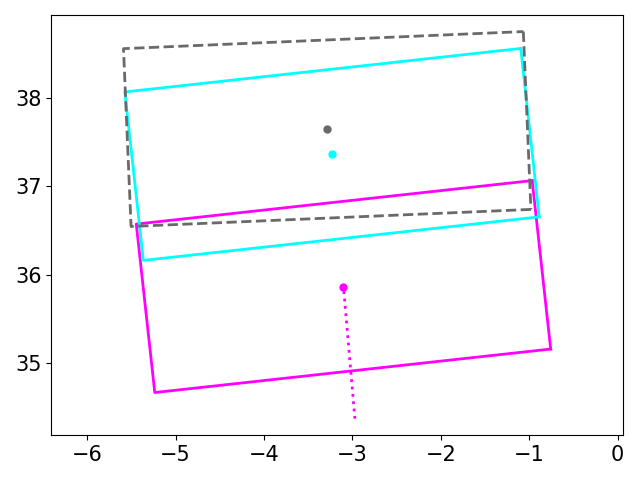}&
				\includegraphics[height=1.2in]{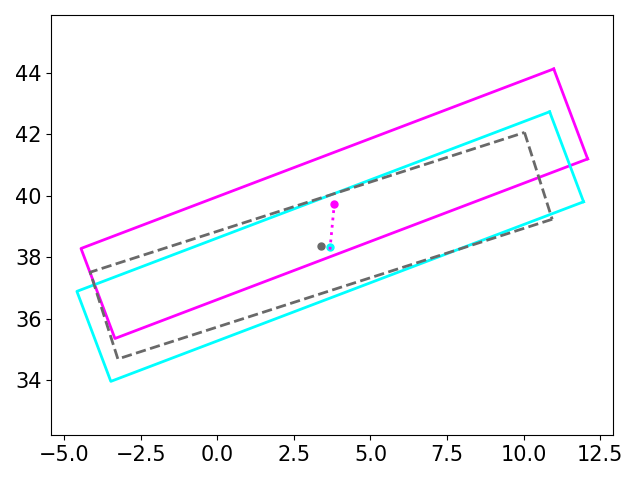}
			\end{tabular}
		}
		\vspace{-2mm}
		\caption{\textbf{Qualitative Results.} \small Visualizations of (a) objects in images, (b) binned \radar points, (c) predicted \radar hits distribution, (d) Stage-2 matching scores, (e) predicted Stage-3 scores, and (f) detections in ego BEV coordinates. [\textbf{Key}: \textcolor{magenta}{Monocular detections}, \textcolor{cyan}{\methodName{} detections}, \textcolor{gray}{GT}. GT boxes are dashed.]
			\vspace{-2mm}
		}
		\label{fig:exp_vis}
	\end{figure*} 
	
	\subsection{Evaluating Stages 2 and 3}
	To evaluate the Stages 2 and 3 in object range estimation, we compare the range error from monocular detection, Stage-2, and Stage-3 estimations. Stage-1 model is trained with \nuscenes training set 
	and Stage-3 model is trained with detections by the monocular method and corresponding Stage-2 matching scores from \nuscenes training set.

	\begin{table*}[t!]
		\caption{Comparison of range estimation accuracy of monocular, Stage-2, and Stage-3 estimates. We use mean/median of absolute range error (meter) as metrics. [\textbf{Key}: Ped.= Pedestrian, Motor.= Motorcycle, CV= Construction Vehicle, TC= Traffic Cone.]}
		\label{tab:stg23}
		\centering
		\vspace{-1mm}
		\scalebox{0.8}{
			\rowcolors{2}{lightgray}{white}
			\setlength\tabcolsep{0.1cm}
			\begin{tabular}{ | c |  c | c | c | c | c | c | c | c | c | c | c |}
				\myTopRule
				Method  & Class-Mean  & Car & Truck & Bus & Trailer & CV & Ped. & Motor. & Bicycle & TC & Barrier\\
				\myTopRule
				Monocular    &   $0.83/0.57$   &  $0.65/0.38$  &  $0.86/0.56$ &  $1.04/0.80$  &  $1.23/1.08$  &  $1.13/0.88$ &  $0.85/0.55$ & $0.81/0.50$  &  $0.65/0.40$  &  $0.51/0.24$  &  $0.59/0.31$  \\
				\stageTwo    &   $0.94/0.52$   &  $0.50/0.21$  &  $0.75/0.39$ &  $0.77/$\best{0.46}  &  $1.53/1.05$  &  $1.11/0.75$ &  $1.13/0.56$ & $0.82/0.35$  &  $0.86/0.44$  &  $0.93/0.41$  &  $1.01/0.55$   \\
				\stageThree  &   \best{0.65/0.36}   &  \best{0.38/0.18}  &  \best{0.59/0.34} &  \best{0.70/0.46}  &  \best{1.13/0.76}  &  \best{0.87/0.63} &  \best{0.71/0.28} & \best{0.57/0.29} &  \best{0.54/0.24}  &  \best{0.46/0.17}  &  \best{0.57/0.27}   \\
				\myTopRule
		\end{tabular}}
	\end{table*}

	\begin{table}[t!]
		\caption{Comparison of \methodName using two baseline \radar hit distributions for \stageOne versus RIC. The metric is mean absolute error (MAE) of range in meters and mean matching score (MMS) between the distributions and actual measurements.}
		\label{tab:distr}
		\centering
		\vspace{-1mm}
		\scalebox{1.0}{
			\footnotesize
			\begin{tabular}{| c | c | c |}
				\hline
				Distribution & MAE $(\downarrow)$ & MMS $(\uparrow)$ \\
				\hline
				L-Shaped  	& 0.77 & 0.059 \\
				Uniform     &  0.67 & 0.078  \\
				\textbf{RIC} & \textbf{0.47}  & \textbf{0.105} \\
				\hline
			\end{tabular}
		}
	\end{table}
	
	To generate Stage-2 estimation, the trained Stage-1 model is applied to monocular detections (\sparseBEV) in \nuscenes validation set to generate \radar distributions, which are subsequently convolved with \radar measurements (in \stageTwo) to obtain matching scores at binned ranges along the ray. 
	The range with the maximum matching score is used as Stage-2 estimation for this evaluation. 
	Finally, we estimate Stage-3 output by feeding the Stage-2 outputs and monocular detections to the Stage-3 model. 
	From \cref{tab:stg23}, we can see that simple extraction from \stageTwo can improve over monocular methods with lower median error but suffers from outliers with larger mean error. 
	\stageThree achieves the best range estimation accuracy across all $10$ categories, by fusing the monocular estimation and Stage-2 outputs.
	
	\subsection{Qualitative Results}
	In Fig.~\ref{fig:exp_vis}, we show examples of \methodName being applied to monocular detections. 
	The radar BEV map in (b) and predicted radar distribution in (c) are both centered at the monocular object center and with X-axis being radial direction and Y-axis tangential direction.  
	We can observe the complexity of predicted radar distributions in (c), which vary according to object size and orientation relative to radar rays. The edges of objects facing the radar tend to have higher densities, and large vehicles have a wider spread (see the 4th column), with reflections by parts under the vehicle near the tires. 
	
	Fig.~\ref{fig:exp_vis}(d) shows Stage-2 matching scores as a function of radial offset (X-axis), with monocular estimated range (magenta) and GT range (gray). Multiple peaks in the 4th column indicate ambiguities in matching radar hits. (e) shows the predicted Stage-3 scores in blue, which typically have a sharper peak than Stage 2, illustrating Stage-3 candidate refinements. In the fourth column Stage 3 resolved ambiguity by enhancing one peak.  Row (f) shows that \methodName improves monocular method in radial position prediction as \textcolor{cyan}{predicted bounding boxes} are closer to (dashed) \textcolor{gray}{GT} compared to \textcolor{magenta}{monocular detections}. The radial directions are plotted as dotted lines for reference.

	\subsection{Ablation on Distribution Models}
	\label{sect:exp_stg1}
	To assess the benefit of using our trained \stageOneName model to represent radar hits on objects, we compare it with using two baseline distributions, {\it i.e.}, a uniform radar distribution within object boundaries and an ``L-shaped" distribution on reflecting sides of bounding boxes. The ``L-shaped" distribution is a simulation of LiDAR point distribution. Each model is passed through \stageTwo, and we estimate object range at the maximum matching score, and compute their range errors. We also use matching score ({\it i.e.},~dot product of predicted distribution and pixel-wise radar point counts from accumulated measurement) as an additional metric for evaluating distribution accuracy. We compute mean matching score (MMS) over all classes. We train \stageOneName with \radar measurements and GT bounding boxes in \nuscenes training set and evaluate on \nuscenes validation set. Note that, in this experiment, GT bounding boxes are used to generate the \radar distribution, and no monocular boxes are involved. \cref{tab:distr} shows that, with the smallest range estimation error and the largest matching score, the \stageOneName distribution captures more accurately the real \radar distribution compared with the two baseline distributions.

	\section{Conclusions}
	This paper presents a novel \radar-camera fusion strategy that utilizes BEV radar distributions to improve object range estimation over monocular methods. Evaluation on the nuScenes dataset shows that \methodName realizes stable and significant improvements in object position estimation over its underlying monocular detector, and achieves the \sota performance in radar-camera-based 3D object detection. We believe this effective method, that is simple to implement, will broadly benefit existing and future camera-radar fusion methods.
	
	\noIndentHeading{Limitations.}
	First, as a detection level fusion, \methodName only uses high-level monocular detection parameters and does not directly utilize low-level image features. Information loss is inevitable in this low-level to high-level feature transition ({\it e.g.}, false negative detections), and it is difficult to make use of radar points for further improvement if an object is missed by the monocular component in the first place.  Second, \methodName adopts BEV representation for radar points distribution. However, BEV representation has an intrinsic limitation to represent radar hits from two vertically positioned targets at the same BEV location ({\it e.g.},  a person riding on a bicycle), since their reflected radar hits are mapped to the same BEV pixel.
	
	\noIndentHeading{Future Work.}
	\methodName assumes the underlying monocular detector has accurate estimation in tangential position, size and orientation and focuses on improving range estimation, and also assumes the distribution remain fixed in the search space. To achieve more precise radar distribution matching, it is worthwhile in future work expanding the search space from a 1D ({\it i.e.},  radial offset) to a high-dimensional space ({\it e.g.}, radial and tangential offsets, size, and orientation) and adopting a variable distribution as a function of locations in the search space during matching.

	\clearpage

	{
		\small
		\bibliographystyle{ieeenat_fullname}
		\bibliography{references}
	}
	
	\clearpage

	\maketitlesupplementary

	\setcounter{figure}{0}
	\setcounter{table}{0}
	\setcounter{section}{0}
	
	\section{Additional Implementation Details}
	\subsection{Detailed Network Structure}
\begin{figure}[t]
		\centering
		\includegraphics[width=\linewidth]{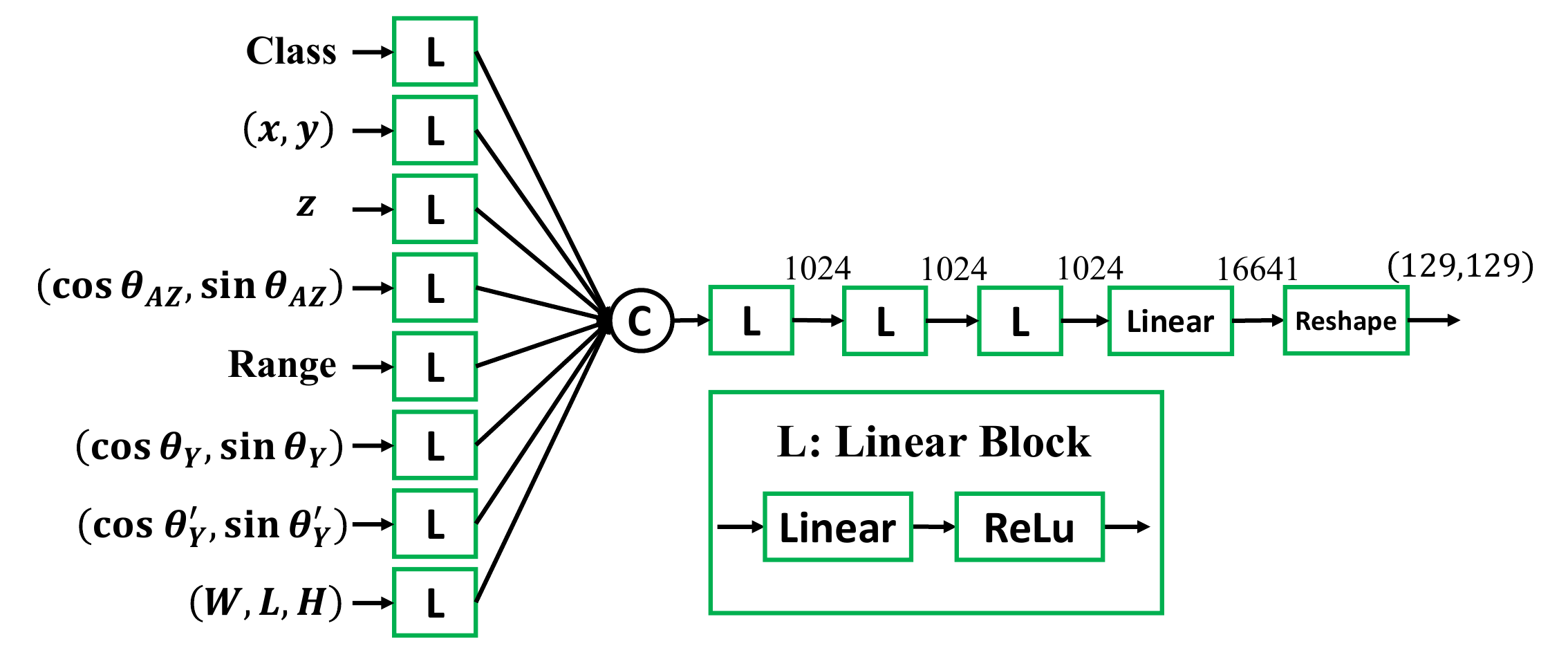}
		\caption{Stage-1 Network Structure. The class input is in one-hot encoding; $z$ represents heights of bounding box bottom faces; $\theta_\text{AZ}$ stands for azimuths of objects in ego coordinates; $\theta_\text{Y}$ and $\theta_\text{Y}^\prime$ are object yaws in ego coordinates and relative yaws ({\it i.e.}, $\theta_\text{Y} - \theta_\text{AZ}$), respectively. ``C" represents concatenation, and ``Linear" denotes a linear transformation layer. Feature sizes are marked besides network layers.}
		\label{fig:stg1_net}
	\end{figure}

\begin{figure}[t]
		\centering
		\includegraphics[width=\linewidth]{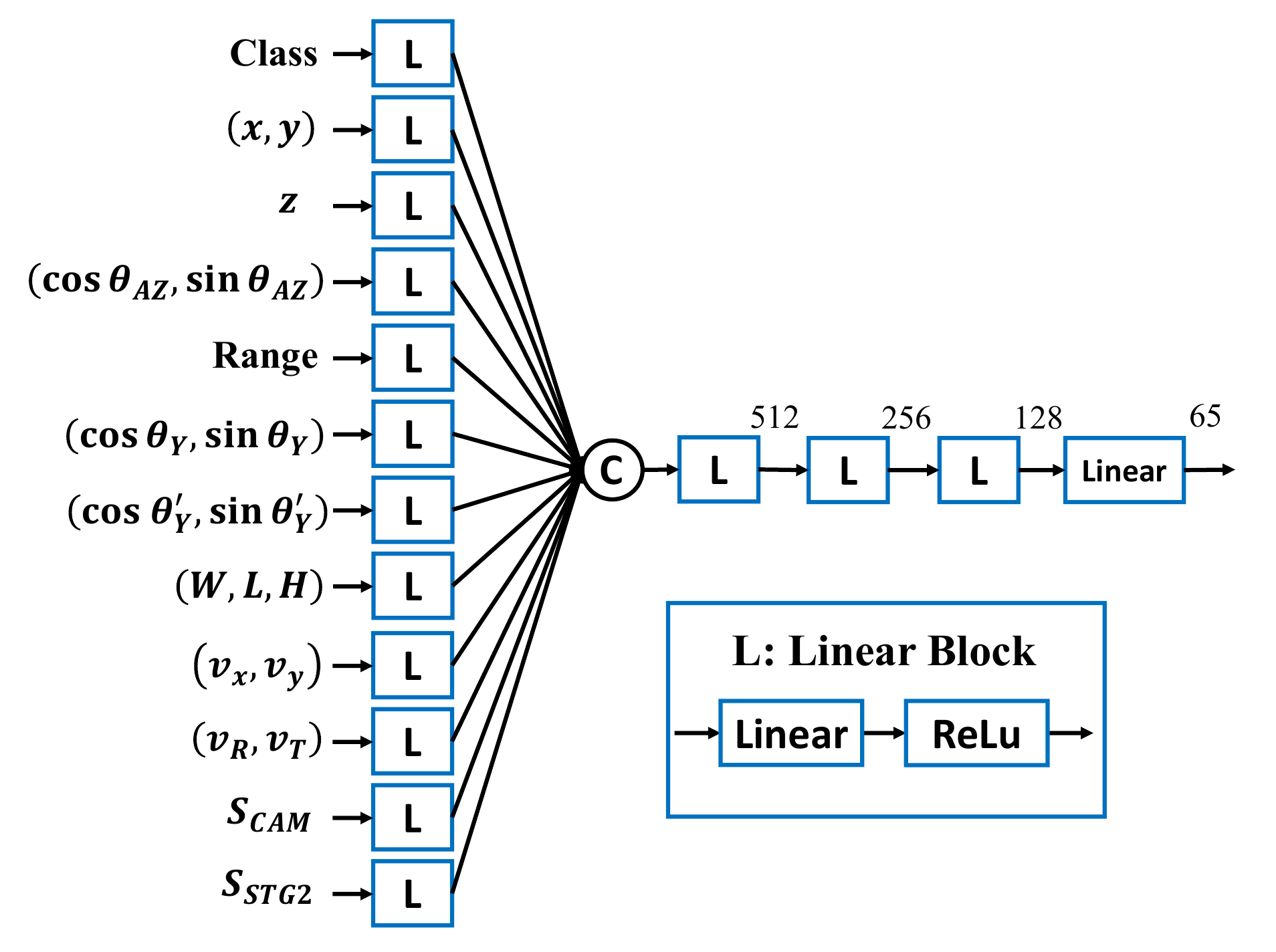}
		\caption{Stage-3 Network Structure. The inputs $v_\text{x}$ and $v_\text{y}$ are monocular estimated object velocities in ego coordinates; $v_\text{R}$ and $v_\text{T}$ are monocular velocities in radial and tangential directions, respectively; $S_\text{CAM}$ and $S_\text{STG2}$ represent monocular detection scores and Stage-2 matching scores, respectively.}
		\label{fig:stg3_net}
	\end{figure}

	Figs.~\ref{fig:stg1_net} and~\ref{fig:stg3_net} show network structures of Stages 1 and 3, respectively.
	
	\subsection{Inference Time}
	Using a NVIDIA V100s GPU and Intel Xeon Platinum 8260 CPUs, we record in Tab.~\ref{tab:time} inference time for different components of \methodName. Radar processing refers to accumulation and BEV binning of $7$ radar sweeps. We can see the monocular component takes most of the inference time and in comparison Stages 1 to 3 are very fast. Radar processing has not been optimized and could be sped up through code optimization.
	
	\begin{table}[t]
		\caption{Inference time for SparseBEV with different backbones, radar processing as well as \methodName Stages 1 to 3.}
		\label{tab:time}
		\centering
		\scalebox{0.65}{
			\begin{tabular}{| c | c | c | c | c | c | c |}
				\myTopRule
				Components & 
				\begin{tabular}{@{}c@{}}SparseBEV \\ (V2-99)\end{tabular}  & 
				\begin{tabular}{@{}c@{}}SparseBEV \\(ResNet101)\end{tabular} & 
				\begin{tabular}{@{}c@{}}Radar 
					\\Processing\end{tabular}  &
				Stage 1 & Stage 2 & Stage 3 \\
				\myTopRule
				Time (ms)  &  $575.7$  &  $211.5$ &  $105.4$ &  $1.0$  &  $12.7$ & $1.2$ \\
				\myTopRule
		\end{tabular}}
	\end{table}
	
	\section{Additional Ablation Studies}
	In the following ablations, we use \sparseBEV~\cite{liu2023sparsebev} with backbone ResNet101~\cite{he2016deep} for the monocular components in \methodName. For efficiency, the data used for evaluation are a subset of nuScenes validation set with $600$ random samples.

	\begin{figure}[b]
		\centering
		\includegraphics[width=0.6\linewidth]{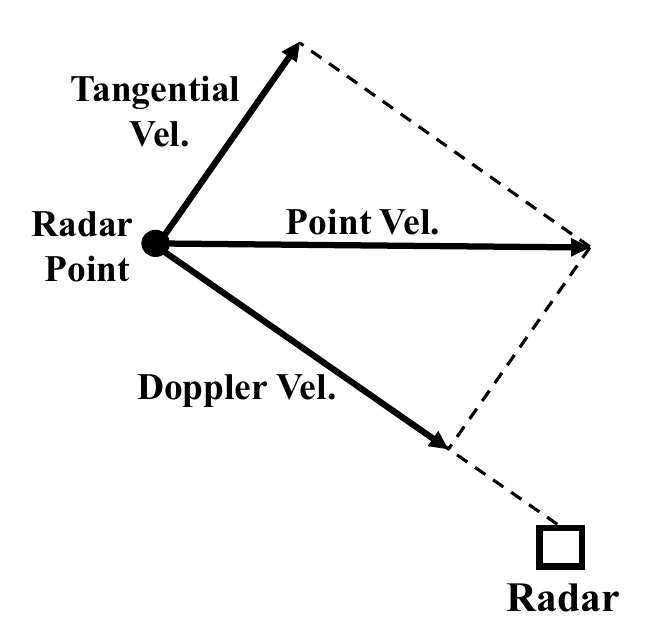}
		\caption{Geometric relation between point velocity and its radial and tangential components.}
		\label{fig:point_vel}
	\end{figure}

	\subsection{Ablation on Velocity Used for Point Motion Compensation}
	When accumulating $7$ radar sweeps in inference, we used estimated radar point velocity to compensate motions of moving points. We implement different velocity estimations and compare resultant detection performance in Tab.~\ref{tab:ab_vel}. The velocity estimations include $0$, {\it i.e.}, no motion compensation, Doppler velocity, Doppler velocity back-projected to estimated object heading direction, Doppler velocity plus tangential component of monocular estimated velocity, and monocular velocity. The geometric relation between full velocity and its tangential and radial components are shown in Fig.~\ref{fig:point_vel}. From Tab.~\ref{tab:ab_vel} we can see performing motion compensation improves detection performance and using full velocity estimates achieves better accuracy compared with only applying Doppler velocity. The three full velocity estimates shown on the $4$th to $6$th rows result in almost the same detection performance.
	
	\begin{table}[t]
		\caption{Ablation on different velocity estimations for compensating motion during radar sweep accumulation. Keys: Vel.= Velocity; Mono.= Monocular}
		\label{tab:ab_vel}
		\centering
		\scalebox{1.0}{
			\rowcolors{2}{lightgray}{white}
			\begin{tabular}{| c | c | c | }
				\myTopRule
				Point Vel. Estimation & 
				NDS (\uparrowRHDSmall)  & 
				mAP (\uparrowRHDSmall)  \\
				\myTopRule
				$0$  &  $0.614$  &  $0.534$ \\
				Doppler Vel.  &  $0.620$  &  $0.544$ \\
				Back-Projected Doppler Vel.  &  \best{0.621}  &  $0.545$ \\
				Doppler + Tangential Mono. Vel.  &  \best{0.621}  &  $0.545$ \\
				Mono. Vel. &  \best{0.621}  &  \best{0.546} \\
				\myTopRule
		\end{tabular}}
	\end{table}

	\subsection{Ablation on Range and Score Updating}
	In Stage-3 inference we update both range and detection score. Detection scores indicate confidence in prediction and have an impact on mAP computation, where predictions with higher scores have priority as true positives to be associated with GT. We update detection scores by adding Stage-3 scores weighted by $\alpha$ to monocular scores. We test different range and score updating options with different $\alpha$ and list resultant detection performance in Tab.~\ref{tab:ab_alpha}.
	We can see both range and score updating improve detection performance while range updating has significantly bigger impacts on performance.
	
	\begin{table}[t]
		\caption{Ablation on updating range and detection score with fusion weight $\alpha$}
		\label{tab:ab_alpha}
		\centering
		\scalebox{1.0}{
			\begin{tabular}{| c | c |  c | c | c | }
				\myTopRule
				\begin{tabular}{@{}c@{}}Update \\ Range \end{tabular} &
				\begin{tabular}{@{}c@{}}Update \\ Score \end{tabular} &
				$\alpha$ & 
				NDS (\uparrowRHDSmall)  & 
				mAP (\uparrowRHDSmall)  \\
				\myTopRule
				&   & -  &  $0.590$  &  $0.501$ \\
				\hline
				& \checkmark & $0.5$  &  $0.593$  &  $0.503$ \\
				\hline
				\checkmark &   & -  &  $0.617$  &  $0.543$ \\
				\hline
				\checkmark & \checkmark & $0.2$  &  $0.620$  &  \best{0.545} \\
				\hline
				\checkmark & \checkmark & $0.5$  &  \best{0.621}  &  \best{0.545} \\
				\hline
				\checkmark & \checkmark & $0.8$  &  \best{0.621}  &  $0.543$ \\
				\hline
				\checkmark & \checkmark & $1.0$ &  $0.620$  &  $0.541$ \\
				\myTopRule
		\end{tabular}}
	\end{table}

	\subsection{Ablation on Number of Radar Sweeps}
	Within $0.5$s time window, there are about 7 sweeps of radar points ({\it i.e.}, $1$ current plus $6$ past ones) from radars running at $13$Hz in \nuscenes Dataset~\cite{caesar2020nuscenes}. We accumulate multiple radar sweeps during inference, and the number of radar sweeps may impact detection performance, as more sweeps provide denser radar measurement used for Stage 2. To verify this, we run \methodName multiple times with radar input from $0$, $1$, $3$, $5$, and $7$ sweeps, respectively and record their detection performance. Note using $0$ radar sweep refers to applying only monocular detector without fusion. As shown in Tab.~\ref{tab:ab_sweeps}, more radar sweeps lead to better detection performance as expected.
	
	\begin{table}[t]
		\caption{Ablation on Number of Radar Sweeps. More radar sweeps result in better detection performance. Key: Num.= Number}
		\label{tab:ab_sweeps}
		\centering
		\scalebox{1.0}{
			\rowcolors{2}{lightgray}{white}
			\begin{tabular}{| c | c | c | }
				\myTopRule
				Num. of Sweeps & 
				NDS (\uparrowRHDSmall)  & 
				mAP (\uparrowRHDSmall)  \\
				\myTopRule
				$0$  &  $0.590$  &  $0.501$ \\
				$1$  &  $0.597$  &  $0.512$ \\
				$3$  &  $0.612$  &  $0.531$ \\
				$5$  &  $0.618$  &  $0.541$ \\
				$7$  &  \best{0.621}  &  \best{0.545} \\
				\myTopRule
		\end{tabular}}
	\end{table}

	\section{Additional Analyses}
	\subsection{Performance at Night}
	Although using radar to handle adverse conditions is a different research focus, we show in Tab.~\ref{tab:night} that \methodName significantly improves detection performance over the underlying monocular detector under challenging lighting conditions at night. We evaluate on 7 object categories, which appear in night scenes in nuScenes validation set. We also list corresponding daytime and overall performance for reference.
	
	\begin{table}[t]
		\caption{Detection performance NDS($\uparrow$) / mAP($\uparrow$) of \methodName and its  underlying monocular detector in night, daytime and all validation scenes, respectively.}
		\label{tab:night}
		\centering
		\scalebox{0.95}{
			\footnotesize
			\setlength\tabcolsep{0.13cm}
			\begin{tabular}{ | c | c | c | c |}
				\hline
				Scene      & Night & Daytime & Overall\\
				\hline
				SparseBEV~\cite{liu2023sparsebev}  & 0.526 / 0.400  & 0.673 / 0.601  & 0.669 / 0.595\\
				SparseBEV + \textbf{RICCARDO} &  \textbf{0.561} / \textbf{0.450} & \textbf{0.704} / \textbf{0.642}   & \textbf{0.699} / \textbf{0.636}\\
				\hline
				Number of Samples & 602  & 5417 & 6019\\
				\hline
			\end{tabular}
		}
	\end{table}
	
	\subsection{Sensitivity to Monocular Prediction Errors}
\begin{figure}[t]
		\centering
		\includegraphics[width=1\linewidth]{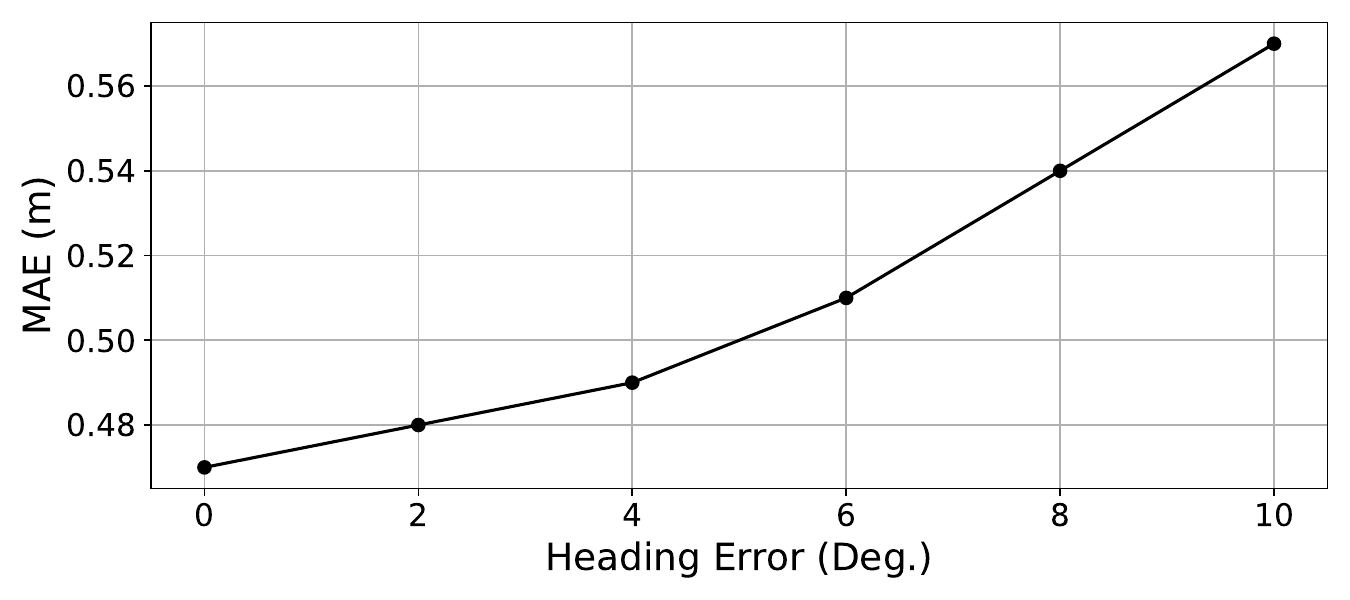}
		\includegraphics[width=1\linewidth]{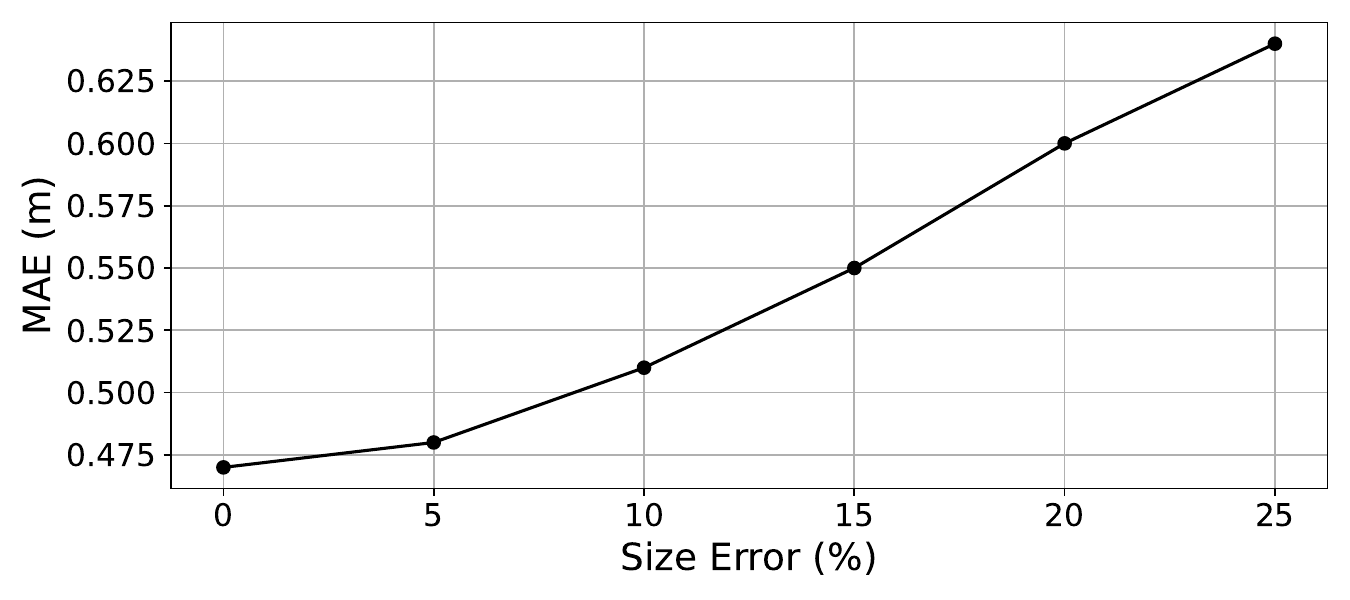}
		\caption{Sensitivity of the Stage-2 prediction to monocular prediction errors in heading (top) and size (bottom).}
		\label{fig:sensitivity}
	\end{figure}

	To assess sensitivity of the Stage-2 prediction to monocular prediction errors, we add errors to the bounding box parameters, which are inputs to the radar distribution model, and compute the MAE of range estimation in Stage 2. The Fig.~\ref{fig:sensitivity} shows the range error modestly increases with errors in heading angle and size, demonstrating good monocular error toleration. 

	\section{Additional Visualizations}
	\subsection{Visualization of Radar Distributions}
\begin{figure*}[t]
		\centering
		\scalebox{1.1}
		{
			\begin{tabular}{@{}c@{}c@{}c@{}c@{}c@{}}
				\\
				{\small $0^\circ$ } &
				\includegraphics[height=1.2in]{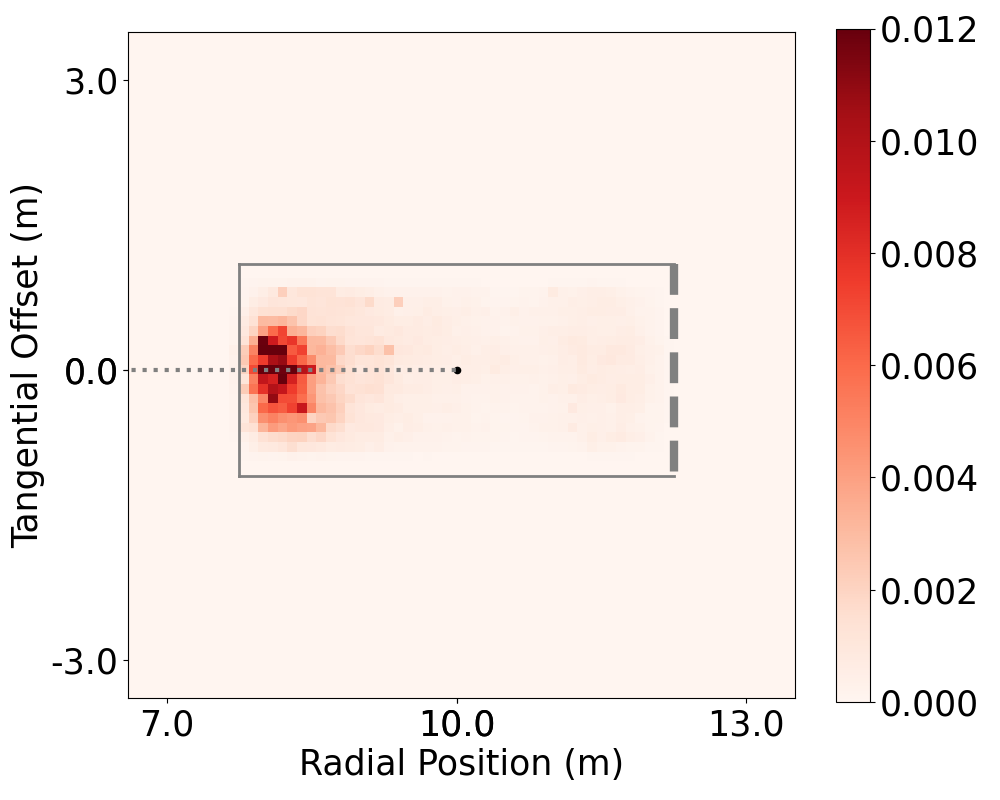}&
				\includegraphics[height=1.2in]{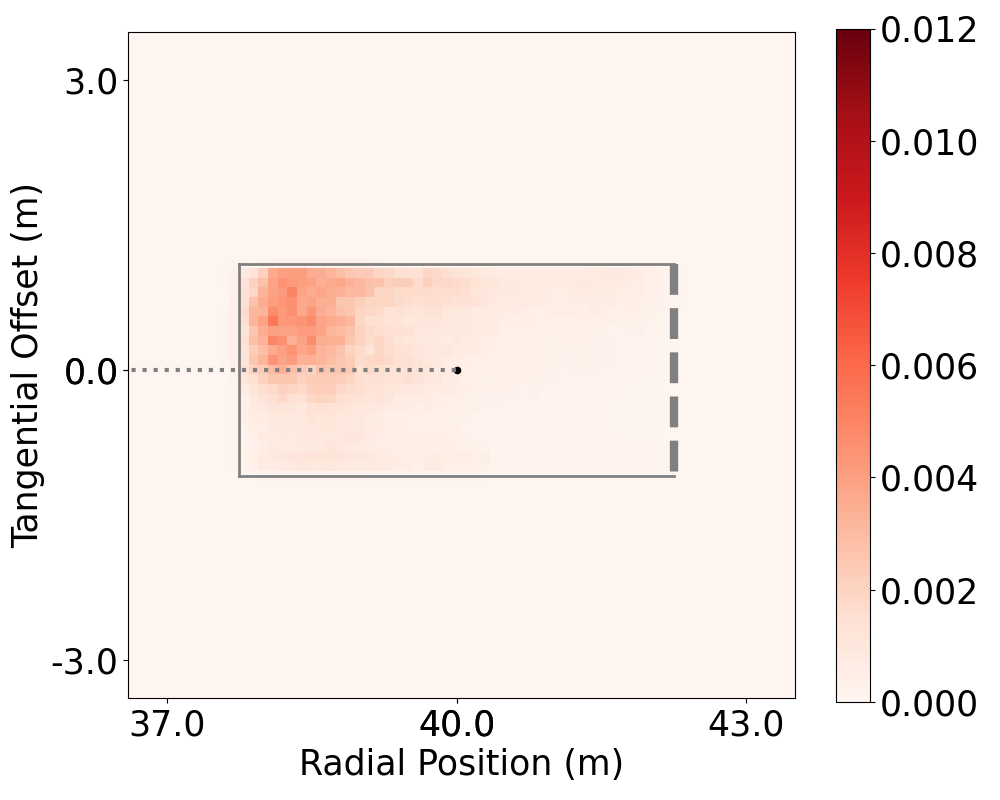}&	
				\includegraphics[height=1.2in]{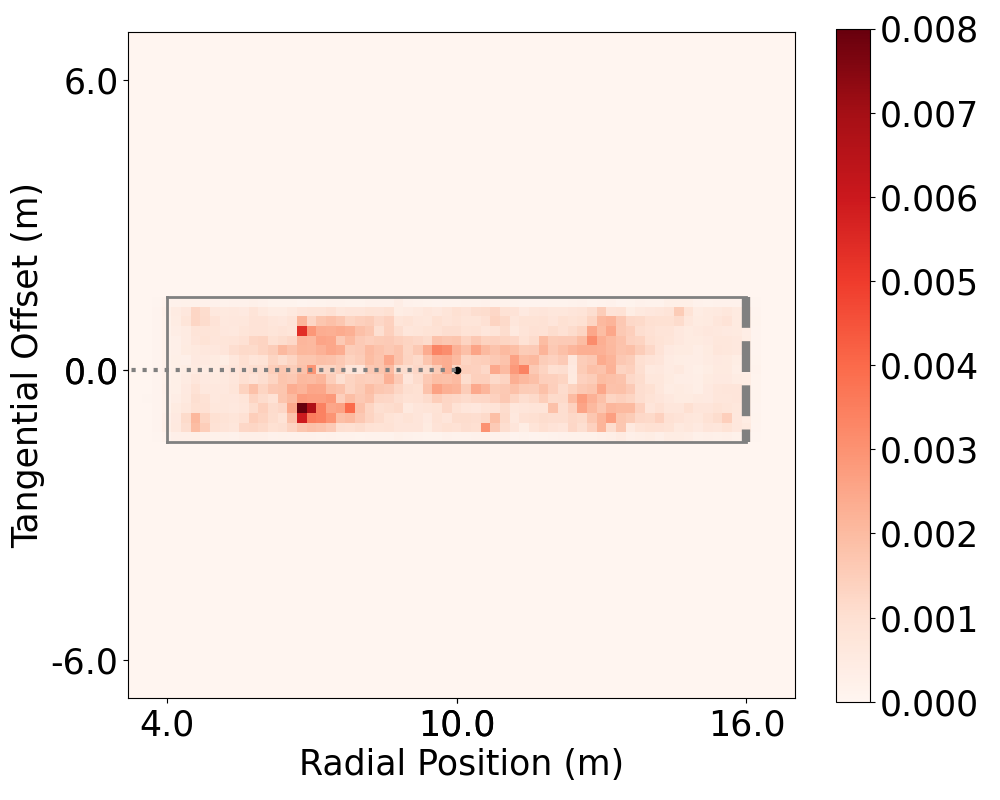}&
				\includegraphics[height=1.2in]{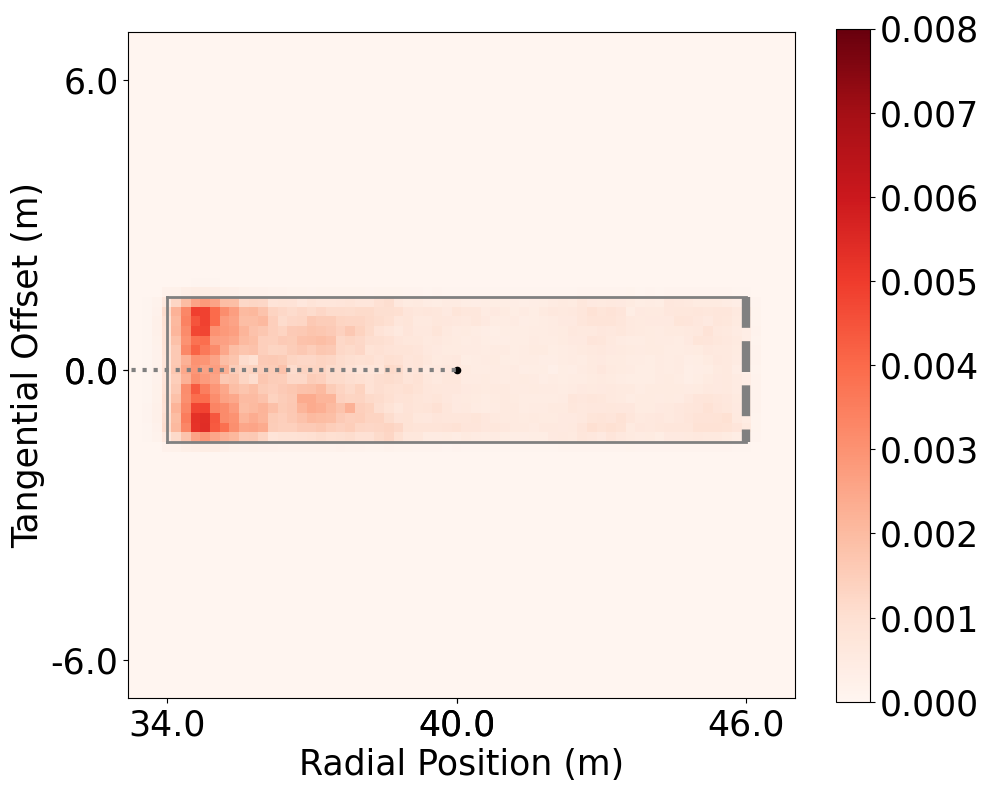}
				\\
				{\small $45^\circ$ } &
				\includegraphics[height=1.2in]{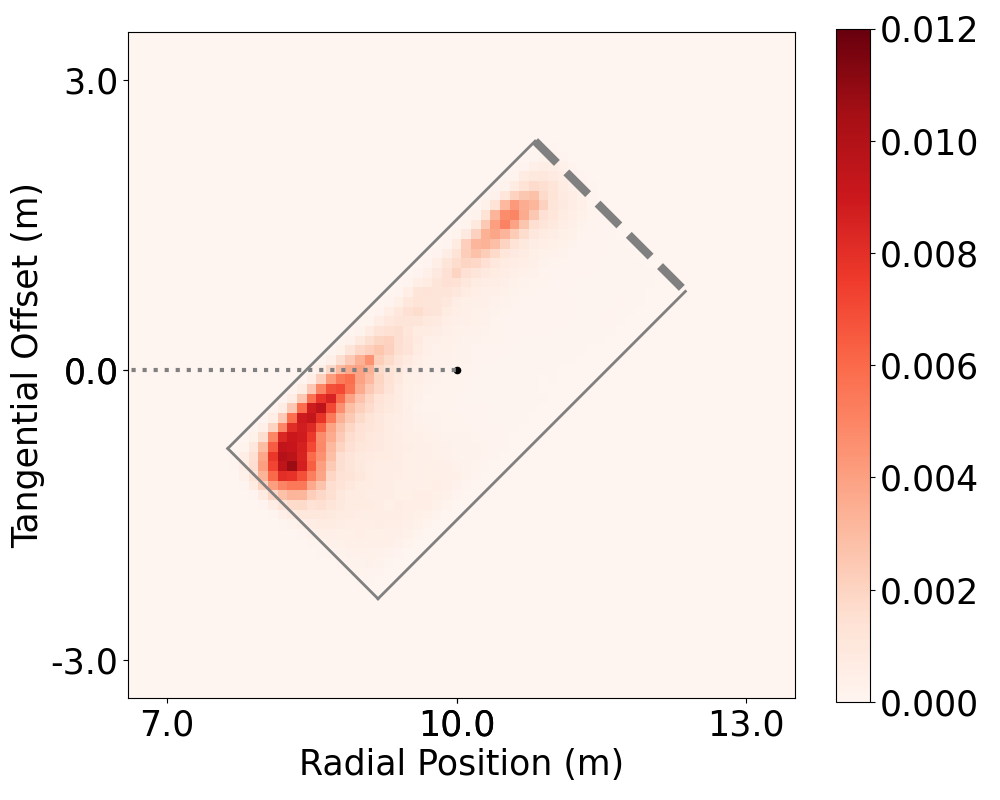}&
				\includegraphics[height=1.2in]{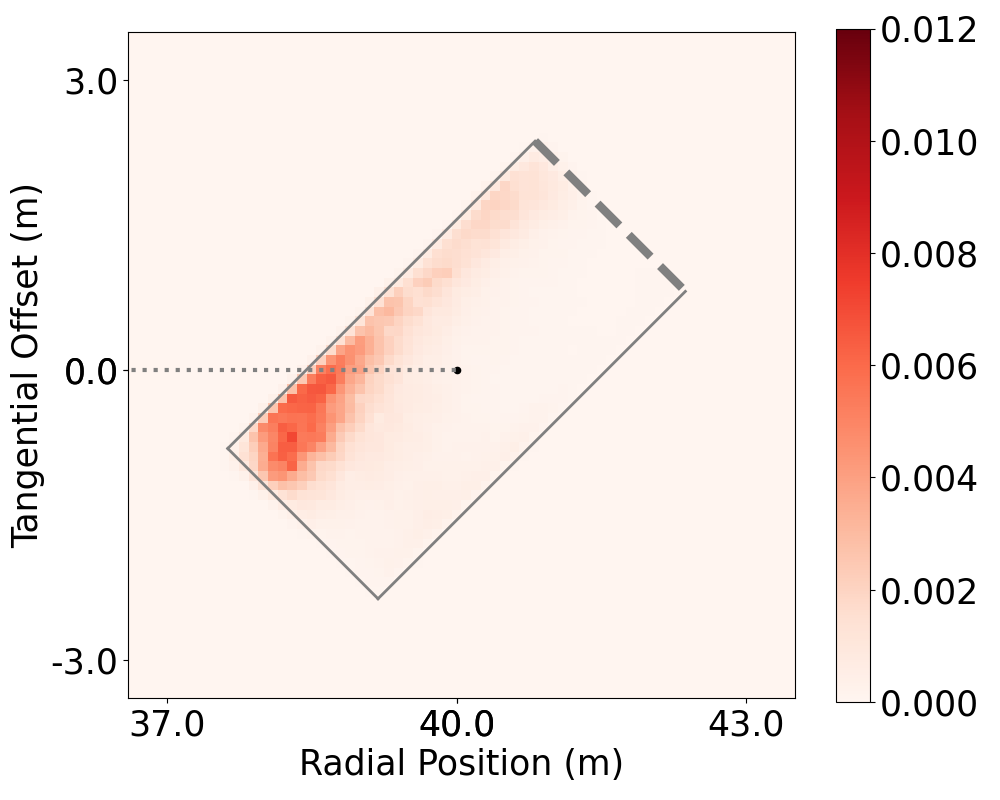}&	
				\includegraphics[height=1.2in]{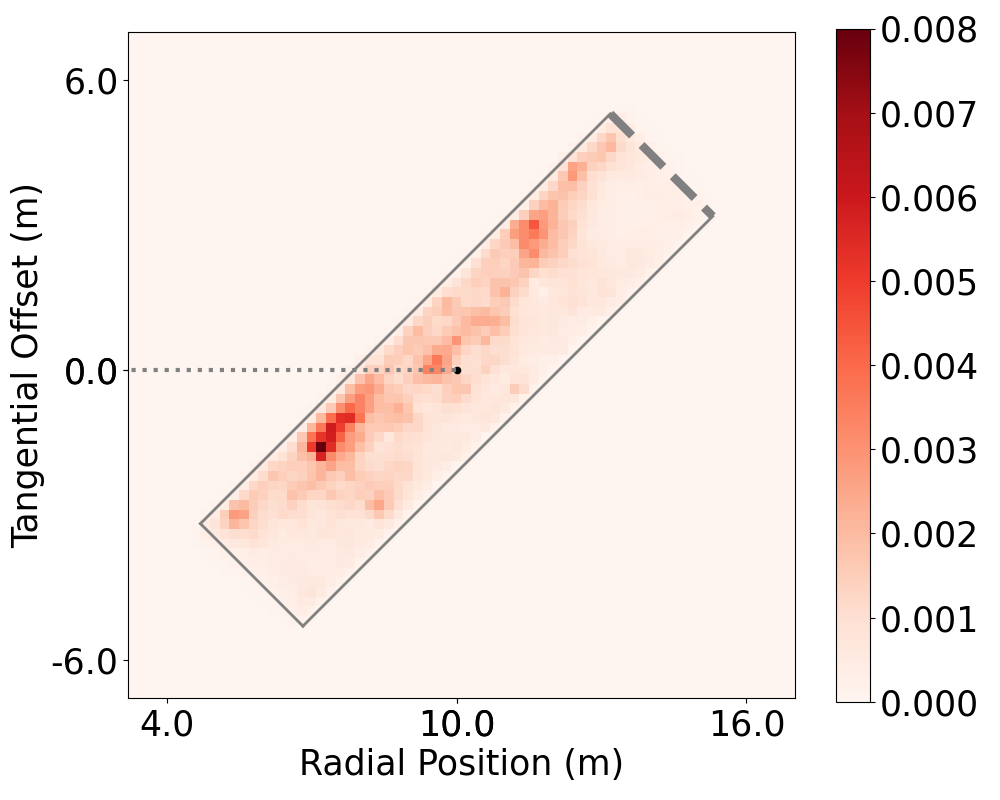}&
				\includegraphics[height=1.2in]{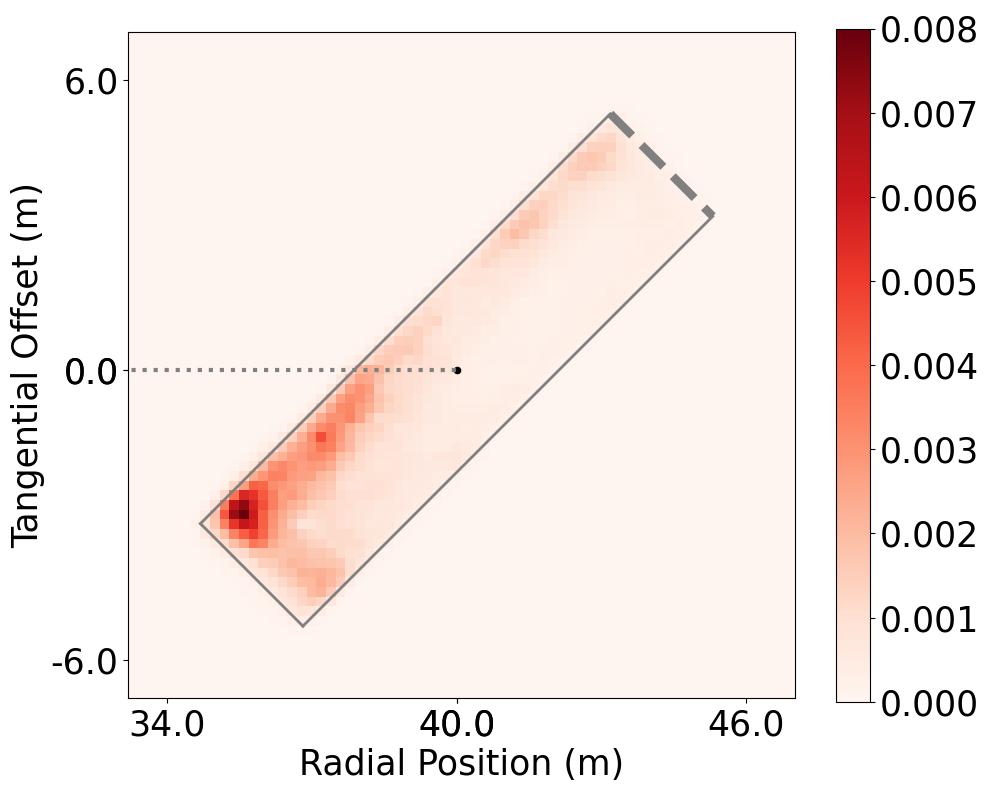}
				\\
				{\small $90^\circ$ } &
				\includegraphics[height=1.2in]{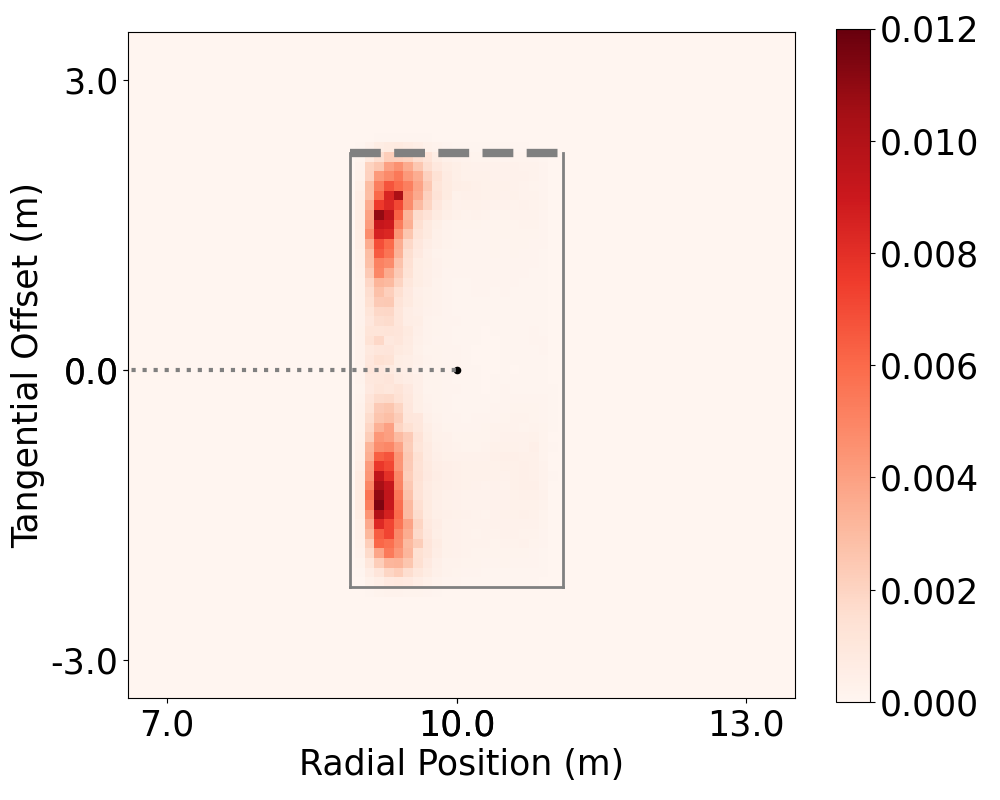}&
				\includegraphics[height=1.2in]{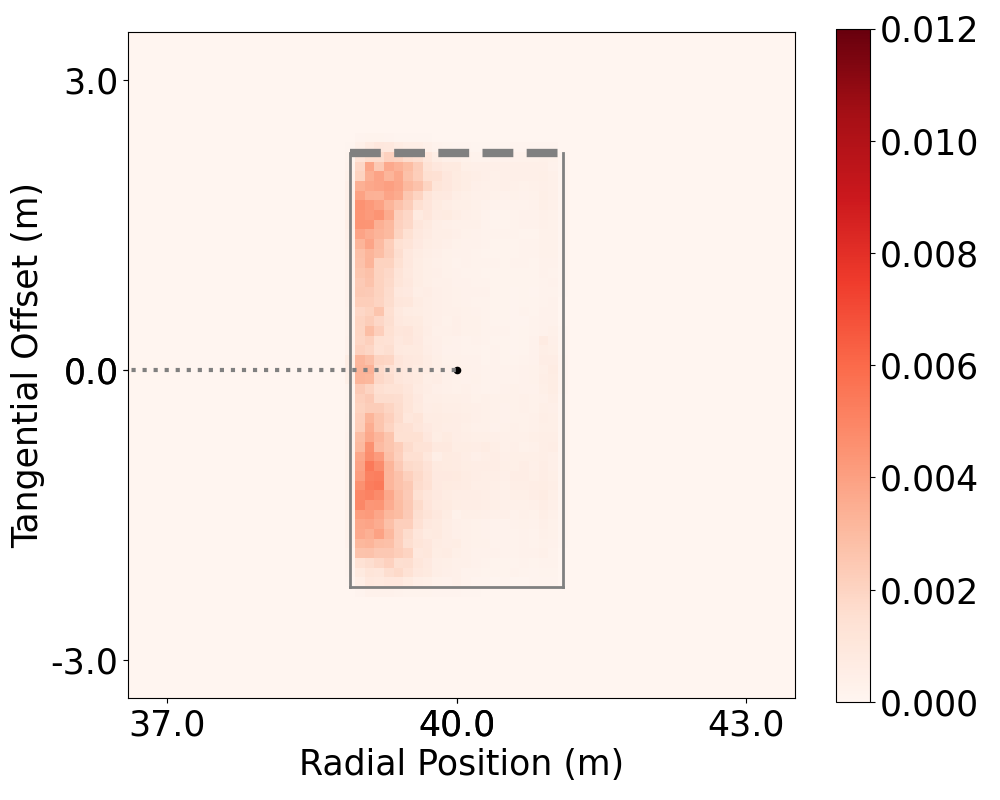}&	
				\includegraphics[height=1.2in]{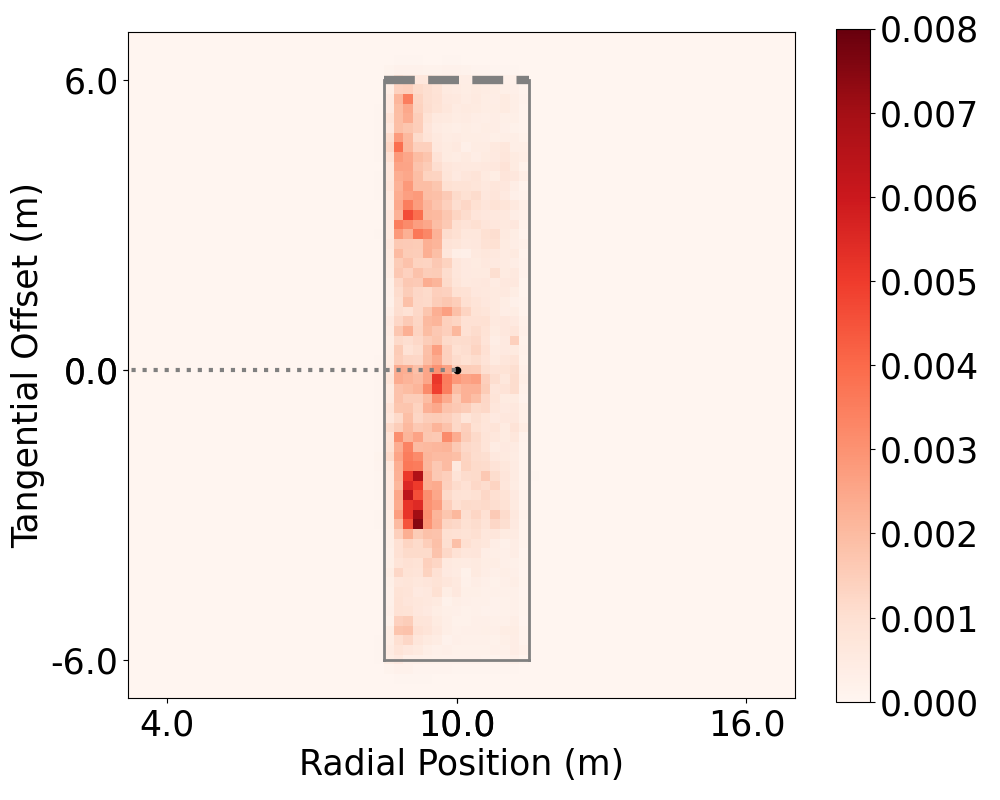}&
				\includegraphics[height=1.2in]{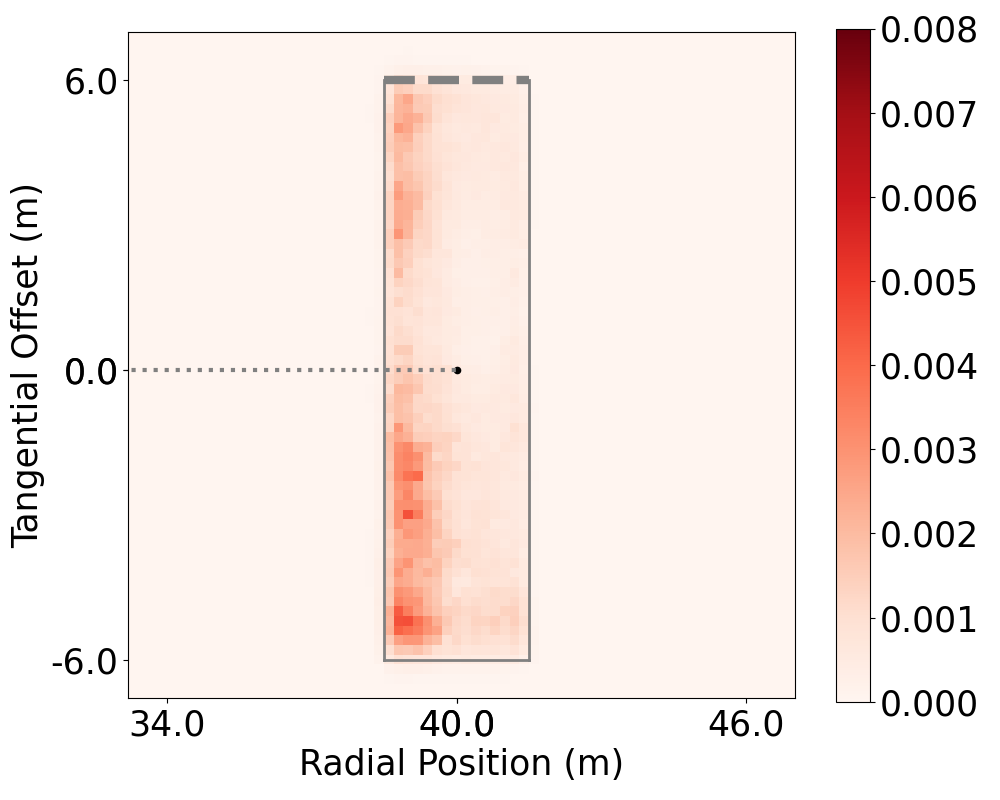}
				\\
				{\small $135^\circ$ } &
				\includegraphics[height=1.2in]{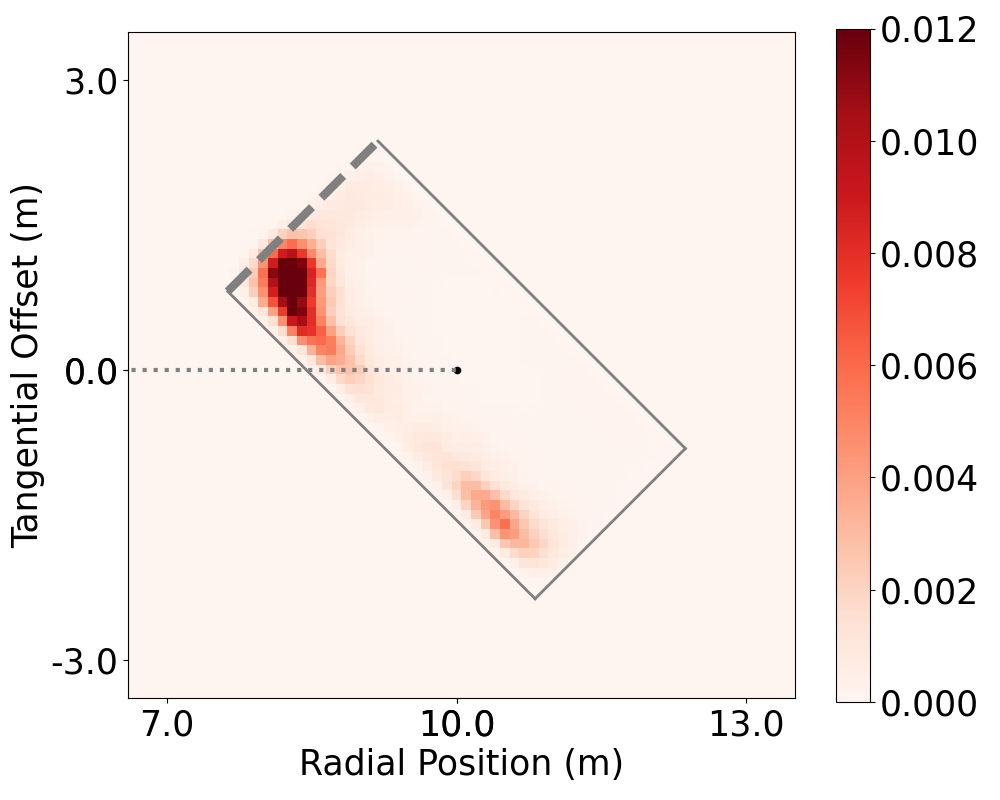}&
				\includegraphics[height=1.2in]{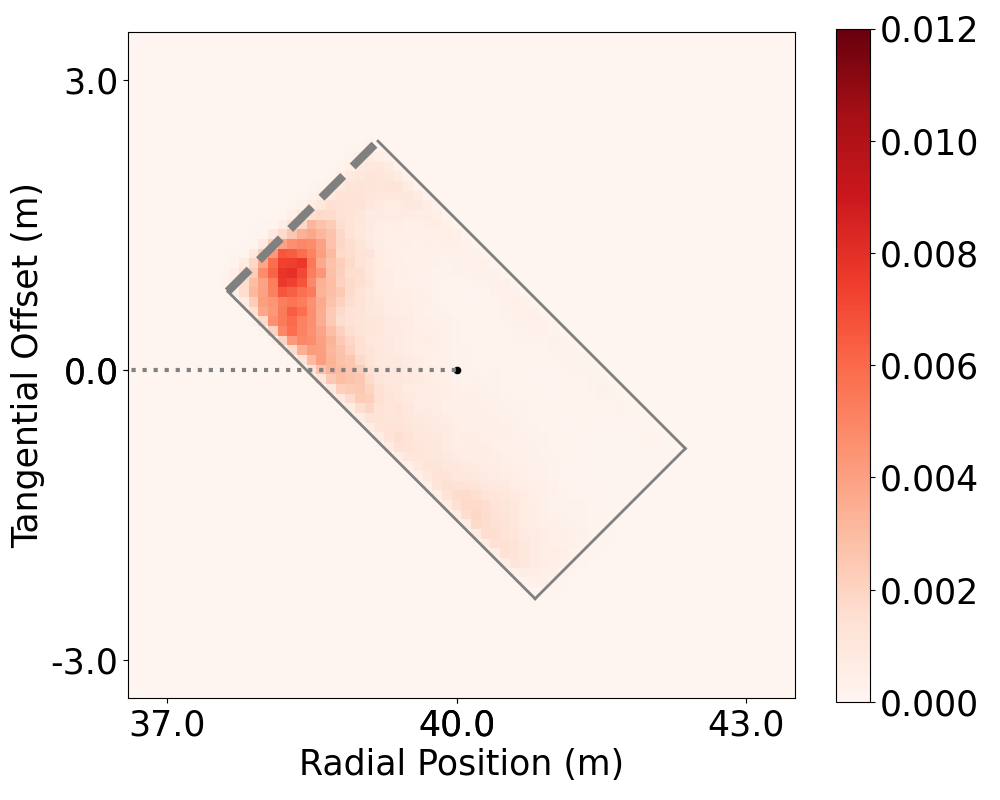}&	
				\includegraphics[height=1.2in]{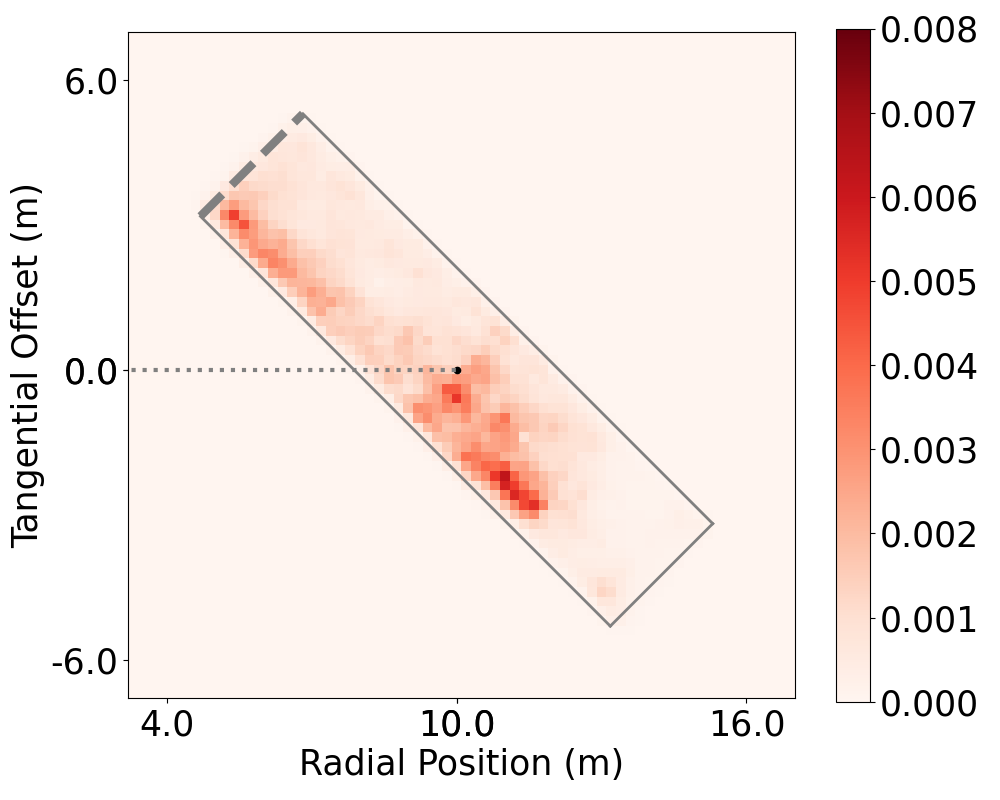}&
				\includegraphics[height=1.2in]{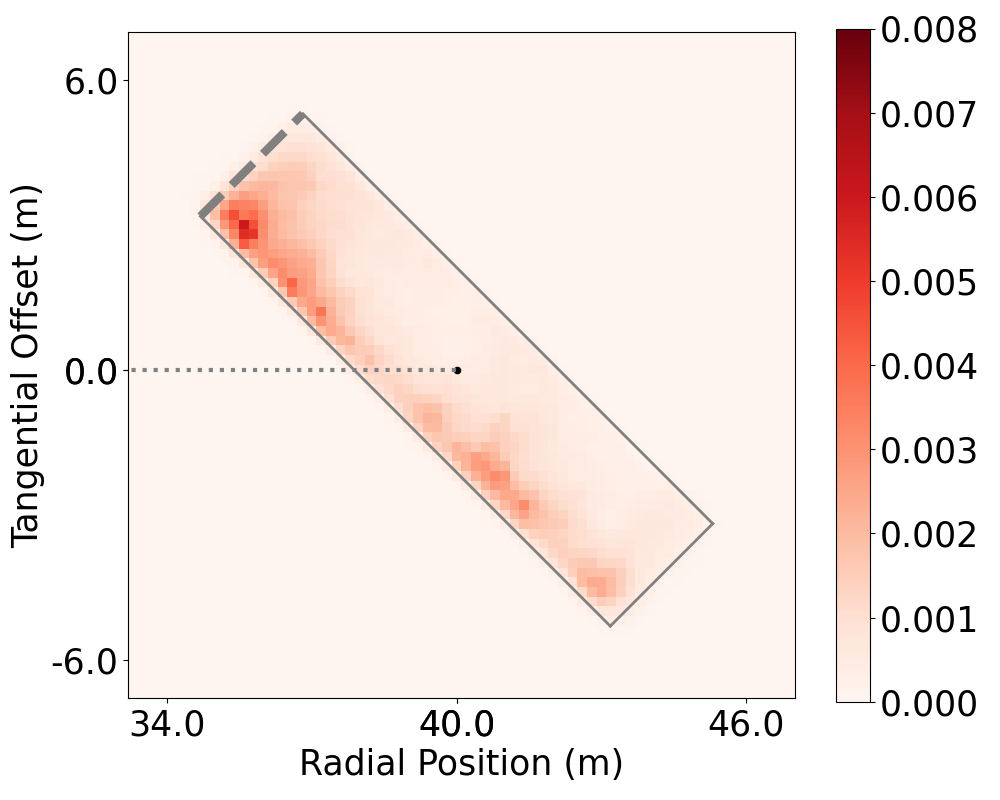}
				\\
				{\small $180^\circ$ } &
				\includegraphics[height=1.2in]{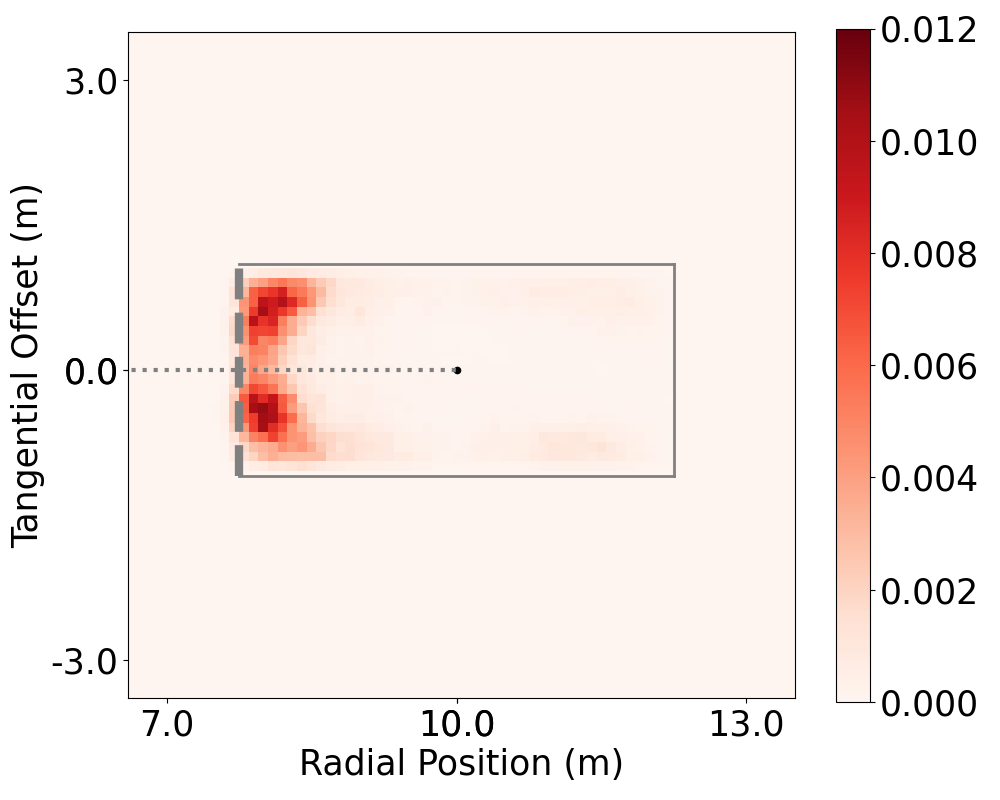}&
				\includegraphics[height=1.2in]{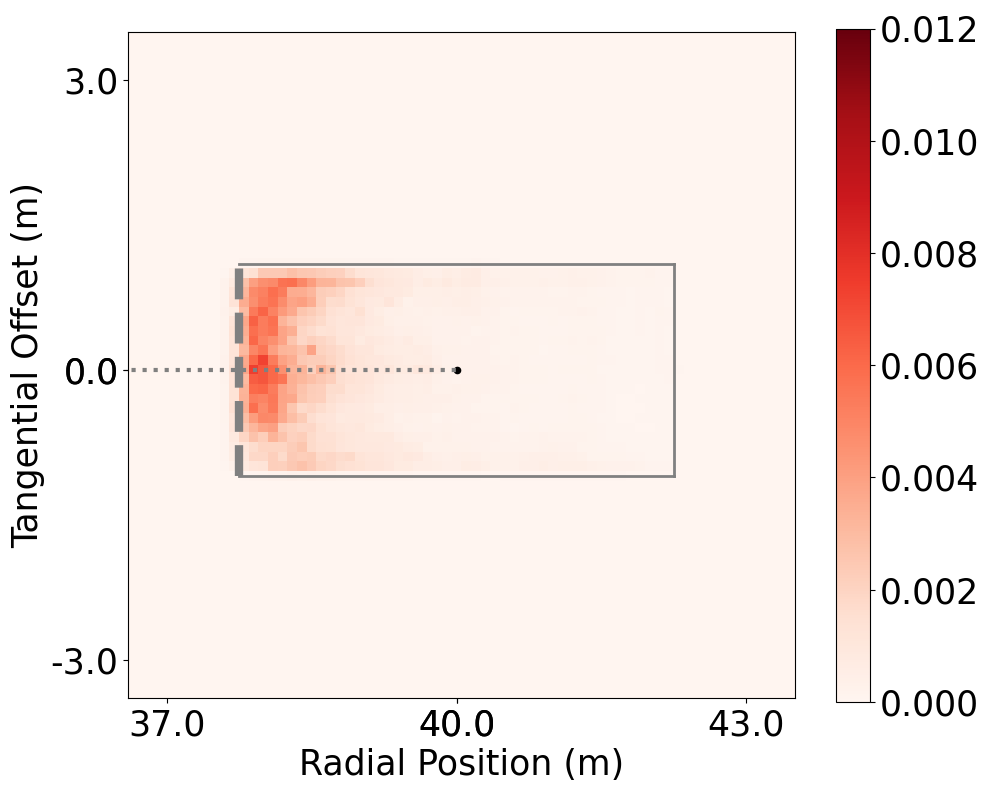}&	
				\includegraphics[height=1.2in]{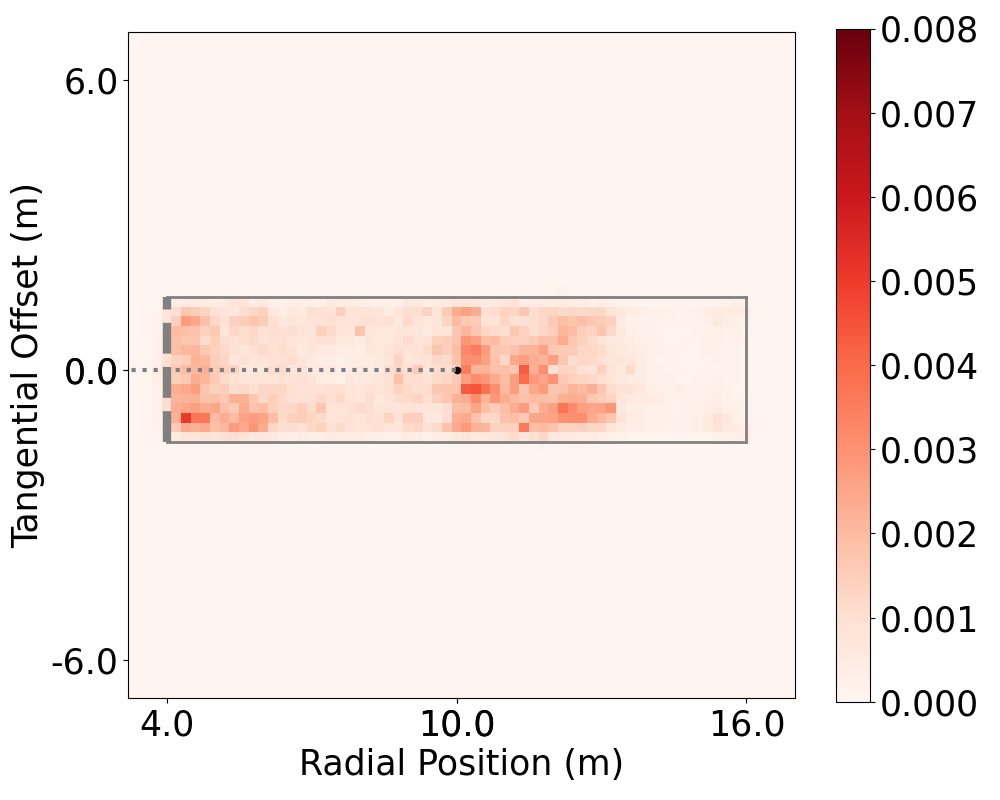}&
				\includegraphics[height=1.2in]{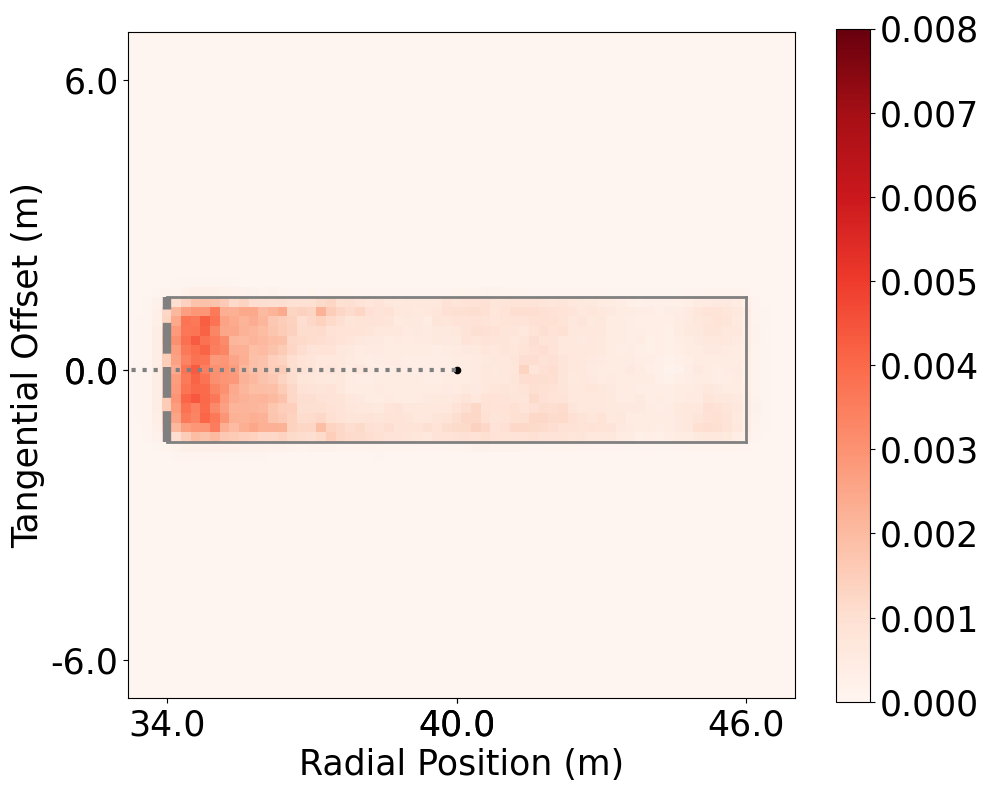}
				\\
				&
				\small (a) Car (Short Range) 	&
				\small (b) Car (Long Range)   &	
				\small (c) Bus (Short Range)  &	
				\small (d) Bus (Long Range) 
				
			\end{tabular}
		}
		\caption{ Visualization of predicted radar distributions of (a)(b) Car and (c)(d) Bus viewed from different angles and distances of $10$ and $40$ meters. X-axis represents radial positions, and Y-axis denotes tangential offsets to object centers. Radial rays are plotted as horizontal dotted lines. Target bounding boxes are shown on top of distributions, and dashed lines represent object head.
		}
		\label{fig:suppl_vis1}
	\end{figure*}

\begin{figure*}[t]
		\centering
		\scalebox{1.1}
		{
			\begin{tabular}{@{}c@{}c@{}c@{}c@{}c@{}}
				\\
				{\small $0^\circ$ } &
				\includegraphics[height=1.2in]{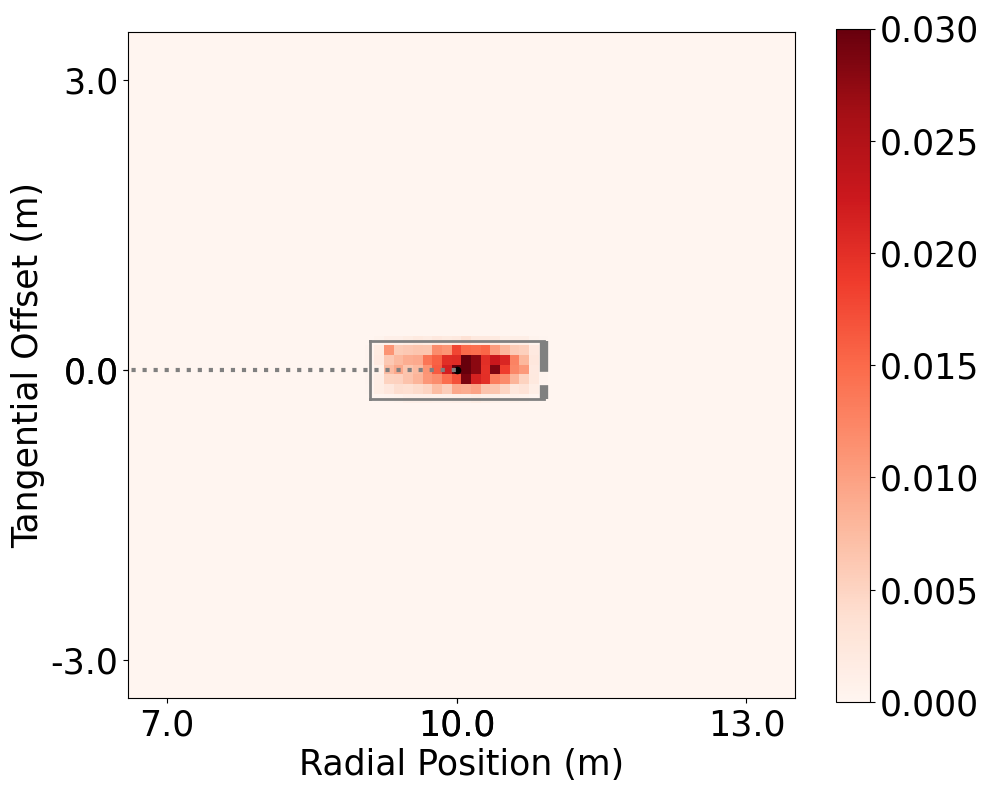}&
				\includegraphics[height=1.2in]{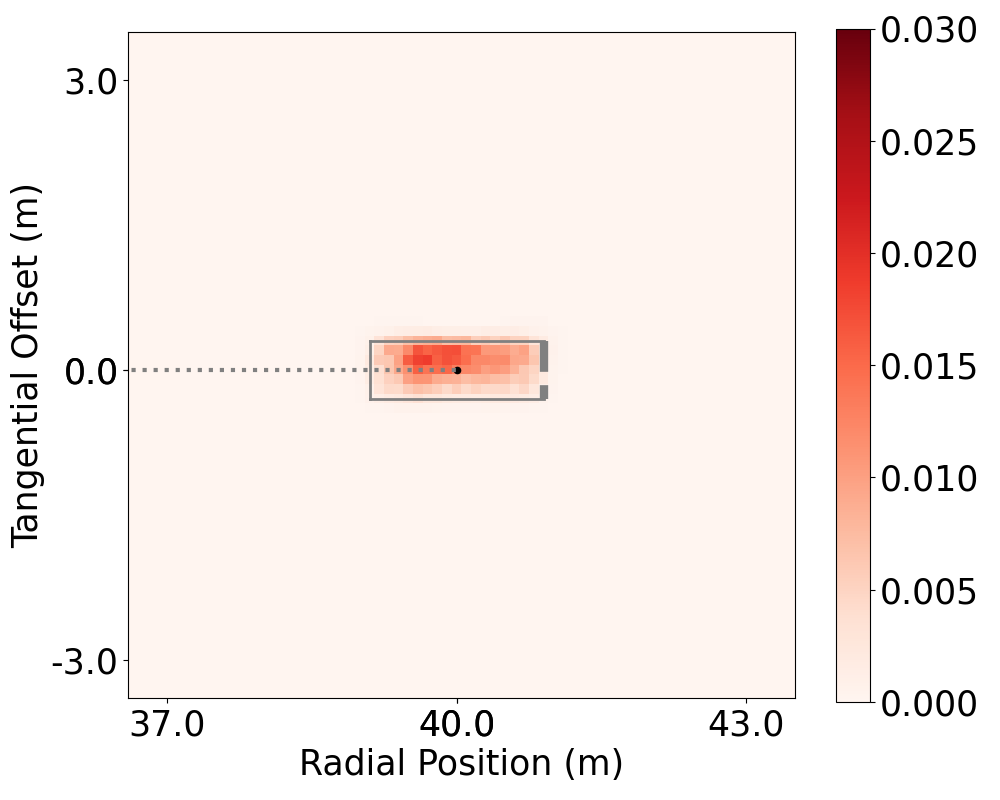}&	
				\includegraphics[height=1.2in]{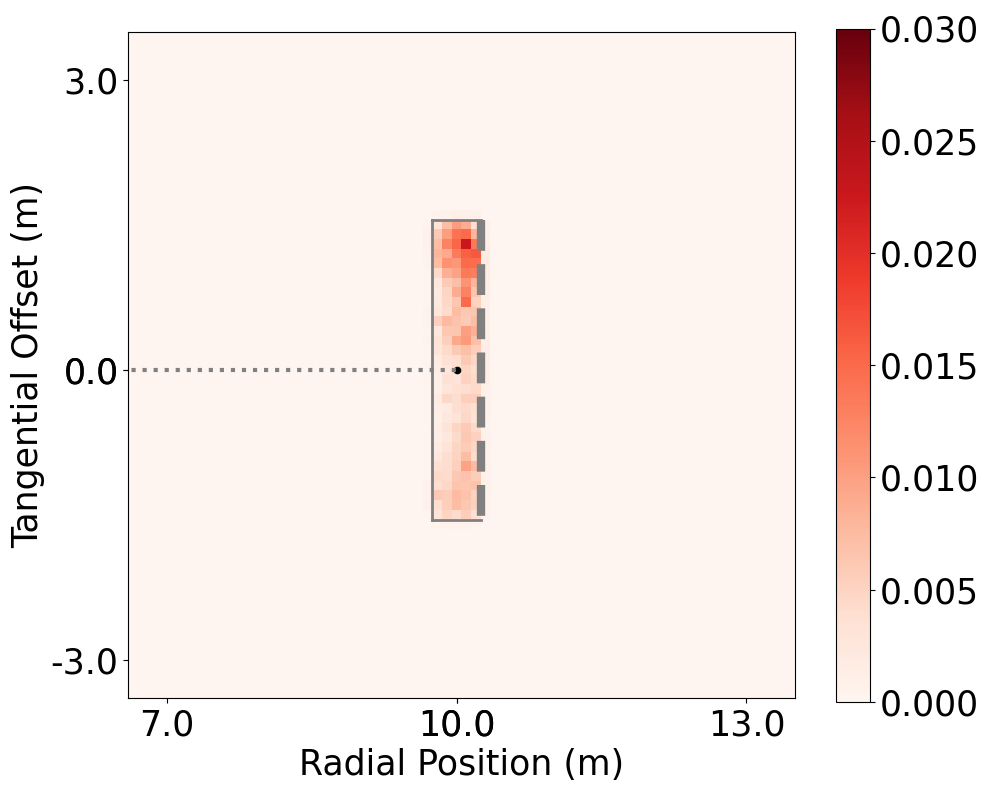}&
				\includegraphics[height=1.2in]{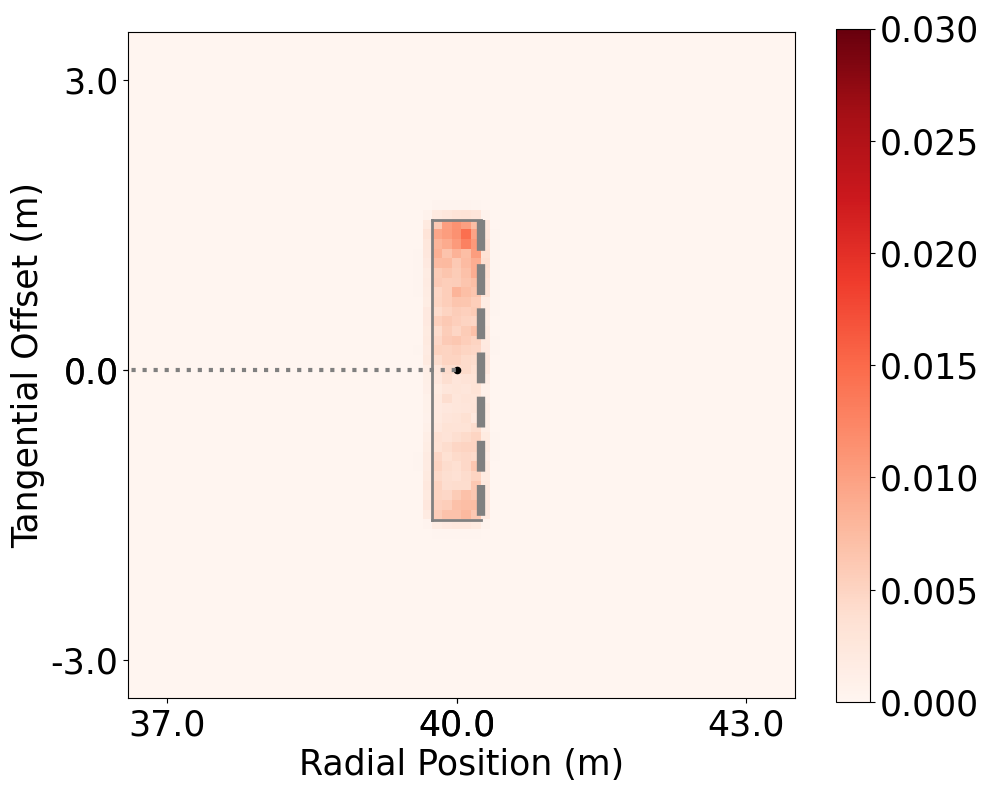}
				\\
				{\small $45^\circ$ } &
				\includegraphics[height=1.2in]{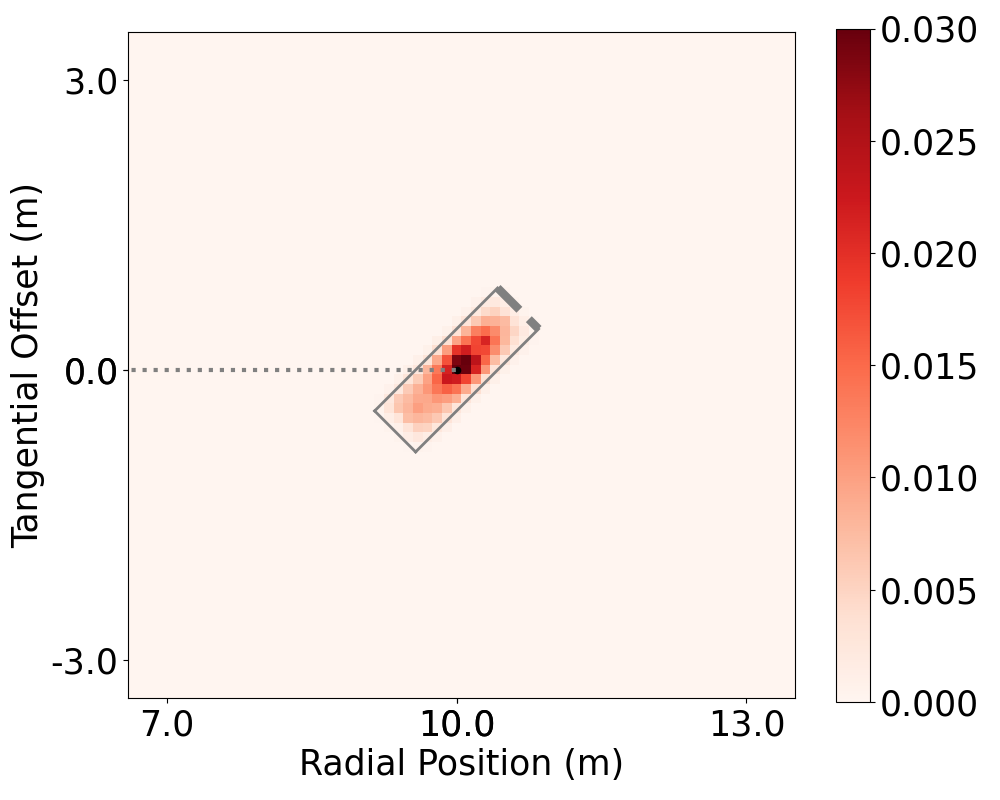}&
				\includegraphics[height=1.2in]{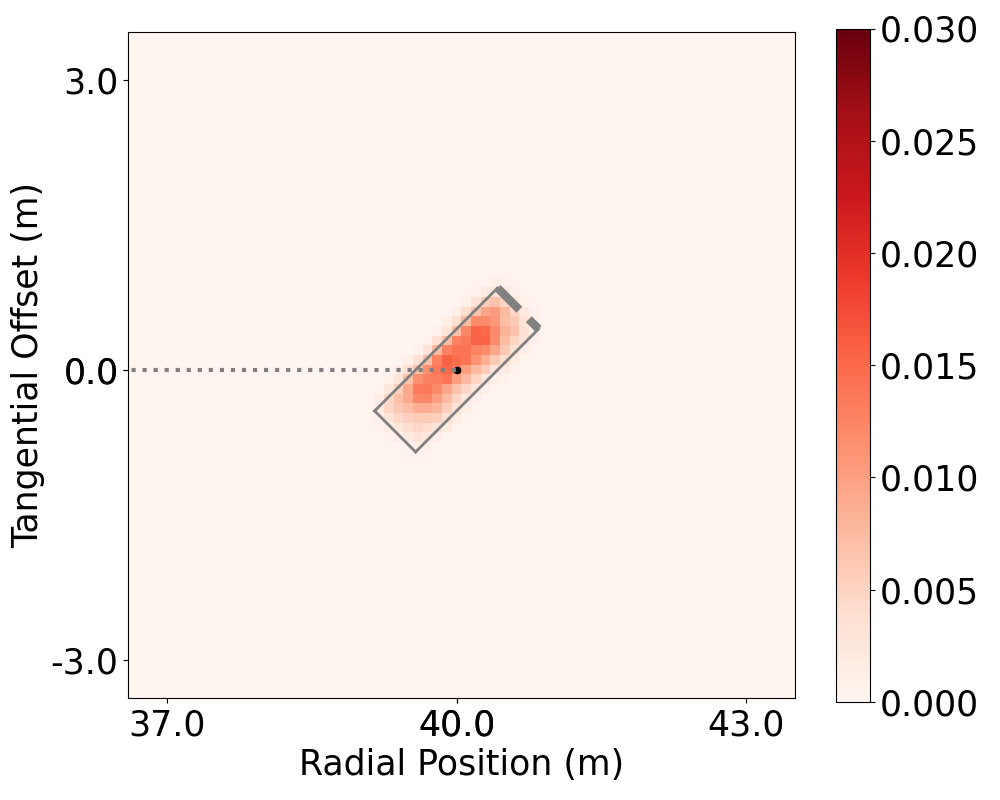}&	
				\includegraphics[height=1.2in]{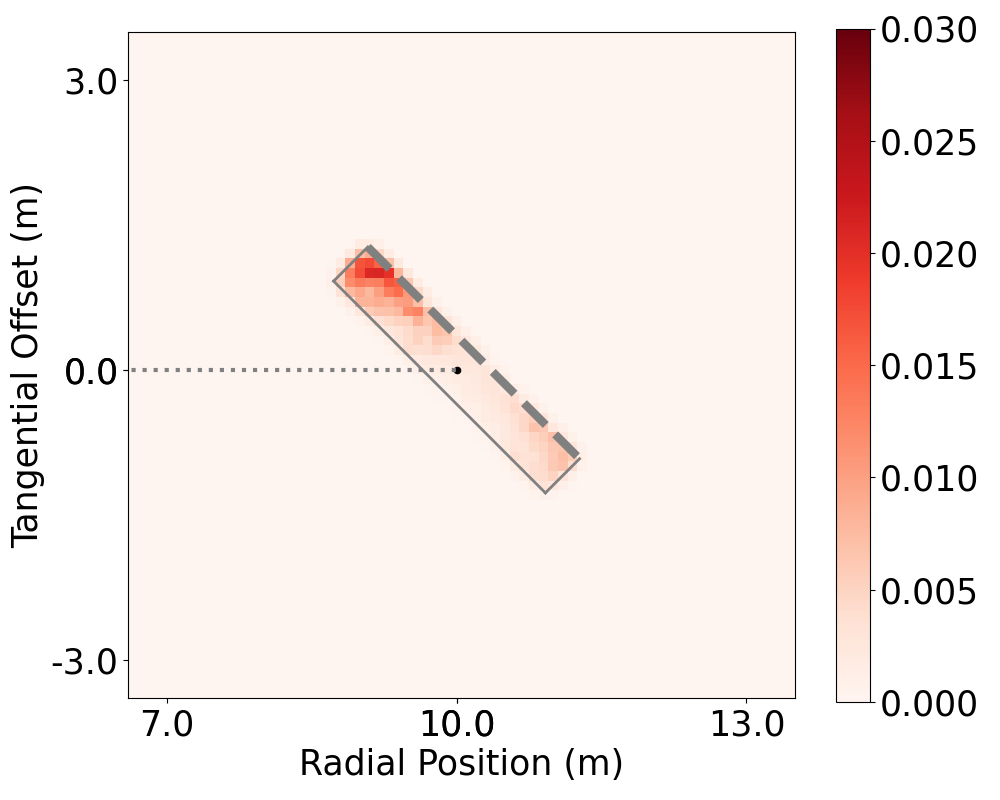}&
				\includegraphics[height=1.2in]{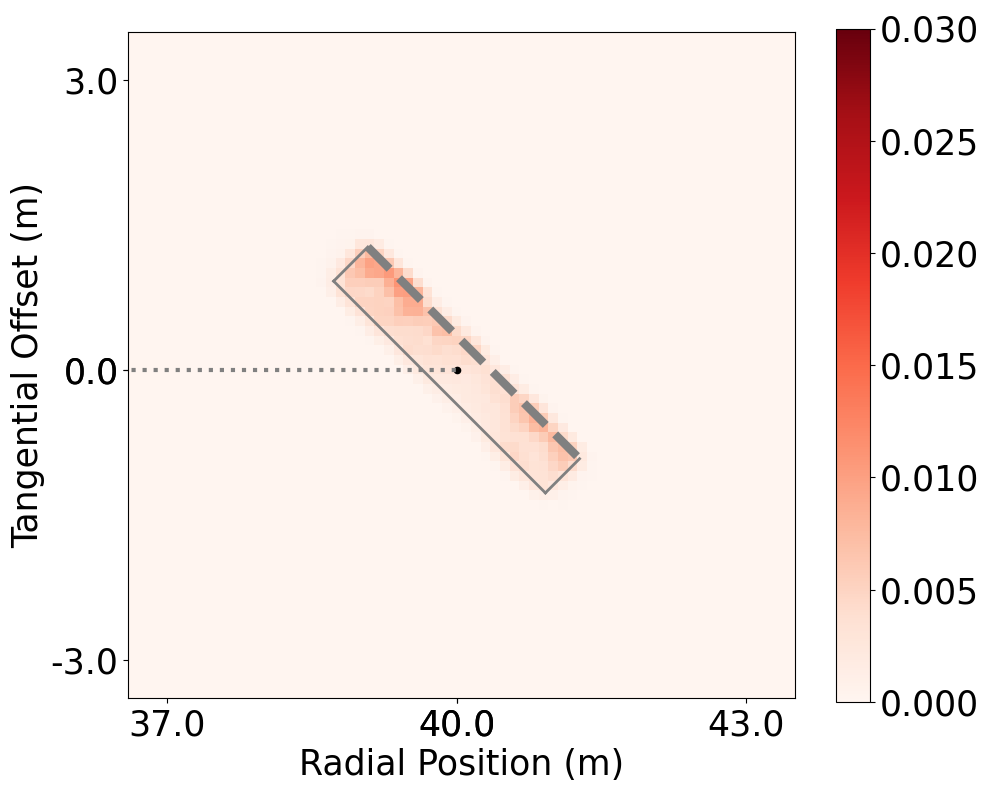}
				\\
				{\small $90^\circ$ } &
				\includegraphics[height=1.2in]{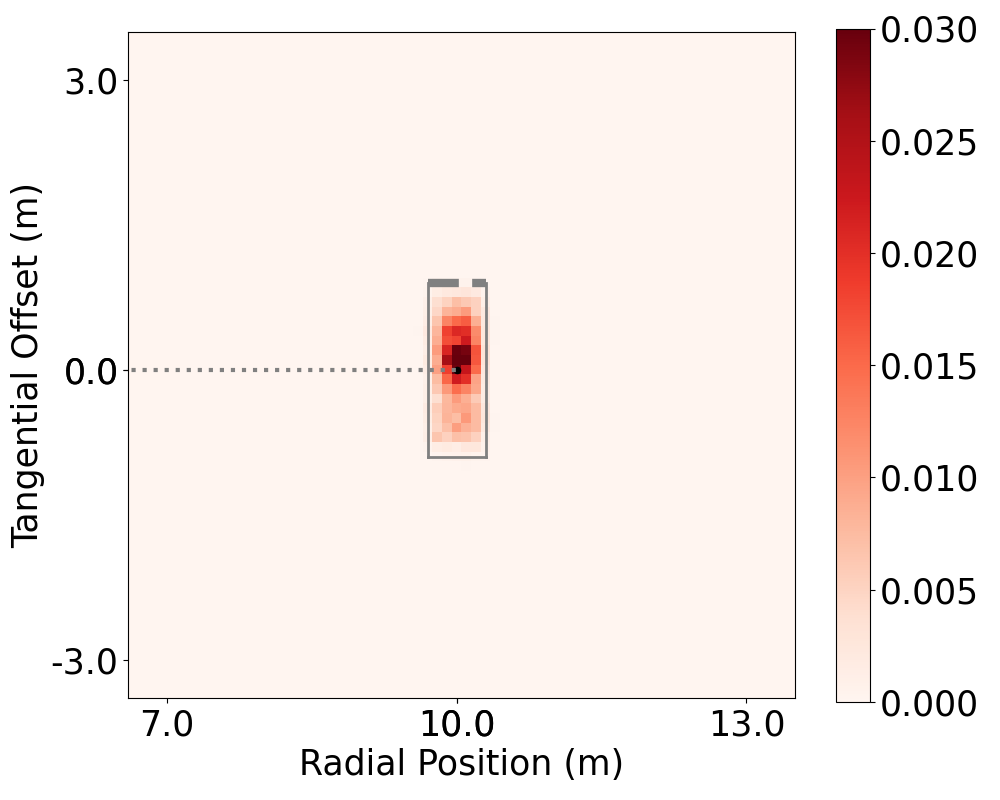}&
				\includegraphics[height=1.2in]{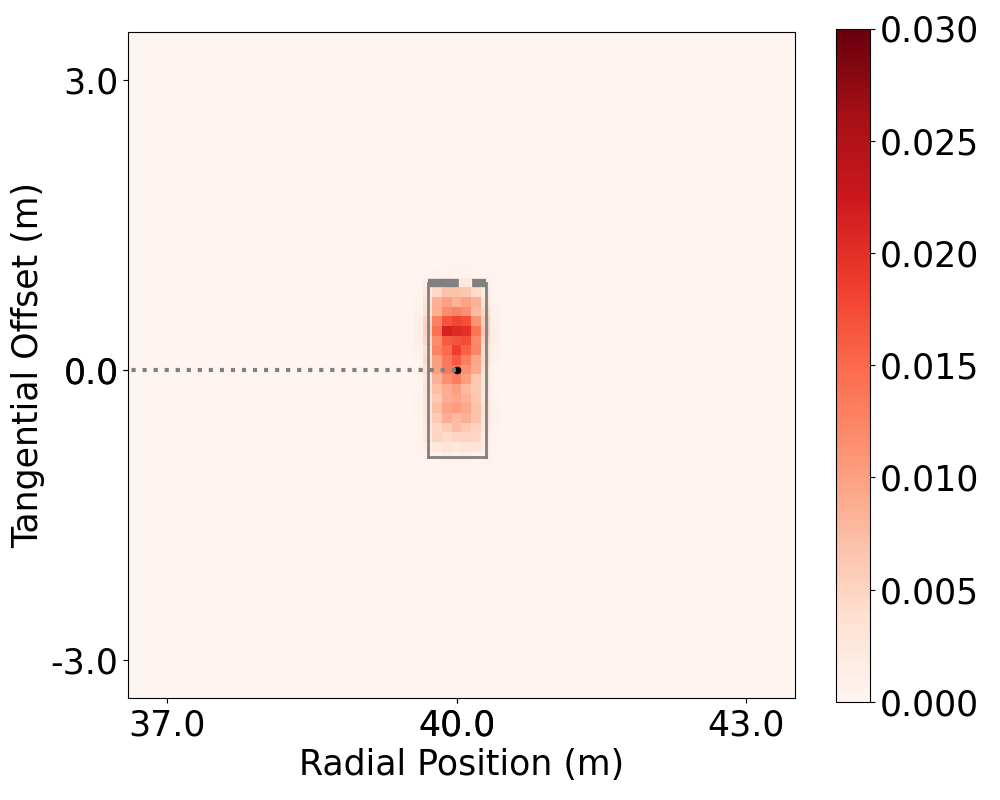}&	
				\includegraphics[height=1.2in]{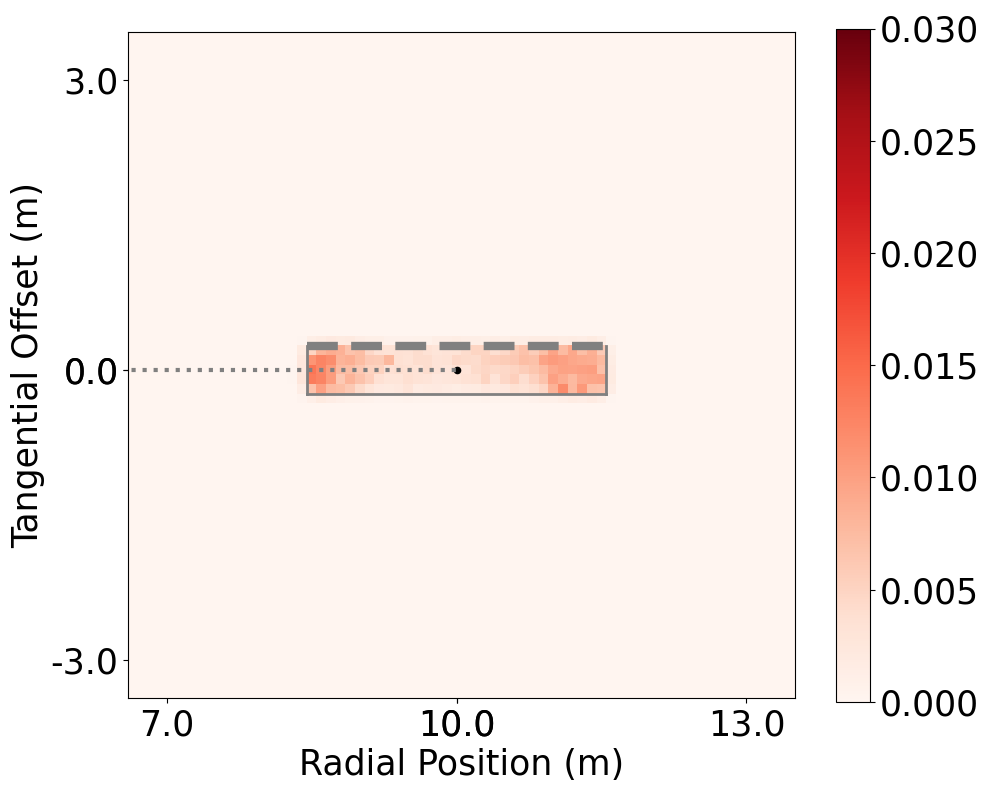}&
				\includegraphics[height=1.2in]{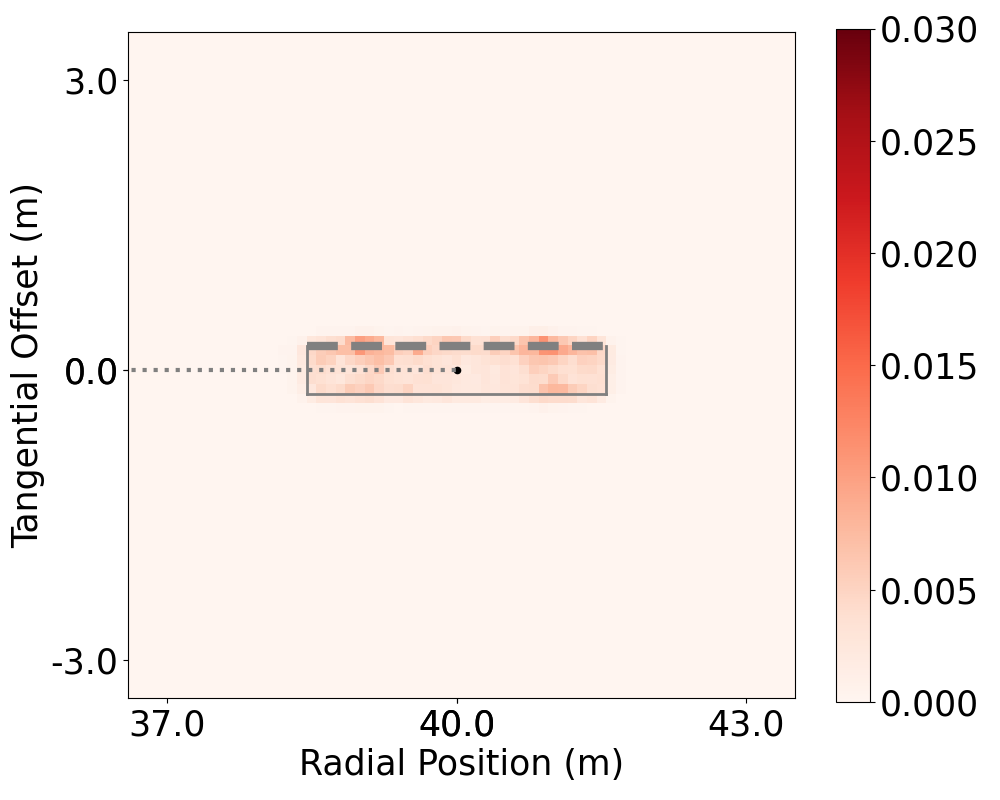}
				\\
				{\small $135^\circ$ } &
				\includegraphics[height=1.2in]{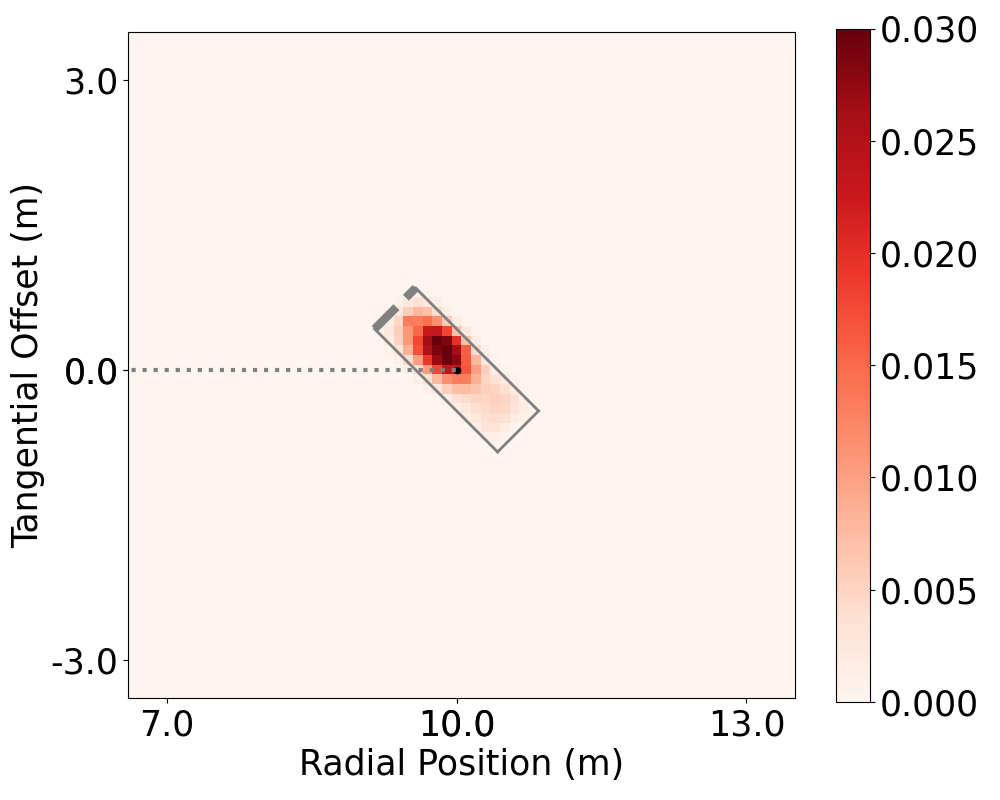}&
				\includegraphics[height=1.2in]{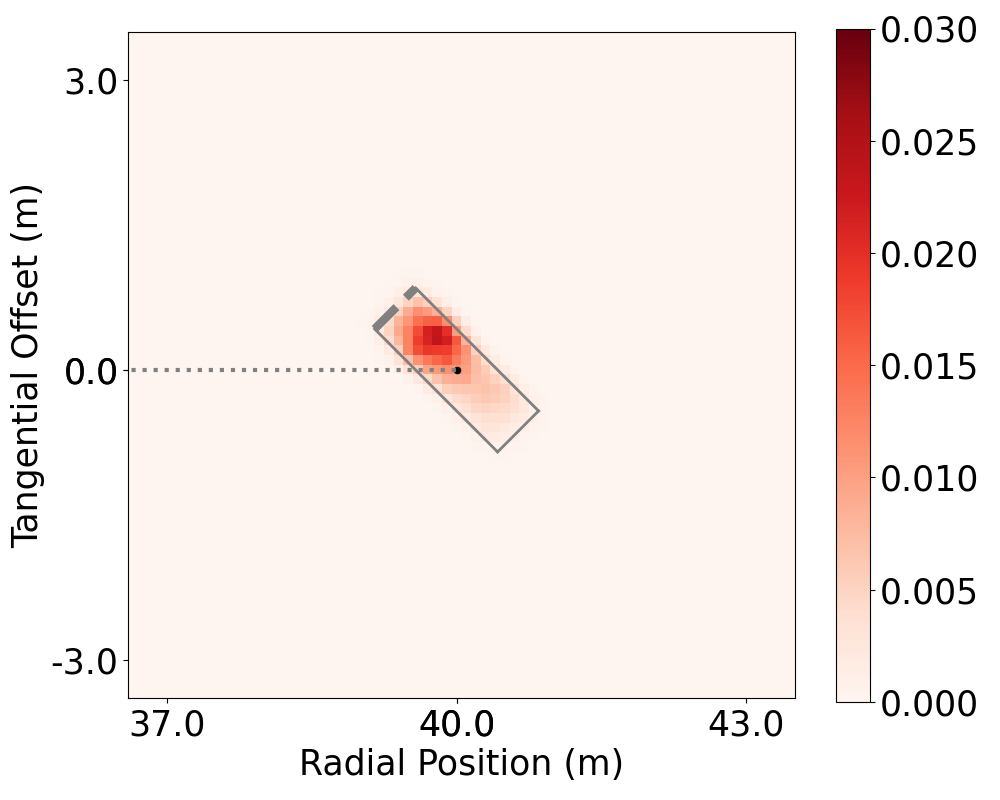}&	
				\includegraphics[height=1.2in]{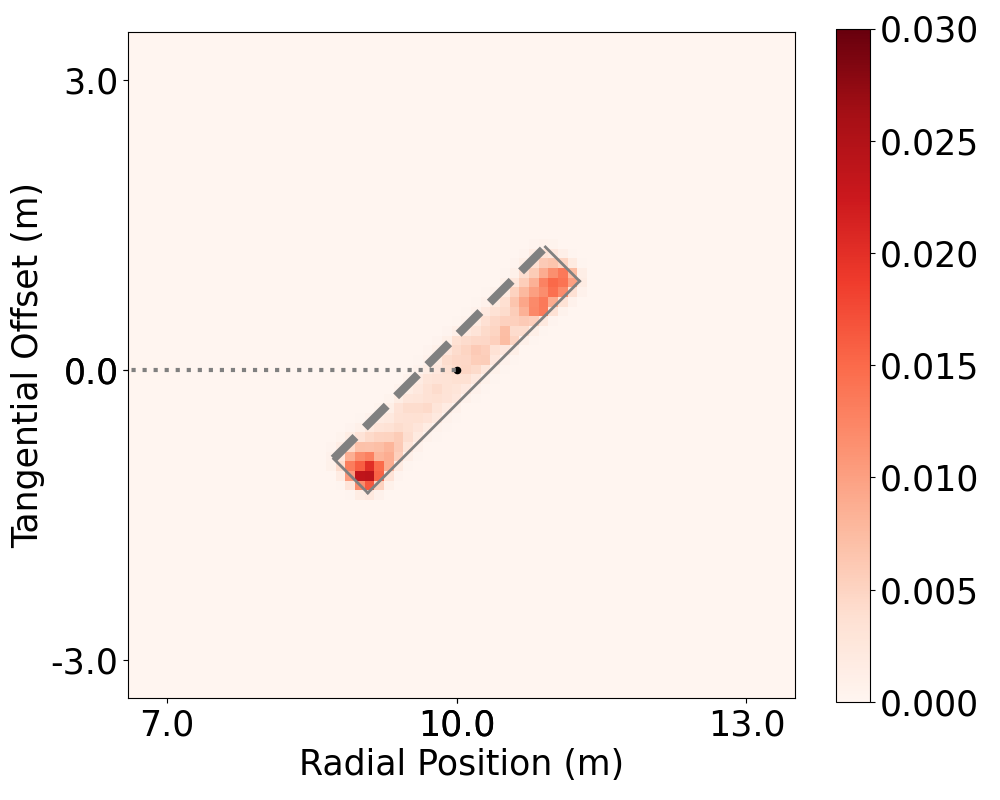}&
				\includegraphics[height=1.2in]{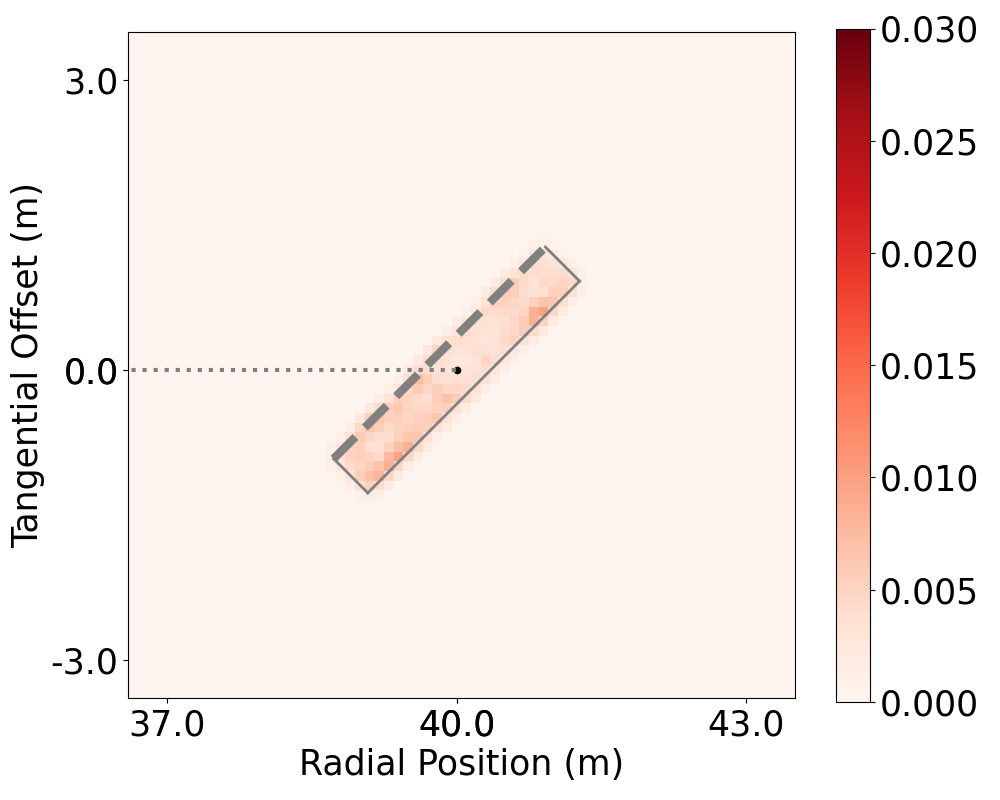}
				\\
				{\small $180^\circ$ } &
				\includegraphics[height=1.2in]{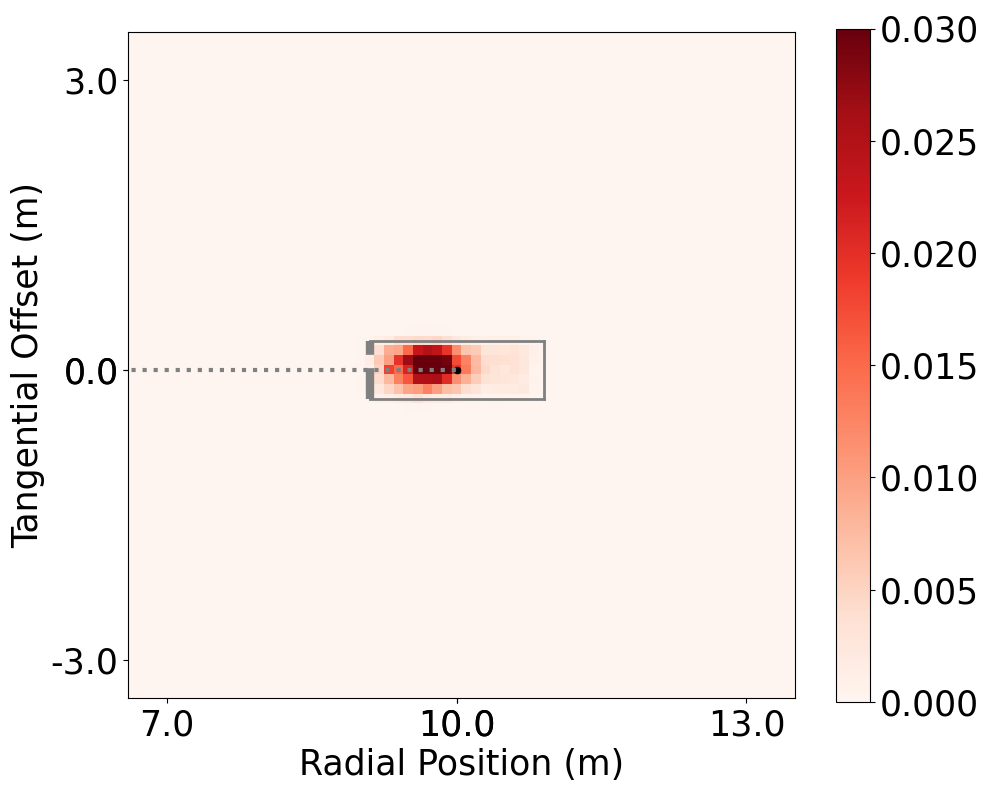}&
				\includegraphics[height=1.2in]{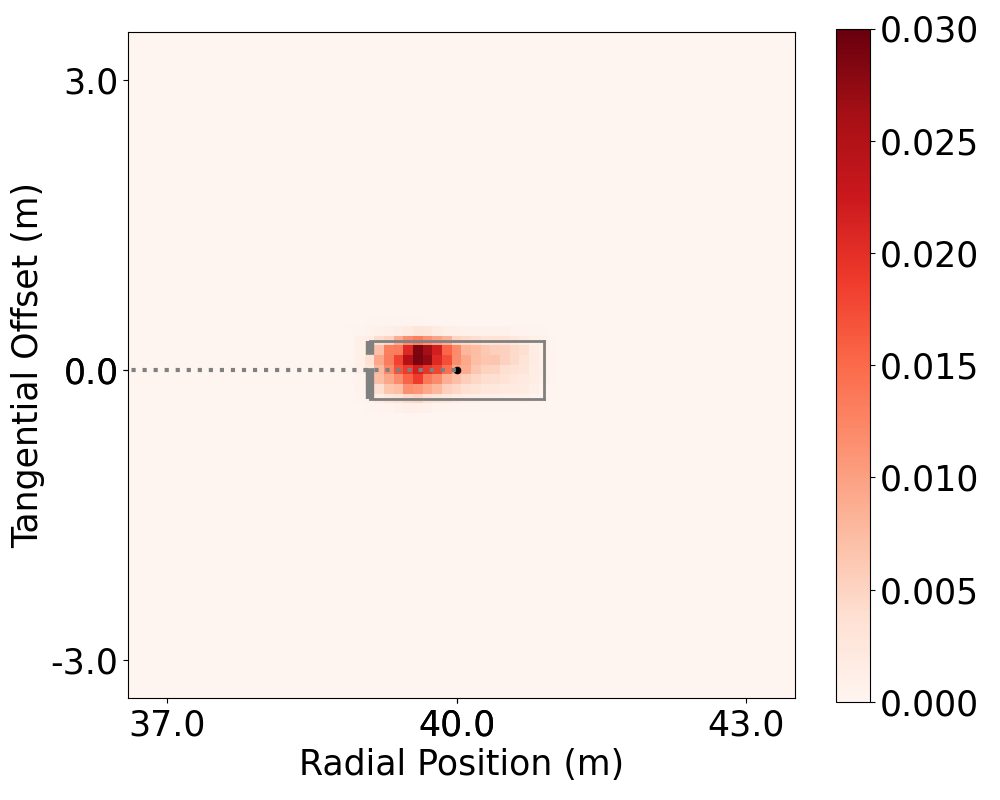}&	
				\includegraphics[height=1.2in]{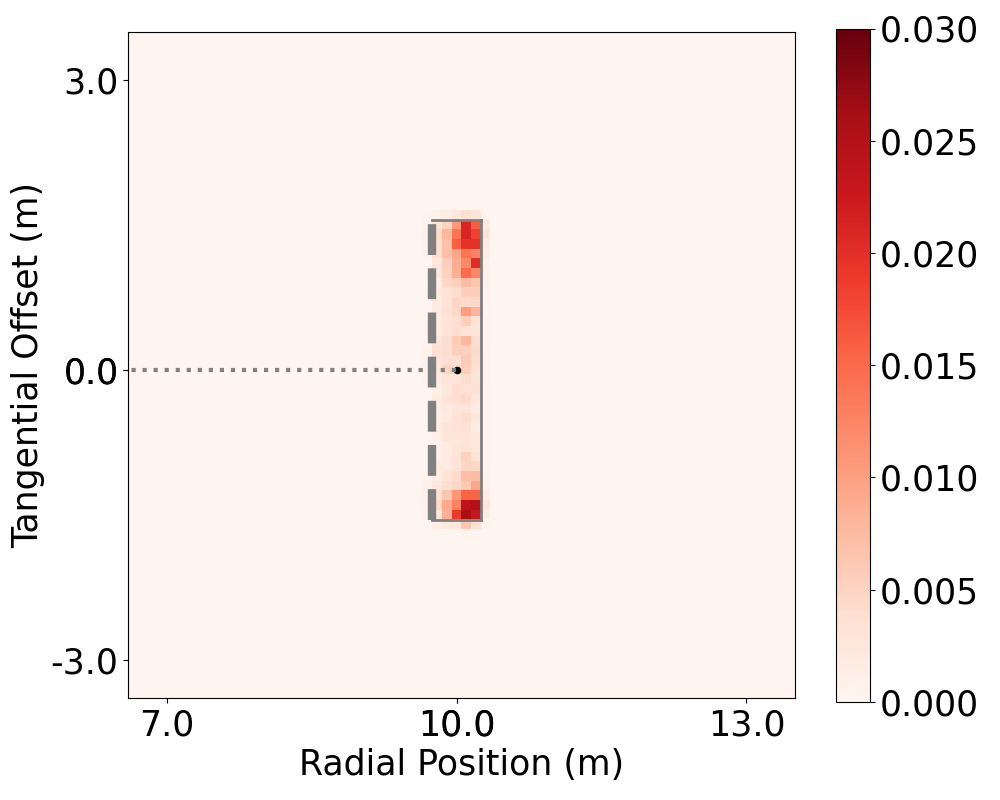}&
				\includegraphics[height=1.2in]{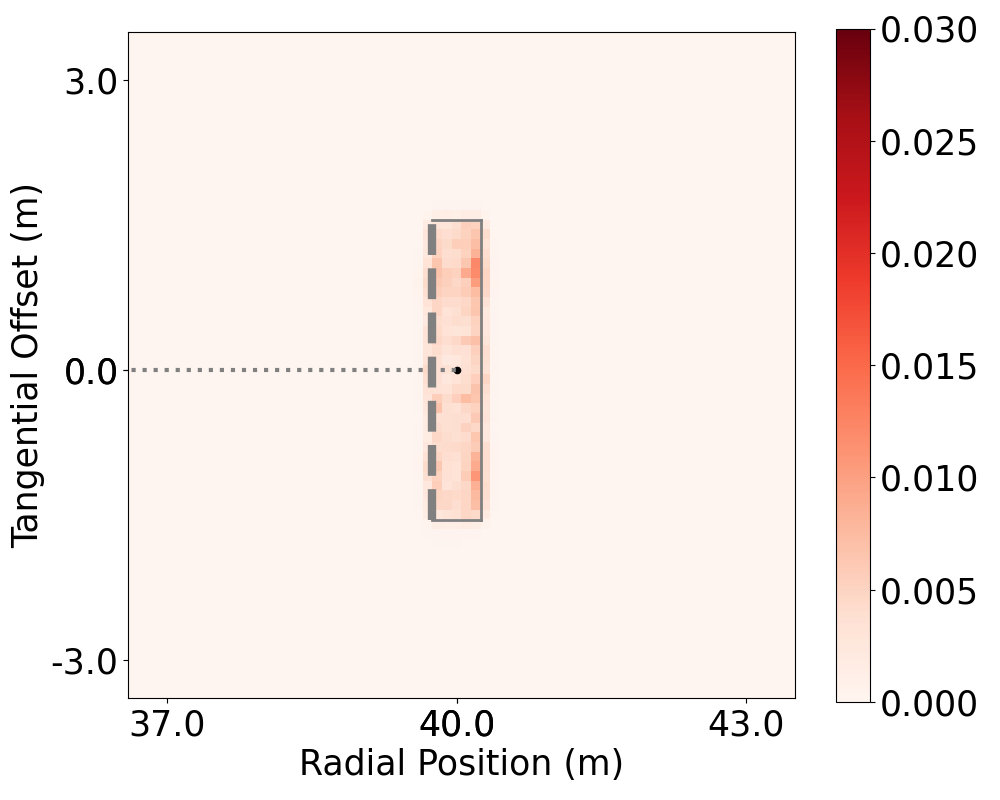}
				\\
				&
				\small (a) Bicycle (Short Range) 	 &
				\small (b) Bicycle (Long Range)      &	
				\small (c) Barrier (Short Range)  &	
				\small (d) Barrier (Long Range) 
				
			\end{tabular}
		}
		\caption{ Visualization of predicted radar distributions of (a)(b) bicycle and (c)(d) barrier viewed from five different angles and from distances of $10$ and $40$ meters.
		}
		\label{fig:suppl_vis2}
	\end{figure*}

	To visualize how predicted distribution varies with viewing angles, we simulate object parameters with different orientations and apply Stage-1 model to generate corresponding radar hit distributions. Figs.~\ref{fig:suppl_vis1} and~\ref{fig:suppl_vis2} shows predicted distributions for car, bus, bicycle, and barrier with different orientations and distances. We can see the distributions vary with category, orientation, and distance. For example, radar distributions are less concentrated spatially at longer range because of larger beam width. We can also notice that distributions of radar points reflected by the tail and head of cars (as shown in the 1st and last row of Fig.~\ref{fig:suppl_vis1}) are different because of their different surface shapes. More visualizations of predicted radar distributions for objects rotating $360$ degrees are shown in the attached video demo.

\end{document}